\documentclass[journal, twoside]{./style/IEEEtran} 

\usepackage{cite}
\usepackage{amsmath,amssymb,amsfonts}
\usepackage{algorithmic}
\usepackage[pdftex]{graphicx} 
\usepackage{textcomp} 
\usepackage{xcolor}
\def\BibTeX{{\rm B\kern-.05em{\sc i\kern-.025em b}\kern-.08em
    T\kern-.1667em\lower.7ex\hbox{E}\kern-.125emX}}

\usepackage{hyperref}

\usepackage{cleveref}
\Crefname{equation}{Eq.}{Eqs.} 
\Crefname{figure}{Fig.}{Figs.}

\graphicspath{{images-src/}{images-bin/}} 

\definecolor{dlrprim1}{HTML}{000000} 
\definecolor{dlrprim2}{HTML}{666666} 
\definecolor{dlrprim3}{HTML}{b9cad2}
\definecolor{dlrprim4}{HTML}{ffffff} 

\definecolor{dlrblue1}{HTML}{00658b} 
\definecolor{dlrblue2}{HTML}{3b98cb}
\definecolor{dlrblue3}{HTML}{6cb9dc}
\definecolor{dlrblue4}{HTML}{a7d3ec}
\definecolor{dlrblue5}{HTML}{d1e8fa}

\definecolor{dlryellow1}{HTML}{d2ae3d}  
\definecolor{dlryellow2}{HTML}{f2cd51} 
\definecolor{dlryellow3}{HTML}{f8de53}
\definecolor{dlryellow3}{HTML}{fcea7a}
\definecolor{dlryellow3}{HTML}{fff8be}

\definecolor{dlrgreen1}{HTML}{82a043} 
\definecolor{dlrgreen2}{HTML}{a6bf51}
\definecolor{dlrgreen3}{HTML}{cad55c}
\definecolor{dlrgreen4}{HTML}{d9df78}
\definecolor{dlrgreen5}{HTML}{e6eaaf}

\definecolor{dlrgray1}{HTML}{666666} 
\definecolor{dlrgray2}{HTML}{868585}
\definecolor{dlrgray3}{HTML}{b1b1b1}
\definecolor{dlrgray4}{HTML}{cfcfcf}
\definecolor{dlrgray5}{HTML}{ebebeb}

\usepackage[textsize=scriptsize]{todonotes}
\newcommand{\annotategraphicsmulti}[3][]{
  \begin{tikzpicture}[%
  every node/.style={draw=black, black, text opacity=1, fill=white, fill opacity=0.75,inner sep=0.5mm, #1},%
  ]
  \node[anchor=south west,inner sep=0, draw=none] (image) at (0,0) {
    #2
  };
  \begin{scope}[x={(image.south east)},y={(image.north west)}]
    #3
  \end{scope}
  \end{tikzpicture}%
}

\newcommand*\lref[1]{\tikz[baseline=(char.base)]{\node[shape=rectangle,draw,inner sep=2pt] (char) {\scriptsize #1};}}

\newenvironment{DIFnomarkup}{}{}

\usepackage{adjustbox}
\newlength{\bioimgw}
\newlength{\bioimgh}
\newcommand{\biophoto}[1]{%
  \clipbox*{%
    {.5\width-.5\bioimgw} {.5\height-.5\bioimgh}
    {.5\width+.5\bioimgw} {.5\height+.5\bioimgh}}{%
    \adjustbox{max size={\bioimgw}{\bioimgh},min size={\bioimgw}{\bioimgh}}{%
      \includegraphics{#1}}}%
}

\usepackage{xfrac}
\usepackage{dblfloatfix}

\usepackage{subcaption}
\usepackage{booktabs}
\usepackage{colortbl}
\definecolor{Light0}{rgb}{0.95, 1, 0.95}
\definecolor{Light1}{rgb}{0.98, 0.95, 0.90}
\definecolor{Light2}{rgb}{0.98, 0.98, 0.93}
\definecolor{Light3}{rgb}{0.98, 0.98, 1}
\definecolor{Light4}{rgb}{0.93, 0.98, 0.98}

\usepackage{bm}
\usepackage{siunitx}
\DeclareSIUnit\gravity{g}
\DeclareSIUnit\year{yrs.}

\usepackage{pgfplots}
\pgfplotsset{compat=1.16}
\usetikzlibrary{arrows.meta}

\usepackage[acronym]{glossaries}
\glsdisablehyper
\newacronym{vf}{VF}{Virtual Fixture}
\newacronym{pcb}{PCB}{printed circuit board}
\newacronym{moe}{MoE}{Mixture of Experts}
\newacronym{poe}{PoE}{Product of Experts}
\newacronym{dtw}{DTW}{dynamic time warping}
\newacronym{gmm}{GMM}{Gaussian Mixture Model}
\newacronym{gmr}{GMR}{Gaussian Mixture Regression}
\newacronym[longplural=Gaussian Processes]{gp}{GP}{Gaussian Process}
\newacronym{kmp}{KMP}{Kernelized Movement Primitive}
\newacronym{rbf}{RBF}{Radial Basis Function}
\newacronym{dmp}{DMP}{Dynamic Movement Primitive}
\newacronym{ds}{DS}{Dynamical System}
\newacronym{seds}{SEDS}{Stable Estimators of Dynamical Systems}
\newacronym[longplural=degrees of freedom]{dof}{DoF}{degree of freedom}

\newcommand{\rmcaffiliation}{German Aerospace Center (DLR), Robotics and Mechatronics Center (RMC), M\"unchener Str. 20, 82234 We\ss ling, Germany.}
\newcommand{\tumaffiliation}{School of Computation, Information and Technology, Sensor Based Robotic Systems and Intelligent Assistance Systems, Technical University of Munich, Friedrich-Ludwig-Bauer-Str. 3, Garching, Germany.}
\newcommand{\idiapaffiliation}{Idiap Research Institute, Martigny, Switzerland.}
\newcommand{\epflaffiliation}{École Polytechnique Fédérale de Lausanne (EPFL), Switzerland.}

\IEEEoverridecommandlockouts

\title{A Unified Framework for Probabilistic Dynamic-, Trajectory- and Vision-based Virtual Fixtures}

\author{Maximilian Mühlbauer$^{1,2}$, Bernhard Weber$^{2}$, Sylvain Calinon$^{3,4}$, Freek Stulp$^2$, Alin Albu-Schäffer$^{2, 1}$,\\João Silvério$^2$
\thanks{The first author partially achieved results within the ACOR project funded by the Federal Ministry BMWK (50RP2240A).}
\thanks{$^1$ \tumaffiliation}  
\thanks{$^2$ \rmcaffiliation}  
\thanks{$^3$ \idiapaffiliation}  
\thanks{$^4$ \epflaffiliation}  
}

\begin{document}

\maketitle


\begin{abstract}
Probabilistic \glspl{vf} enable the adaptive selection of the most suitable haptic feedback for each phase of a task, based on learned or perceived uncertainty.
While keeping the human in the loop remains essential, for instance, to ensure high precision, partial automation of certain task phases is critical for productivity.
We present a unified framework for probabilistic \glspl{vf} that seamlessly switches between manual fixtures, semi-automated fixtures (with the human handling precise tasks), and full autonomy.
We introduce a novel probabilistic \Acrlong{ds}-based \gls{vf} for coarse guidance, enabling the robot to autonomously complete certain task phases while keeping the human operator in the loop.
For tasks requiring precise guidance, we extend probabilistic position-based trajectory fixtures with automation, allowing for seamless human interaction, geometry-awareness and optimal impedance gains.
For manual tasks requiring very precise guidance, we also extend visual servoing fixtures with the same geometry-awareness and impedance behavior.
We validate our approach on different robots, including an evaluation with expert users, showcasing operation modes, the ease of programming fixtures and lower interaction forces and favorable usability compared to a baseline.
\end{abstract}


\section{Introduction}
\IEEEPARstart{V}{irtual} Fixtures (VFs)~\cite{rosenberg1993virtualfixtures,bowyer2014active} guide humans through tasks by providing haptic feedback.
They have been applied to diverse areas such as medical robotics \cite{hagmann2024continous}, manufacturing on Earth~\cite{wicht2025human} and in space~\cite{leutert2024ai,muehlbauer2024ai, muehlbauer2022multiphase, muehlbauer2024probabilistic} as well as in underwater manipulation~\cite{birk2018dexterous}.
Depending on the task phase, fixtures can be based on different perceptual input, e.g., on robot proprioception or visual measurements \cite{muehlbauer2022multiphase,muehlbauer2024probabilistic}.
One main limitation of state-of-the-art \glspl{vf} is, however, that tasks cannot be accomplished without a human in the loop, raising the need for fixtures that can progress autonomously by outputting actions, e.g., velocities, while still keeping the human in full control.
Another limitation is that, when having a set of complementary fixtures with different input modalities, output types and models, as summarized in \Cref{tab:problem_statement}, a principled arbitration between them is required, which is currently not available.
Our main contribution is a framework for the \textit{arbitration} of probabilistic fixtures, ensuring both an optimal guidance for the human and automated operation when needed.
A novel set of \glspl{vf}, providing different types of assistance depending on required guidance precision (\Cref{tab:problem_statement}), forms the backbone of our framework.
\begin{figure}
	\centering
	\begin{DIFnomarkup}
	\tikzstyle{l} = [draw, -latex',thick]
	\resizebox{\columnwidth}{!}{%
	\begin{tikzpicture}
		\node[rectangle,draw=teal,line width=0.5mm,inner sep=10pt] at (-2, 0) (arbitration) {$\hat{\bm{\Sigma}}_{\mathrm{VF}} \sum\limits_{i=1}^{N_\mathrm{VF}} \bm{\Sigma}^{-1}_{\mathrm{VF},i} \bm{\mu}_{\mathrm{VF},i}$};
		\node[above=0cm of arbitration] (arbitration_text) {Arbitration: \ref{sec:problem:arbitration}};
		
		\node[right=0.5cm of arbitration] (wrench_out) {$\hat{\bm{w}}$};
		\draw[-Stealth] (arbitration.east) -- (wrench_out.west);

		\node[rectangle,draw=red,line width=0.5mm,fill=white,inner sep=0pt,minimum width=2cm,minimum height=2.3cm] at (-7.7,3.1) {};
		\node[rectangle,draw=red,line width=0.5mm,fill=white,inner sep=0pt,minimum width=2cm,minimum height=2.3cm] at (-7.8,3) {};
		\node[rectangle,draw=red,line width=0.5mm,fill=white,inner sep=0pt,minimum width=2cm,minimum height=2.3cm] at (-7.9,2.9) {};
		\node[rectangle,draw=red,line width=0.5mm,fill=white,inner sep=0pt,minimum width=2cm,minimum height=2.3cm] at (-8,2.8) (ds) {};
		\node[right=0.3cm of ds,align=center] (dstext) {$\bm{\mu}_{\mathrm{DS},i}$\\$ \bm{\Sigma}_{\mathrm{DS},i}$};
		\draw[-Stealth] (dstext.east) -- (arbitration.north west);
		\node[above right=-0.6cm and 0.3cm of ds,align=left] {Dynamical\\System VF};
		\node[above left=0cm and 0cm of ds] {$N_\mathrm{DS}$};
		
		\node[rotate=90] at (-8.8,3.4) {\small Model};
		\draw[->] (-8.6,3.7) -- (-8.3,3.6);
		\draw[->] (-8.6,3.5) -- (-8.3,3.4);
		\draw[->] (-8.5,3.1) -- (-8.2,3.3);
		\draw[->] (-8.2,3.7) -- (-7.9,3.7);
		\draw[->] (-8.2,3.4) -- (-7.9,3.55);
		\draw[->] (-7.8,3.7) -- (-7.5,3.6);
		\draw[->] (-7.8,3.5) -- (-7.5,3.4);
		\draw[->] (-7.4,3.5) -- (-7.1,3.6);
		\node at (-7.65, 3.1) {\small \ref{sec:non_parametric_DS} / \hyperref[sec:dynamic_fixtures:base_policy]{B}};
		\node[rotate=90] at (-8.8,2.25) {\small Control};
		\node at (-7.85, 2.3) {\small Proportional};
		\node at (-7.4, 1.9) {\small \ref{sec:dynamic_fixtures:vf}};
		
		\node[rectangle,draw=orange,line width=0.5mm,fill=white,inner sep=0pt,minimum width=2cm,minimum height=2.3cm] at (-7.7,0.3) {};
		\node[rectangle,draw=orange,line width=0.5mm,fill=white,inner sep=0pt,minimum width=2cm,minimum height=2.3cm] at (-7.8,0.2) {};
		\node[rectangle,draw=orange,line width=0.5mm,fill=white,inner sep=0pt,minimum width=2cm,minimum height=2.3cm] at (-7.9,0.1) {};
		\node[rectangle,draw=orange,line width=0.5mm,fill=white,inner sep=0pt,minimum width=2cm,minimum height=2.3cm] at (-8,0) (pb) {};
		\node[right=0.3cm of pb,align=center] (pbtext) {$\bm{\mu}_{\mathrm{PB},i}$\\$ \bm{\Sigma}_{\mathrm{PB},i}$};
		\draw[-Stealth] (pbtext.east) -- (arbitration.west);
		\node[above right=-0.6cm and 0.3cm of pb,align=left] {Position-based\\Trajectory VF};
		\node[above left=0cm and 0cm of pb] {$N_\mathrm{PB}$};
		
		\node[rotate=90] at (-8.8,0.6) {\small Model};
		\draw plot[smooth] coordinates {
		    (-8.4,0.5) (-8,0.7) (-7.9,0.5) (-7.6,0.4)
		    (-7.2,0.8) };
		\node at (-7.2, 0.3) {\small \ref{sec:auto_position_based}};
		\node[rotate=90] at (-8.8,-0.55) {\small Control};
		\node[align=center] at (-7.85, -0.5) {\small Variable\\[-5pt] Impedance};
		\node at (-7.4, -0.9) {\small \ref{sec:problem:variable_impedance}};
		
		\node[rectangle,draw=cyan,line width=0.5mm,fill=white,inner sep=0pt,minimum width=2cm,minimum height=2.3cm] at (-7.7,-2.5) {};
		\node[rectangle,draw=cyan,line width=0.5mm,fill=white,inner sep=0pt,minimum width=2cm,minimum height=2.3cm] at (-7.8,-2.6) {};
		\node[rectangle,draw=cyan,line width=0.5mm,fill=white,inner sep=0pt,minimum width=2cm,minimum height=2.3cm] at (-7.9,-2.7) {};
		\node[rectangle,draw=cyan,line width=0.5mm,fill=white,inner sep=0pt,minimum width=2cm,minimum height=2.3cm] at (-8,-2.8) (vs) {};
		\node[right=0.3cm of vs,align=center] (vstext) {$\bm{\mu}_{\mathrm{VS},i}$\\$ \bm{\Sigma}_{\mathrm{VS},i}$};
		\draw[-Stealth] (vstext.east) -- (arbitration.south west);
		\node[above right=-0.6cm and 0.3cm of vs,align=left] {Visual Serv.\\Fixture};
		\node[above left=0cm and 0cm of vs] {$N_\mathrm{VS}$};
		
		\node[rotate=90] at (-8.8,-2.2) {\small Model};
		\node at (-8.5,-1.9)[circle,fill,inner sep=1.5pt]{};
		\node at (-8,-2.5)[circle,fill,inner sep=1.5pt]{};
		\node at (-7.2,-2.05)[circle,fill,inner sep=1.5pt]{};
		\node at (-7.6,-2.2)[circle,fill,inner sep=1.5pt]{};
		\node at (-7.4,-2)[circle,fill,inner sep=1.5pt]{};
		\node at (-8.2,-2)[circle,fill,inner sep=1.5pt]{};
		\node at (-7.25,-2.5) {\small \ref{sec:vs_fixture}};
		\node[rotate=90] at (-8.8,-3.35) {\small Control};
		\node[align=center] at (-7.85, -3.3) {\small Variable\\[-5pt] Impedance};
		\node at (-7.4, -3.7) {\small \ref{sec:problem:variable_impedance}};
		
		\node[left=2cm of pb] (xee) {$\bm{x}_\mathrm{ee}$};
		\draw[-Stealth] (xee.east) -- (ds.west);
		\draw[-Stealth] (xee.east) -- (pb.west);
		\draw[-Stealth] (xee.east) -- (vs.west);
		
		\node[left=2cm of vs] (i) {$\bm{I}$};
		\draw[-Stealth] (i.east) -- (vs.west);
	\end{tikzpicture}}
	\end{DIFnomarkup}
	\caption{\label{fig:intro:overview} Overview of our unified framework. We propose a new type of learned, \textcolor{red}{\Acrlongpl{ds} based \Acrlong{vf}} to assist an operator with \textbf{coarse guidance} in progressing along the task while staying near the training data. For \textbf{precise guidance}, \textcolor{orange}{position-based trajectory fixtures} support the human operator while geometric \textcolor{cyan}{visual servoing fixtures} provide \textbf{very precise guidance}. Input to the fixtures is the end effector pose $\bm{x}_{\mathrm{ee}}$ and an image $\bm{I}$ while the output of each fixture is a probabilistic wrench $\bm{w}_i \sim \mathcal{N} \left( \bm{\mu}_i, \bm{\Sigma}_i \right)$. Core of our framework is a novel variable impedance control scheme as well as an \textcolor{teal}{optimal arbitration} of all probabilistic fixture wrenches.}
\end{figure}
All our \glspl{vf} output a probabilistic wrench with covariance in pose space.
We formulate the fusion of different \glspl{vf} in a principled way using an arbitration scheme (\Cref{sec:problem_statement}), thus solving the problem of selecting and switching between \glspl{vf} in a flexible and adaptive manner based on learned and/or perceived uncertainties.
\Cref{fig:intro:overview} provides an overview of our unified framework.

To create automated \glspl{vf}, we propose a novel probabilistic \gls{vf} for \textbf{coarse guidance} based on learned \textit{\glspl{ds}} \cite{ijspeert2002movement,ijspeert2013dynamical} which have been studied extensively in robotics \cite{saveriano2023dynamic,lopez2023non,fichera2023implicit,khansari2011learning,fichera2022linearization,fichera2024learning,mohammadi2023neural,perez2024deep,perez2023stable} and are a promising approach to model a wide range of autonomously executed tasks.
Commonly, they are equipped with a single attractor and the method ensures that the system converges to that point.
Using probabilistic methods providing epistemic uncertainty we model \glspl{ds} that do not necessarily converge to a single attractor but may also contain recurring motions and can be composed of multiple demonstrated dynamics in different areas of the robot's workspace while supporting user interaction (\Cref{sec:dynamic_fixtures}).

While \gls{ds}-based fixtures offer a very flexible automated assistance, probabilistic \textit{position-based trajectory fixtures} are better suited for \textbf{precise guidance} when a demonstrated path has to be followed.
Although such fixtures are abundant in the literature \cite{raiola2017comanipulation,muehlbauer2024probabilistic}, in scenarios of potential data scarcity, learning efficiency becomes crucial.
To facilitate data efficient learning when object and task geometries are known, we propose an extension of a state-of-the-art formulation \cite{ti2023geometric} to take different geometries into account.
Furthermore, extracting the preferred direction from demonstration data, we automate the execution of such fixtures by extending them with a novel control scheme inspired by the \gls{ds} based \gls{vf} (\Cref{sec:auto_position_based}).

For \textbf{very precise guidance}, particularly near relevant objects in the robot workspace, we leverage probabilistic visual measurements.
The \textit{visual servoing fixture} formulation \cite{muehlbauer2024probabilistic} arbitrates between multiple possible targets.
We extend this formulation to cylindrical and spherical manifolds, taking special arrangements of those targets into account (\Cref{sec:vs_fixture}).

\begin{table}
\scriptsize
\caption{\label{tab:problem_statement}\centering \scshape Types of fixtures considered in our framework.}
\centering
\begin{tabular}{@{}c|cccc@{}}
Fixture & Input & Output & Accuracy & Workspace\\
\hline
\textit{Dynamical System} (\ref{sec:dynamic_fixtures}) & Pose & Velocity & Coarse & Everywhere\\
\textit{Position-based Trajectory} (\ref{sec:auto_position_based}) & Pose & Pose & Precise & Near trajectory\\
\textit{Visual servoing} (\ref{sec:vs_fixture}) & Image & Pose & Very precise & Near target\\
\end{tabular}
\end{table}
Our proposed probabilistic \gls{vf} framework reproduces the whole range of automation levels.
In \textit{manual} mode, a user is guided by non-progressing \glspl{vf} in teleoperation or hand-guided mode.
\textit{Semi-automated} \glspl{vf} automate certain tasks while others are still in the responsibility of a human operator.
Finally, \textit{fully automated} \glspl{vf} act autonomously while a human can still intervene.
The key contributions of our work are:
\begin{enumerate}
	\item An extended arbitration scheme taking different types of \glspl{vf} (\Cref{tab:problem_statement}) and different underlying geometries into account (\Cref{sec:problem:arbitration});
	\item A novel variable stiffness formulation that models couplings between positional and orientational \glspl{dof} (\Cref{sec:problem:variable_impedance});
	\item A novel dynamical system based \gls{vf} (\Cref{sec:dynamic_fixtures});
	\item An extension of position-based fixtures to different manifolds as well as their automation (\Cref{sec:auto_position_based});
	\item An extension of visual servoing fixtures to different manifolds (\Cref{sec:vs_fixture}).
\end{enumerate}

For evaluating our framework, we have implemented it on different robotic systems under different automation levels.
After experimentally validating the individual fixtures (\Cref{sec:evaluation:coupled_stiffness,sec:evaluation:position_stiffness,sec:evaluation:dynamic_vf,sec:evaluation:auto_pb_robograv}),
we evaluate the combination of fixtures both in a fully automated scenario on a space-ready robot (\Cref{sec:evaluation:dynamic_pb_robograv}) as well as in semi-automated scenarios with human interaction (\Cref{sec:evaluation:dynamic_vs_hug,sec:evaluation:all_fixtures_sara}).
We provide background on Riemannian manifolds, probabilistic learning from demonstration and impedance control in \Cref{sec:background}.
Key notations used throughout the work are listed in \Cref{tab:notation}.

\section{Related Work}
Our approach builds on a range of techniques from robot learning and control which we review in this section.
The proposed \gls{ds}-based \gls{vf} formulation (\Cref{sec:dynamic_fixtures}) builds on the learning of \gls{ds} body of work \cite{ijspeert2002movement,ijspeert2013dynamical,khansari2011learning,mohammadi2023neural} as well as on methods for interaction with them, explored in \Cref{sec:rel:dynamical_systems,sec:rel:dynamical_systems_interaction}.
For position-based trajectory fixtures (\Cref{sec:auto_position_based}), automation approaches (\Cref{sec:rel:vf_automation}) and adaptive stiffness scaling (\Cref{sec:rel:stiffness_scaling}) are explored.
Finally, we review techniques for the adaptive arbitration of \glspl{vf} in \Cref{sec:rel:arbitration}.

\subsection{Dynamical Systems}
\label{sec:rel:dynamical_systems}
\glsreset{ds}
\glspl{ds} model actions as a function of the system state.
As such, they can for example encode velocity policies $\dot{\bm{x}} = f(\bm{x})$.
\glspl{dmp} \cite{ijspeert2002movement,ijspeert2013dynamical} model such systems with a single attractor by combining a stable attractor dynamic with forcing terms.
Those forcing terms deform the attractor field to, e.g., follow complex trajectories.
A \gls{dmp} can be learned by optimizing the parameters of its forcing terms to follow a set of demonstrations as closely as possible.
For an overview of popular \gls{dmp}-based approaches, the reader is referred to \cite{saveriano2023dynamic}.

\gls{seds} \cite{khansari2011learning} learns a \gls{gmm} and ensures stability of the resulting \gls{ds} towards an attractor point.
For the case of multiple attractors, \cite{fichera2022linearization} proposes a clustering method.
For complex \glspl{ds}, \cite{fichera2024learning} proposes to learn the non-linearity of the \gls{ds}, allowing for an easy adaptation of the learned dynamics with the use case of obstacle avoidance.
Learning implicit manifolds promises to better model complex \glspl{ds} \cite{fichera2023implicit}.
Neural networks allow for even more powerful estimations of \glspl{ds}.
In \cite{mohammadi2023neural}, global stability of the learned system is ensured through a special network architecture.
Other approaches \cite{perez2024deep,perez2023stable} design a special loss function to shape the learned system.
At the cost of a loss of a global stability proof, the more powerful expressivity of deep neural networks can be leveraged.
Closest to our approach are \cite{franzese2021ilosa,meszaros2022learning,pignat2019bayesian}, which also utilize a stabilizing policy together with the learned nonparametric velocity field.
They however integrate the velocity field to positions, leaving user interaction for future work, which our method enables.

While ensuring convergence to an attractor point, many of the proposed methods exhibit motions which were not demonstrated when far away from demonstrated data and therefore might surprise the operator.
Furthermore, it is not possible to fuse multiple of such motion policies in one unified framework.
We here propose a non-parametric approach based on \glspl{kmp} leveraging the fusion of different policies (\Cref{sec:dynamic_fixtures}).
This allows us to have multiple \glspl{ds} active in the workspace in parallel that can, e.g., each handle a specific portion of a task.
Our formulation without fixed attractor points can also model periodic motions, e.g., limit cycles.
A probabilistic stabilizing policy ensures that the robot always stays close to demonstrated data by computing a velocity towards the closest known dynamic.
By using appropriate Riemannian distance metrics in the kernel function, we can account for the full pose as state input.

\subsection{User Interaction with Dynamical Systems}
\label{sec:rel:dynamical_systems_interaction}
Conventionally, \glspl{ds} are used to program autonomous robot motions.
Interactions from a human are treated as perturbations and therefore cancelled.
In contrast, \cite{papageorgiou2019kinesthetic} design a \gls{dmp} for human interaction.
The \gls{dmp} evolution is synchronized with the human movement and a penetrable \gls{vf} provides force feedback when the user would deviate from the programmed trajectory.
Stiffness scaling is used to allow for user corrections.
In \cite{pervez2019motion}, a similarly time-synchronized \gls{dmp} is designed allowing for a shared control scheme where the autonomous agent controls repetitive \glspl{dof} while the human operator is responsible for the required accuracy in the more variable \glspl{dof}.
Chen et al. \cite{chen2021closed} sample attractor points from a \gls{ds} and employ variable stiffness with impedance control which they also combine with a human operator \cite{xue2023shared}.

Closest to our approach is \cite{amanhoud2019dynamical} who hand-program a \gls{ds} for user interaction.
An impedance controller is used for torque control of the robot allowing for human interactions.
For force-based tasks, a force overlay is added when the robot is in surface contact.
The stabilizing properties of the \gls{ds} are used to bring the robot back to the surface in case of perturbations.

In general, using learned \glspl{ds} as haptic aids for human operators has received little attention so far.
We aim to fill this gap by employing a damping controller (\Cref{sec:dynamic_fixtures:vf}) enabling the guidance of a human operator through a state-based \gls{ds}.
The system state evolves both through the actions of the \gls{ds} as well as through user input.
The probabilistic formulation achieves an optimal fusion of different policies as well as between different \glspl{vf}.

\subsection[Position-based Trajectory VF Automation]{Position-based Trajectory \gls{vf} Automation}
\label{sec:rel:vf_automation}
Position-based \glspl{vf}, which take the robot pose as input and output a desired pose (\Cref{tab:problem_statement}), have been applied to many teleoperation tasks, e.g., in medical \cite{hagmann2021digital,hagmann2024continous} and industrial robotics \cite{pruks2022method}.
They have also been used together with vision \cite{bettini2004vision}.
Closest to our work are the probabilistic trajectory fixtures \cite{raiola2017comanipulation,zeestraten2018programming}.
An overview of different types of fixtures can be found in \cite{abbott2007haptic,bowyer2014active}.
Traditionally, \glspl{vf} constrain the user by keeping them outside of forbidden regions or guiding them along a path without directional guidance \cite{bowyer2014active}.

For our framework, an automated version of such fixtures is required.
To this end, \cite{pezzementi2007dynamic} propose to use the path direction to guide the user along a path.
In \cite{hagmann2024continous}, a cylindrical \gls{vf} is used to move along a path.
For full automation, radius and length of the cylinder are set to $0$ which makes the user follow a given trajectory.
Automation is achieved by introducing a point mass which is accelerated by a user-defined force - this force is counteracted by a virtual damping as well as a damping potentially introduced by the user, therefore limiting the maximum velocity.
Transitions between different levels of autonomy are possible by enlarging or shrinking the guiding cylinder.
None of these approaches however implements more than one concurrently active fixture.
Furthermore, no fusion with other types of fixtures is possible with those works.
Through a probabilistic fusion, our approach allows us to both model multiple concurrently active trajectory fixtures as well as to fuse the automated fixture with other fixtures.

\subsection[Virtual Fixtures with Variable Stiffness]{\Acrlongpl{vf} with Variable Stiffness}
\label{sec:rel:stiffness_scaling}
Impedance control \cite{hogan1984impedance} allows for compliant interaction with the environment and is therefore crucial for safe manipulation as well as robot-human interaction.
A key element of this approach is the choice of stiffness matrix, as this matrix determines the relationship between position offset and exerted forces.
This key property can be changed according to the task needs using variable stiffness formulations.

To this end, \cite{michel2021bilateral} suggest to learn the coupling stiffness for teleoperation.
Learned from task properties, the coupling between input device and remote robot is realized with a low stiffness for safe interactions or with high stiffness to achieve a high tracking accuracy.
Other approaches \cite{chen2021closed,xue2023shared} learn a \gls{ds} with variable stiffness along an orthogonal to the motion direction and furthermore demonstrate it in a shared control application.
The approach is also extended to incorporate a learned stiffness for rotational \glspl{dof} \cite{michel2023orientation}.

Closer to our requirements is \cite{balachandran2023passive}, where, based on the uncertainty of a perception algorithm, the stiffness of the robot controller is adapted through a scaling factor.
Covariance matrices can, however, model complex relationships between the individual \glspl{dof} which cannot be represented by a scalar factor.
In \cite{abifarraj2017learning}, a block diagonal matrix with submatrices for position and orientation stiffness based on the covariance matrix of a learned trajectory is used.
While this choice of stiffness matrix can model a wide range of stiffness behaviors, couplings between positional and orientational \glspl{dof} which, e.g., full covariance matrices can represent are neglected.

As our experiments underline (\Cref{sec:evaluation:coupled_stiffness}), covariance matrices with couplings between position and orientation are important for variable stiffness control.
Note that simply using a scaled version of the precision matrix $\bm{P} = \bm{\Sigma}^{-1}$ is not possible as this could lead to stiffness values too high  for a physical system to be stable.
We therefore build upon the findings of \cite{huang1998achieving,huang2000eigenscrew} who decompose a stiffness matrix into eigenscrews.
This can be performed similarly using precision matrices to design a desired stiffness matrix with such couplings while respecting stiffness limits (\Cref{sec:problem:variable_impedance}).

\subsection[Virtual Fixture Arbitration]{\Acrlong{vf} Arbitration}
\label{sec:rel:arbitration}
As shown previously \cite{muehlbauer2024probabilistic}, a function to arbitrate between different \glspl{vf} is required, which extends the concept of \textit{arbitration} between human operator and system (see \cite{selvaggio2021autonomy} for a survey).
To this end, special controllers to stabilize hard switches \cite{selvaggio2016enhancing} or hand-tuned weights \cite{firas2020haptic,muehlbauer2022multiphase} can be used.
One major limitation of these works is that the arbitration function or stabilizing controller does not make use of information from the fixtures but instead needs to be handcrafted.

This limitation can be resolved using probabilistic formulations \cite{raiola2017comanipulation}.
Further extensions allow for an individual weighting along all \glspl{dof} instead of a single scalar weight value.
This can also be used to arbitrate between system and human operator \cite{zeestraten2018programming,michel2021bilateral}.
In a previous work \cite{muehlbauer2024probabilistic} a Gaussian product has been used to perform \gls{dof}-specific arbitration.
We build on this foundation to include all fixtures highlighted in \Cref{tab:problem_statement} in a unified framework, showing how \gls{ds} based \glspl{vf} (\Cref{sec:dynamic_fixtures}) and \glspl{vf} on different geometries can be fused (\Cref{sec:problem:arbitration}).

\section{Background}
\label{sec:background}
\subsection{Riemannian Manifolds and Probabilities}
\label{sec:background:on_manifold}
\begin{figure}
\centering
\begin{subfigure}{0.32\columnwidth}
  \centering
  \includegraphics[width=\columnwidth,trim={120 65 120 65},clip]{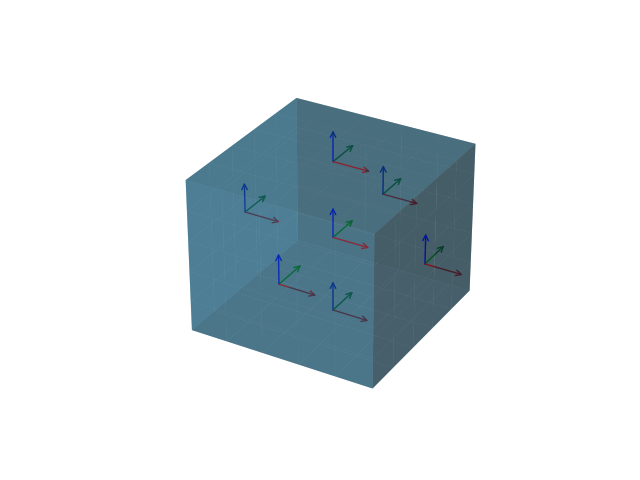}
  \subcaption{$\mathcal{M}_1$: $\mathbb{R}^3 \times \mathcal{S}^3$.}
\end{subfigure}
\hfill
\begin{subfigure}{0.32\columnwidth}
  \centering
  \includegraphics[width=\columnwidth,trim={120 65 120 65},clip]{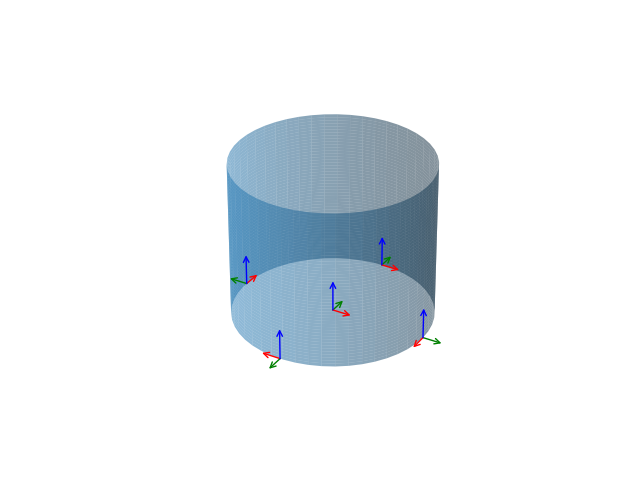}
  \subcaption{$\mathcal{M}_2$: $\mathcal{S}^1 \times \mathbb{R}^2 \times \mathcal{S}^3$.}
\end{subfigure}
\hfill
\begin{subfigure}{0.32\columnwidth}
  \centering
  \includegraphics[width=\columnwidth,trim={150 80 130 100},clip]{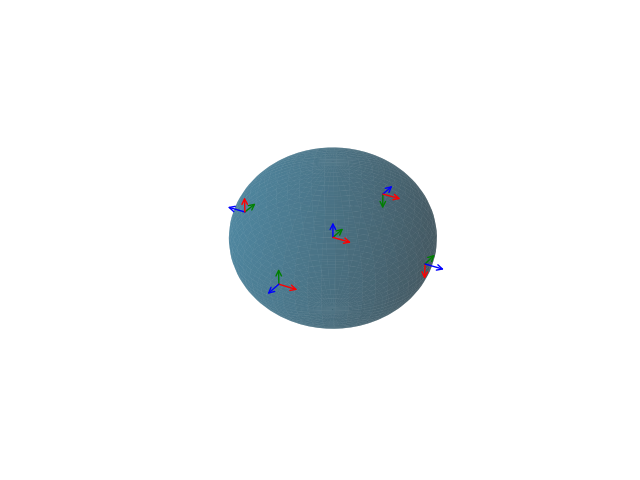}
  \subcaption{$\mathcal{M}_3$: $\mathcal{S}^2 \times \mathbb{R}^1 \times \mathcal{S}^3$.}
\end{subfigure}%
\caption{Manifolds used in this work inspired by and using the notation of \cite{ti2023geometric}. Depending on the task, properties can be expressed more efficiently in cylindrical ($\mathcal{M}_2$) or spherical ($\mathcal{M}_3$) compared to Cartesian ($\mathcal{M}_1$) coordinates. The coordinate systems in each image depict the orientation basis, i.e., the unit quaternion $(0, 0, 0, 1)^\top$ for different positions on the manifold.}
\label{fig:background:manifolds}
\end{figure}
The geometry of a task can be modeled using Riemannian manifolds -- this is already required for orientations which are non-Euclidean.
Following the notation and intuition of \cite{ti2023geometric} on when to use which manifold, we consider the manifolds $\mathcal{M}$ shown in \Cref{fig:background:manifolds} to express a full pose $\bm{x}$:
\begin{enumerate}
	\item $\mathcal{M}_1$ with ${\bm{x} \in \mathbb{R}^3 \times \mathcal{S}^3}$: We use this manifold to represent Cartesian poses as product of the position expressed in 3-dimensional Euclidean space and a unit quaternion\footnote{To avoid issues with $\mathcal{S}^3$ double-covering $SO(3)$, we wrap the logarithm at a full rotation, ensuring that $\mathrm{Log}_{\bm{q}}\!\!\left(-\bm{q}\right) = \bm{0}$.}.
	\item $\mathcal{M}_2$ with $\bm{x} \in \mathcal{S}^1 \times \mathbb{R}^2 \times \mathcal{S}^3$: This manifold represents cylindrical poses as product of the angle $\phi$ from the $x$-axis, the radius $r$ measured as distance from the origin in the $xy$ plane and the $z$ coordinate and a unit quaternion. The base of the orientation is adjusted such that its $y$ axis is always pointing in direction of increasing radius.
	\item $\mathcal{M}_3$ with $\bm{x} \in \mathcal{S}^2 \times \mathbb{R}^1 \times \mathcal{S}^3$: We use this manifold to represent spherical poses as product of the angles from the $x$ and $z$ axis (often denoted as \textit{azimuthal} angle $\phi$ and \textit{polar} angle $\theta$), the radius $r$ and a unit quaternion. The base of the orientation is adjusted such that the $z$ axis is always pointing in direction of increasing radius.
\end{enumerate}

The manifold logarithm $\mathrm{Log}_{\bm{x}_1}^{\mathcal{M}} \bm{x}_2$ on $\mathcal{M}$ calculates the tangent vector $\bm{u}_{12} \in \mathcal{T}_{\bm{x}_1}\mathcal{M}$ from $\bm{x}_1$ in the direction of $\bm{x}_2$; its magnitude is equal to the geodesic distance between the points.
Its inverse, the manifold exponential $\mathrm{Exp}_{\bm{x}_1}^{\mathcal{M}} \bm{u}_{12}$, recovers $\bm{x}_2$ on the manifold.
Parallel transport is required to move vectors between different tangent spaces centered at $\bm{x}_i$.
The reader is referred to  \cite{zeestraten2017manifold,calinon2020gaussians} for a more exhaustive treatment of these Riemannian operations.
Between tangent spaces of different manifolds, manifold-specific Jacobians $\bm{J}_\mathcal{M}$ (see \Cref{sec:appendix:coordsys}) transform vectors \textit{contravariantly} \cite{lee2003introduction}.

With the logarithm map and the Gaussian distribution proposed by \cite{zeestraten2017manifold,calinon2020gaussians}, we compute the probability of $\bm{x}$ to
\begin{equation}
	\mathcal{N}\left(\bm{x}\vert\bm{\mu},\bm{\Sigma}\right) = \frac{1}{\sqrt{\left(2\pi\right)^d\vert\bm{\Sigma}\vert}}e^{-\frac{1}{2}\mathrm{Log}_{\bm{\mu}}^{\mathcal{M}}\!\left(\bm{x}\right)^\top\bm{\Sigma}^{-1}\mathrm{Log}_{\bm{\mu}}^{\mathcal{M}}\!\left(\bm{x}\right)},
\end{equation}
parameterized by mean $\bm{\mu}\in\mathcal{M}$ and covariance $\bm{\Sigma} \in \mathcal{T}_{\bm{\mu}} \mathcal{M}$.

Finally, a weighted distance measure between $\bm{x}_1$ and $\bm{x}_2$ is often required.
For this, we define the on-manifold distance
\begin{align}
	d_{\bm{A}}^{\mathcal{M}}(\bm{x}_1, \bm{x}_2) &= \left|\left| \mathrm{Log}_{\bm{x}_1}^{\mathcal{M}} \bm{x}_2 \right|\right|_{\bm{A}}^2\nonumber\\
	&= \mathrm{Log}_{\bm{x}_1}^{\mathcal{M}} \left(\bm{x}_2\right)\!\!^\top \bm{A} \>  \mathrm{Log}_{\bm{x}_1}^{\mathcal{M}} \left(\bm{x}_2\right).
	\label{eq:distance_function_onman_weighted}
\end{align}
with a weighting matrix $\bm{A}$ expressed in tangent space $\mathcal{T}_{\bm{x_1}} \mathcal{M}$.
Note that, e.g., in the case of $\mathcal{S}^2$, $\mathrm{Log}_{\bm{x}_1}^{\mathcal{S}^2} \bm{x}_2 \neq -\mathrm{Log}_{\bm{x}_2}^{\mathcal{S}^2} \bm{x}_1$ as would hold for the other manifolds used in this work.
For a change of basis, $\bm{A}$ therefore would have to be rotated.

\subsection[Impedance-controlled Virtual Fixtures]{Impedance-controlled \Acrlongpl{vf}}
Our \Acrlongpl{vf} framework (\Cref{sec:problem_statement}) outputs a Cartesian wrench $\bm{w}_\mathrm{VF}$ which is applied to the end effector.
Assuming a gravity-compensated, torque-controlled manipulator, the corresponding desired joint torques evaluate to
\begin{equation}
	\label{eq:robot_torque_control}
	\bm{\tau} = \bm{J}^\top \bm{w}_{\mathrm{VF}}.
\end{equation}
Note that $\bm{w}_\mathrm{VF}$ is a covector in cotangent space $\mathcal{T}^*_{\bm{x}_\mathrm{ee}} \mathcal{M}$ ($\eta: \mathcal{T}_{\bm{x}_\mathrm{ee}} \mathcal{M} \rightarrow \mathbb{R}$) requiring the covariant transformation with $\bm{J}_\mathcal{M}^\top$ \cite{lee2003introduction} following from the conservation of power $\bm{\tau}^\top \dot{\bm{q}} = \bm{w}^\top \dot{\bm{x}}$.
We assume that $\bm{w}_{\text{VF}}$ is a combination of $N_\mathrm{VF} = N_\mathrm{DS}+N_\mathrm{PB}+N_\mathrm{VS}$ individual wrenches associated with different \glspl{vf}.
For position-based \glspl{vf} (\Cref{sec:auto_position_based}), this wrench is computed as
\begin{equation}
	\bm{w}_{\mathrm{VF},i} = \bm{K}_{\mathrm{VF},i} \mathrm{Log}_{\bm{x}_{\mathrm{ee}}}^{\mathcal{M}}\!\!\left(\bm{x}_{\mathrm{VF},i}\right) + \bm{D}_{\mathrm{VF},i} \frac{\mathrm{d}}{\mathrm{d}t}\mathrm{Log}_{\bm{x}_{\mathrm{ee}}}^{\mathcal{M}}\!\!\left(\bm{x}_{\mathrm{VF},i}\right),
	\label{eq:VF_wrench}
\end{equation}
where $\bm{x}_{\mathrm{ee}} \in \mathcal{M}$ is the end effector pose.
$\bm{K}_{\mathrm{VF},i}$, $\bm{D}_{\mathrm{VF},i}$ and $\bm{x}_{\mathrm{VF},i}$ are the stiffness, damping and attractor of the fixture.
$\mathrm{Log}_{\bm{x}_{\mathrm{ee}}}^{\mathcal{M}}\!\!\left(\bm{x}_\mathrm{VF}\right)$ denotes the manifold logarithm \cite{zeestraten2017manifold} of $\bm{x}_\mathrm{VF}$ at $\bm{x}_{\mathrm{ee}}$, which is the on-manifold equivalent to the Euclidean $\bm{x}_{\mathrm{VF},i}-\bm{x}_{\mathrm{ee}}$, taking orientation and different manifolds $\mathcal{M}$ into account.
$\frac{\mathrm{d}}{\mathrm{d}t}\mathrm{Log}_{\bm{x}_{\mathrm{ee}}}^{\mathcal{M}}\!\!\left(\bm{x}_\mathrm{VF}\right)$ is the corresponding time derivative.
We compute a single Cartesian wrench $\bm{w}_{\mathrm{VF}}$ for \eqref{eq:robot_torque_control} from impedance- and velocity-controlled fixtures $\bm{w}_{\mathrm{VF},i}$ on different manifolds through a probabilistic fusion (\Cref{sec:problem:arbitration}).

\subsection{Probabilistic Learning from Demonstration}
\label{sec:background:lfd}
Different probabilistic models encoding assistive behaviors can be learned from demonstration data using Gaussian distributions.
We are specifically interested in modeling both the \textit{aleatoric} uncertainty which is inherent to the data, i.e., the variability in demonstrations, and the \textit{epistemic} uncertainty, which is the uncertainty caused by a lack of data.
We argue that fixture activation should be inversely proportional to uncertainty -- both epistemic and aleatoric -- such that strong guidance corresponds to low uncertainty levels.

\glspl{gmm}\cite{calinon2015tutorial,calinon2020gaussians} encode the joint distribution between input $\bm{x}$ and output $\bm{y}$ with $M$ Gaussians, i.e.,
\begin{equation}
		\label{eq:background:probabilistic:joint_gmm_encoding}
	\begin{bmatrix} \bm{x} \\ \bm{y} \end{bmatrix} \sim \sum_{m=1}^{M} \pi_m \mathcal{N} \left(\left.\begin{bmatrix} \bm{x} \\ \bm{y} \end{bmatrix}\right| \bm{\mu}_m,\bm{\Sigma}_m\right).
\end{equation}
The conditional distribution of $\bm{y}$ given $\bm{x}$ can be computed as
\begin{equation}
	p(\bm{y}\vert \bm{x}) = \sum_{m=1}^{M} \pi_m(\bm{x}) \mathcal{N} (\bm{y}\vert\bm{\mu}_{m|\bm{x}},\bm{\Sigma}_{m|\bm{x}})
	\label{eq:background:probabilistic:gmm_multimodal}
\end{equation}
using \gls{gmr}.
Note that this is a \gls{moe} \cite{jacobs1991adaptive} model, computing a multi-modal weighted sum of different experts represented by Gaussian distributions in an ``or'' operation.
Subsequently, a unimodal approximation can be computed as \cite{calinon2015tutorial}
\begin{equation}
	\label{eq:background:probabilistic:gmm_unimodal}
	p(\bm{y}\vert \bm{x}) = \mathcal{N} (\bm{y}\vert\bm{\mu}_\mathrm{GMR},\bm{\Sigma}_\mathrm{GMR})
\end{equation}
which is used in (\Cref{sec:auto_position_based}) to derive position-based trajectory fixtures.
For geometry-awareness we use the on-manifold Gaussian operations from \cite{zeestraten2017manifold}.
The \gls{gmm}'s covariance reflects \textit{aleatoric} uncertainty which is used to scale stiffness (\Cref{sec:problem:variable_impedance}), preserving the demonstrated variability.

Kernel-based methods such as \glspl{gp} \cite{williams2006gaussian} and \glspl{kmp} \cite{huang2019kernelized} output \textit{epistemic} uncertainty which increases with the distance between training data and new inputs.
Notably, the latter is capable of capturing both \textit{aleatoric} and \textit{epistemic} uncertainties.
This is used in the \gls{ds}-based \gls{vf} (\Cref{sec:dynamic_fixtures}) to find validity regions for learned dynamics.
In \glspl{kmp}, the posterior mean and covariance of a function $y = f(x)$ are computed.
For a test point $\bm{x}^*$, we obtain
\begin{align}
	\bm{\mu}^* &= \bm{K}(\bm{x}^*, \bm{x}) \left( \bm{K}(\bm{x}, \bm{x}) + \lambda \bm{\Sigma} \right)^{-1} \bm{\mu}\label{eq:background:kmp_pred_1}\\
	\bm{\Sigma}^* &= \alpha \left( \bm{K}(\bm{x}^*, \bm{x}^*) \right. \nonumber\\
	 &~~~- \left. \bm{K}(\bm{x}^*, \bm{x}) \left( \bm{K}(\bm{x}, \bm{x}) + \lambda_c \bm{\Sigma} \right)^{-1} \bm{K}(\bm{x}, \bm{x}^*) \right). \label{eq:background:kmp_pred_2}
\end{align}
$\bm{k}(\bm{x}_i,\bm{x}_j) = k(\bm{x}_i,\bm{x}_j)\bm{I}$, where $k(\bm{x}_i, \bm{x}_j)$ is a kernel function\footnote{We use \eqref{eq:distance_function_onman_weighted} with $\bm{A} = \bm{I}$ to calculate the on-manifold distance in the kernel. More complex treatments might be required in other use cases \cite{jacquier2022geometry}.}, ${\bm{\mu}=[\bm{\mu}^\top_1,\ldots,\bm{\mu}^\top_N]^\top}$ and ${\bm{\Sigma}=\mathrm{blockdiag}\left(\bm{\Sigma}_1,\ldots,\bm{\Sigma}_N\right)}$ are set from a probabilistic reference.
$\lambda$, $\lambda_c$ and $\alpha$ are regularization and scaling hyperparameters.
Please see \cite{huang2019kernelized} for a detailed derivation. Note the connection to classical \glspl{gp} \cite{williams2006gaussian}, where we have $\lambda_c=\lambda$ and $\bm{\Sigma}=\bm{I}$ (\textit{homoscedasticity}), $\alpha=1$ and $\bm{\mu}$ are directly the observations $\bm{y}$.

Treating \gls{gmm} and \gls{kmp} predictions as experts, we fuse outputs using a \gls{poe} \cite{hinton1999product} in an ``and'' operation where each constraint is satisfied approximately, allowing for an arbitration between different types and representations of \glspl{vf} (\Cref{sec:problem:arbitration}).

\section[Virtual Fixture Framework]{\Acrlong{vf} Framework}
\label{sec:problem_statement}
\glsreset{vf}
Depending on the task at hand, \glspl{vf} with different properties are required to optimally guide an operator or to automate it.
\Cref{tab:problem_statement} summarizes properties of individual \glspl{vf} we consider.
All our fixtures output probabilistic wrenches with on-manifold, multivariate Gaussian uncertainties
\begin{equation}
	p(\bm{w}_{\mathrm{VF},i}) = \mathcal{N}(\bm{\mu}_{\mathrm{VF},i}, \bm{\Sigma}_{\mathrm{VF},i})
\end{equation}
where $\bm{\mu}_{\mathrm{VF},i}$ is the mean of the wrench calculated by each fixture on its specific manifold (\Cref{fig:background:manifolds}), under the assumption that $\bm{w}_{\mathrm{VF},i}$ is Gaussian-distributed.
This allows for a natural arbitration of fixtures using the pose-space covariance $\bm{\Sigma}_{\mathrm{VF},i}$, taking \gls{dof}-specific uncertainty and correlations into account (\Cref{sec:problem:arbitration}).
For impedance-controlled \glspl{vf}, we modulate the stiffness using the covariance.
Our novel approach to compute fully populated stiffness matrices is detailed in \Cref{sec:problem:variable_impedance}.

\Cref{fig:problem:arbitration_coords} shows the different coordinate systems used by our method.
An easy transfer between different object placements is possible through the task coordinate system.
Coordinates for individual fixtures placed relative to this coordinate system allow us to exploit the properties of different geometries.

\subsection[Virtual Fixture Arbitration]{\Acrlong{vf} Arbitration}
\label{sec:problem:arbitration}
Previous works \cite{muehlbauer2024probabilistic} use a \gls{poe} \cite{hinton1999product} to arbitrate different fixture wrenches expressed in the same manifold.
We additionally consider per-fixture manifolds depicted in \Cref{fig:background:manifolds}.
Each fixture calculates a mean wrench $\bm{\mu}_{\mathrm{VF},i,\mathcal{M}}$ in cotangent space $\mathcal{T}^*_{\bm{x}_\mathrm{ee}}\mathcal{M}$ through \eqref{eq:VF_wrench}.
This prohibits a direct fusion of the wrenches as the entries in the covector correspond to different \glspl{dof}.
To stay robot-agnostic, we propose to transform covariances and mean wrenches into the (co)tangent space $\mathcal{T}^{(*)}_{\bm{x}_\mathrm{ee}}\mathcal{M}_1$ using
\begin{align}
	\bm{\mu}_{\mathrm{VF},i,\mathcal{M}_1} &= \bm{J}_{i,\mathcal{M}}^\top \bm{\mu}_{\mathrm{VF},i,\mathcal{M}}\\
	\bm{\Sigma}_{\mathrm{VF},i,\mathcal{M}_1} &= \bm{J}_{i,\mathcal{M}}^{-1} \bm{\Sigma}_{\mathrm{VF},i,\mathcal{M}} {\bm{J}_{i,\mathcal{M}}^{-1}}^{\top}
\end{align}
with the manifold Jacobian $\bm{J}_{i,\mathcal{M}} = \frac{\partial \bm{x}_{\mathrm{ee},i,\mathcal{M}}}{\partial \bm{x}_{\mathrm{ee},\mathcal{M}_1}}$ \cite{dyck2022impedance} given in \Cref{sec:appendix:jacobians}.
This corresponds to the transformation of the covector $\bm{\mu}_{\mathrm{VF},i}$ from $\mathcal{T}^*_{\bm{x}_\mathrm{ee}}\mathcal{M}$ into $\mathcal{T}^*_{\bm{x}_\mathrm{ee}}\mathcal{M}_1$ as well as of the twice contravariant tensor $\bm{\Sigma}_\mathrm{VF}$ from $\mathcal{T}_{\bm{x}_\mathrm{ee}}\mathcal{M}$ into $\mathcal{T}_{\bm{x}_\mathrm{ee}}\mathcal{M}_1$.

\begin{figure}
	\centering
	\annotategraphicsmulti{
	    \includegraphics[width=\columnwidth,trim={500 370 600 180},clip]{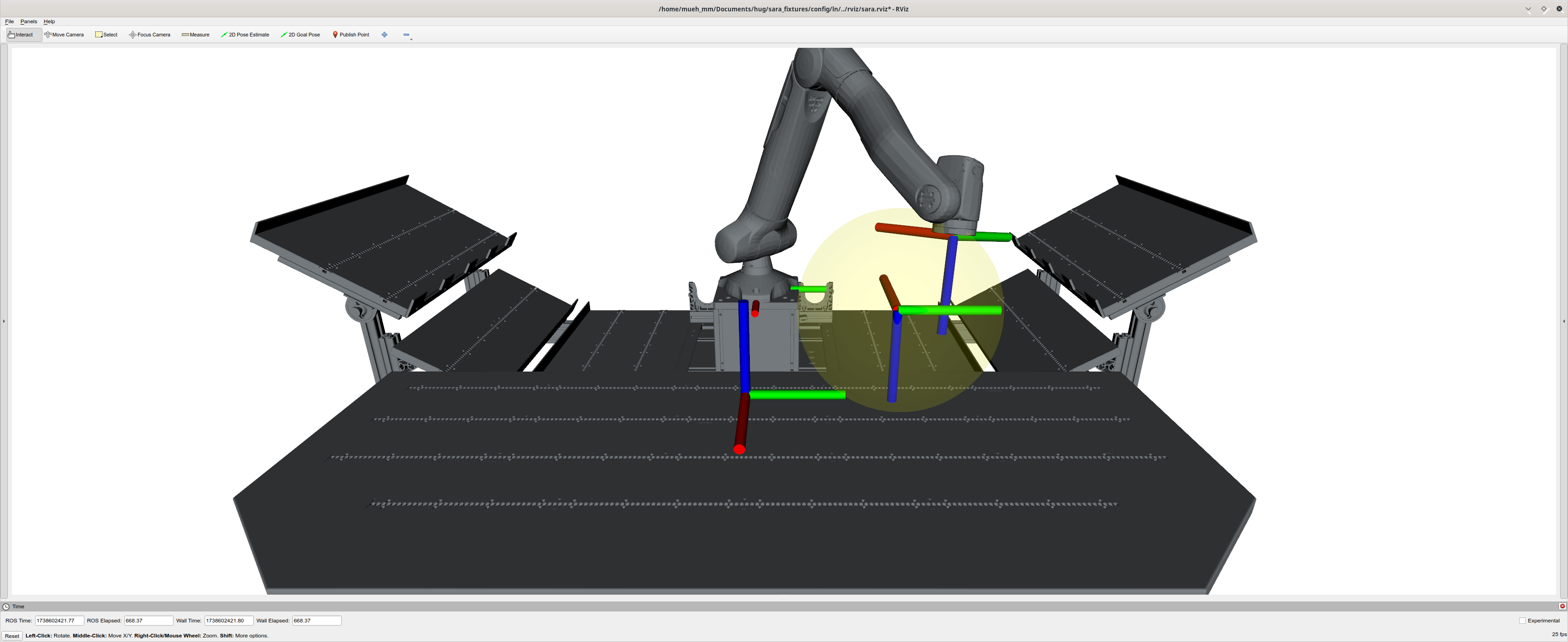}
	}{
        \node at (0.4,0.6) {BASE};
        \node at (0.75,0.6) {TOOL};
        \node at (0.55,0.3) {TASK};
        \node at (0.72,0.42) {FIXTURE};
    }
	\caption{\label{fig:problem:arbitration_coords} The~\protect\lref{TASK} coordinate system defined with respect to the robot~\protect\lref{BASE} accounts for different object placements in the workspace. The specification of~\protect\lref{FIXTURE} coordinate systems relative to the \protect\lref{TASK} coordinate system is crucial for cylindrical and spherical coordinates. Finally, the~\protect\lref{TOOL} frame depends on the current end effector pose of the robot.}
\end{figure}

The wrench $\bm{\mu}_{\mathrm{VF},i,\mathcal{M}_1}$ in cotangent space $\mathcal{T}^*_{\bm{x}_\mathrm{ee}}\mathcal{M}_1$ expresses the forces of the wrench in the \lref{FIXTURE} coordinate system while the torques are expressed in \lref{TOOL} coordinates (\Cref{fig:problem:arbitration_coords}).
Compared to that, the $\mathrm{SE}(3)$ wrench required by \eqref{eq:robot_torque_control} expects also the forces in \lref{TOOL} coordinates which corresponds to a rotation of the position part of the tangent space.
The mean wrench $\bm{\mu}_{\mathrm{VF},i}$ of each fixture $i$ to be commanded at the end effector is thus transformed from $\mathcal{T}^*_{\bm{x}_\mathrm{ee}} \mathcal{M}_1$ to $\mathfrak{se}(3)$ using
\begin{equation}
	\bm{\mu}_{\mathrm{VF},i,\mathrm{SE3}} =
		\begin{bmatrix}
			\bm{R}_{\mathrm{ee}}^\top & \bm{0}\\
			\bm{0} & \bm{I}
		\end{bmatrix}
		\bm{\mu}_{\mathrm{VF},i,\mathcal{M}_1},
\end{equation}
where $\bm{R}_{\mathrm{ee}}$ is the rotation of $\bm{x}_{\mathrm{ee}}$ in the fixture's coordinate system.
The covariance has to be rotated as well using
\begin{equation}
	\bm{\Sigma}_{\mathrm{VF},i,\mathrm{SE3}} = \begin{bmatrix}
			\bm{R}_{\mathrm{ee}}^\top & \bm{0}\\
			\bm{0} & \bm{I}
		\end{bmatrix}
		\bm{\Sigma}_{\mathrm{VF},i,\mathcal{M}_1}
		\begin{bmatrix}
			\bm{R}_{\mathrm{ee}} & \bm{0}\\
			\bm{0} & \bm{I}
		\end{bmatrix}.
\end{equation}
$\bm{\mu}_{\mathrm{VF},i,\mathrm{SE3}}$ and $\bm{\Sigma}_{\mathrm{VF},i,\mathrm{SE3}}$ are then used as experts in a \gls{poe} (cf. \cite{muehlbauer2024probabilistic}, we drop the subscript ${}_{\mathrm{SE3}}$ for readability) to calculate the arbitrated $\hat{\bm{w}}$ as result of the optimization
\begin{equation}
	\hat{\bm{w}} = \mathrm{arg}~\underset{\bm{w}}{\mathrm{min}} \sum_{i=1}^{N_\mathrm{VF}} \left( \bm{w} - \bm{\mu}_{\mathrm{VF},i} \right)^\top \bm{\Sigma}^{-1}_{\mathrm{VF},i} \left( \bm{w} - \bm{\mu}_{\mathrm{VF},i} \right),\nonumber\label{eq:poe_wrench_optimization}
\end{equation}
solved as product of $N_\mathrm{VF} = N_\mathrm{DS} + N_\mathrm{PB} + N_\mathrm{VS}$ Gaussians
\begin{equation}
	\hat{\bm{w}} = \hat{\bm{\Sigma}}_{\mathrm{VF}} \sum_{i=1}^{N_\mathrm{VF}} \bm{\Sigma}^{-1}_{\mathrm{VF},i} \bm{\mu}_{\mathrm{VF},i}, \quad \hat{\bm{\Sigma}}_{\mathrm{VF}} = \left( \sum_{i=1}^{N_\mathrm{VF}} \bm{\Sigma}^{-1}_{\mathrm{VF},i} \right)^{-1}\hspace{-1em}.
	\label{eq:vf_covariance_arbitration}
\end{equation}
The resulting $\hat{\bm{w}}$ with the optimally weighted wrenches\footnote{We assume independent and calibrated uncertainties in the fusion. \gls{kmp} covariances align with \gls{gmm} by construction \eqref{eq:background:kmp_pred_2}. Simulated visual detections are set empirically to match the scale of learned ones while covariances of real detections are typically scaled appropriately by design \cite{muehlbauer2024probabilistic,kuehnoel2024visual}.} of the individual fixtures is then applied to the robot through \eqref{eq:robot_torque_control}.

\subsection{Variable Impedance Control}
\label{sec:problem:variable_impedance}
By modulating the controller stiffness, as a form of authority allocation, we assign a higher importance to the fixture in case of low uncertainty, and give the operator more freedom otherwise.
Using a full covariance matrix, our fixtures can also express \gls{dof} specific and coupled uncertainties.
To reproduce these properties in the robot's impedance behavior, we propose a method to match the stiffness characteristics to those of the covariance matrix.
Unlike previous works \cite{balachandran2023passive,abifarraj2017learning}, we create a full stiffness matrix with nonzero coupling terms $\bm{K}_{tr}$
\begin{equation}
	\bm{K} =
		\begin{bmatrix}
			\bm{K}_t & \bm{K}_{tr}\\
			\bm{K}_{tr}^\top & \bm{K}_r
		\end{bmatrix}.
\end{equation}
A first approach could compute $\bm{K}_i = k \bm{\Sigma}^{-1}_{\mathrm{VF},i} = k \bm{P}_{\mathrm{VF},i}$ using $k$  to scale the precision matrix.
This is, however, too naive, as maximum stiffness values in a robotic system are not respected.
Furthermore, vastly different scales of stiffness values for translation (e.g., \SI{2000}{\newton\per\metre}) and rotation (e.g., \SI{50}{\newton\metre\per\radian}) are neglected.

We therefore propose to decompose the precision matrix $\bm{P}_{\mathrm{VF},i}$ and reassemble it to a stiffness matrix while preserving its anisotropy to the extent possible.
Building on the findings of \cite{chen2015principal,huang2015comments} on the decomposition of spatial stiffness matrices, we decompose a rotated precision matrix $\bm{P}_{\mathrm{VF},i}'$ into translational ($\bm{P}_{t,\mathrm{VF},i}$) and rotational ($\bm{P}_{r,\mathrm{VF},i}$) components
\begin{align}
	\bm{P}_{\mathrm{VF},i}' =
		\begin{bmatrix}
			\bm{A} & \bm{B}\\
			\bm{B}^\top & \bm{C}
		\end{bmatrix} =
		\bm{P}_{t,\mathrm{VF},i} + \bm{P}_{r,\mathrm{VF},i} =\nonumber\\
		\begin{bmatrix}
			\bm{A} & \bm{B}\\
			\bm{B}^\top & \bm{B}^\top\bm{A}^{-1}\bm{B}
		\end{bmatrix} +
		\begin{bmatrix}
			\bm{0} & \bm{0}\\
			\bm{0}´ & \bm{C} - \bm{B}^\top\bm{A}^{-1}\bm{B}
		\end{bmatrix}.
\end{align}
where $\bm{P}_{\mathrm{VF},i}$ is rotated with $\bm{R}_{\mathrm{diag},i}$ to obtain
\begin{equation}
	\bm{P}_{\mathrm{VF},i}' = \begin{bmatrix}
		\bm{R}_{\mathrm{diag},i} & \bm{0}\\
		\bm{0} & \bm{R}_{\mathrm{diag},i}
	\end{bmatrix}^\top
	\bm{P}_{\mathrm{VF},i}
	\begin{bmatrix}
		\bm{R}_{\mathrm{diag},i} & \bm{0}\\
		\bm{0} & \bm{R}_{\mathrm{diag},i}
	\end{bmatrix}
\end{equation}
and $\bm{R}_{\mathrm{diag},i}$ is chosen s.t. $\bm{A}$ is diagonal.

$\bm{P}_{t,\mathrm{VF},i}$ and $\bm{P}_{r,\mathrm{VF},i}$ can be further decomposed and after a scaling be realized using three screw springs and three rotational springs, respectively \cite{chen2015principal}.
An eigendecomposition of $\bm{P}_{r,\mathrm{VF},i}$ yields the torsional spring axes ($j = 4, 5, 6$) as eigenvectors $\bm{w}_j = (0, 0, 0, w_{\mathrm{rx},j}, w_{\mathrm{ry},j}, w_{\mathrm{rz},j})^\top$ and corresponding eigenvalues $\lambda_j$.
Those eigenvalues allow to compute a scaling
\begin{equation}
	\label{eq:variable_stiffness:torque_scaling}
	s_j =
		\begin{cases}
			0,& \lambda_j < \lambda_\mathrm{rot}^-\\
			\frac{\lambda_j - \lambda_\mathrm{rot}^-}{\lambda_\mathrm{rot}^+ - \lambda_\mathrm{rot}^-},& \lambda_\mathrm{rot}^- < \lambda_j < \lambda_\mathrm{rot}^+\\
			1,& \lambda_j \ge \lambda_\mathrm{rot}^+
		\end{cases},
\end{equation}
where $\lambda_\mathrm{rot}^-$ and $\lambda_\mathrm{rot}^+$ are empirically determined hyperparameters for ``low'' and ``high'' eigenvalues resulting in full and zero stiffness.
The torsional springs are then realized as ($j = 4, 5, 6$)
\begin{equation}
	\label{eq:variable_stiffness:torque_realization}
	\bm{K}_j' = k_{\mathrm{nom},j} s_j \bm{w}_j \bm{w}_j^\top
\end{equation}
with the nominal stiffnesses ${\bm{k}_\mathrm{nom} = (k_t, k_t, k_t, k_r, k_r, k_r)^\top}$ where $k_t$ is the translational and $k_r$ the rotational stiffness.
Eigenvalues $\ge \lambda_\mathrm{rot}^+$ result in a nominal stiffness along the corresponding rotational \gls{dof} while eigenvalues $< \lambda_\mathrm{rot}^-$ result in zero stiffness with linear scaling in between.

The screw springs ($j = 1, 2, 3$) are given by wrench axes $\bm{w}_j = (e_j^\top, w_{\mathrm{tx},j}, w_{\mathrm{ty},j}, w_{\mathrm{tz},j})^\top$ with $\begin{bmatrix} e_1 & e_2 & e_3 \end{bmatrix} = \bm{I}_3$ and corresponding $\lambda_j$ from the diagonal entries of $\bm{A}$ in $\bm{P}_t$.
The scaling $s_j$ of \eqref{eq:variable_stiffness:torque_scaling} is also computed with factors $\lambda_\mathrm{trans}^-$ and $\lambda_\mathrm{trans}^+$.
We limit the rotational stiffness of $w_j$ by ensuring
\begin{equation}
	s_j \le \frac{k_r}{k_{\mathrm{nom},j}\left(w_{\mathrm{tx},j}^2 + w_{\mathrm{ty},j}^2 + w_{\mathrm{tz},j}^2\right)}.
	\label{eq:variable_stiffness:limit_rot}
\end{equation}
While this operation impacts the coupling of translations and rotations in the screw spring, it ensures attainable stiffness values close to the characteristics of the original precision matrix $\bm{P}_{\mathrm{VF},i}$.
The wrench springs are then realized to ($j = 1, 2, 3$)
\begin{equation}
	\label{eq:variable_stiffness:wrench_realization}
	\bm{K}_j' = k_{\mathrm{nom},j} s_j \bm{w}_j \bm{w}_j^\top.
\end{equation}
The resulting stiffness matrix is then calculated from wrench ($j = 1, 2, 3$) and torsional ($j = 4, 5, 6$) springs and rotated
\begin{equation}
	\bm{K}_i = \begin{bmatrix}
		\bm{R}_{\mathrm{diag},i} & \bm{0}\\
		\bm{0} & \bm{R}_{\mathrm{diag},i}
	\end{bmatrix}
	\left(\sum_{j=1}^6 \bm{K}_j'\right)
	\begin{bmatrix}
		\bm{R}_{\mathrm{diag},i} & \bm{0}\\
		\bm{0} & \bm{R}_{\mathrm{diag},i}
	\end{bmatrix}^\top\hspace{-1em}.\nonumber
\end{equation}
$\bm{K}_i$ is then used for impedance control \eqref{eq:VF_wrench} of the $i$-th fixture.

In \Cref{sec:evaluation:coupled_stiffness}, we show how this approach couples translational and rotational \glspl{dof} in the stiffness matrix.
See \Cref{sec:appendix:var_stiffness_rot} for the extension to manifolds $\mathcal{M}_2$ and $\mathcal{M}_3$ and \Cref{sec:appendix:damping} for an optimal damping calculation.
We passivate variable stiffness and fusion \eqref{eq:poe_wrench_optimization} using \cite{muehlbauer2026stabilizing} as \textit{offline} methods such as \cite{kronander2016stability} are not suited to our \textit{online} stiffness calculation.

\section[Probabilistic Dynamical System Virtual Fixtures]{Probabilistic Dynamical System \Acrlongpl{vf}}
\label{sec:dynamic_fixtures}
\begin{figure}
	\centering
	\includegraphics[width=.297\columnwidth,trim={29 20 29 40},clip]{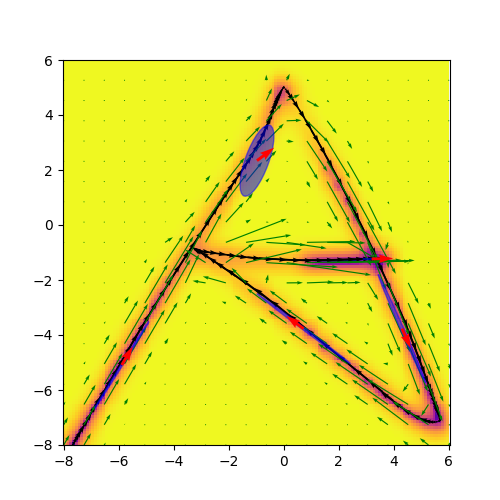}
	\includegraphics[width=.297\columnwidth,trim={29 20 29 40},clip]{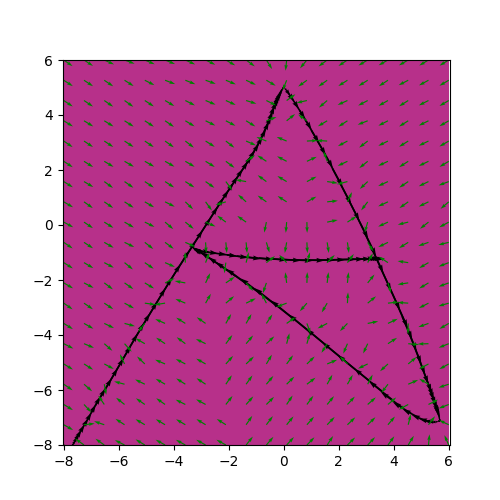}
	\includegraphics[width=.376\columnwidth,trim={32 20 32 40},clip]{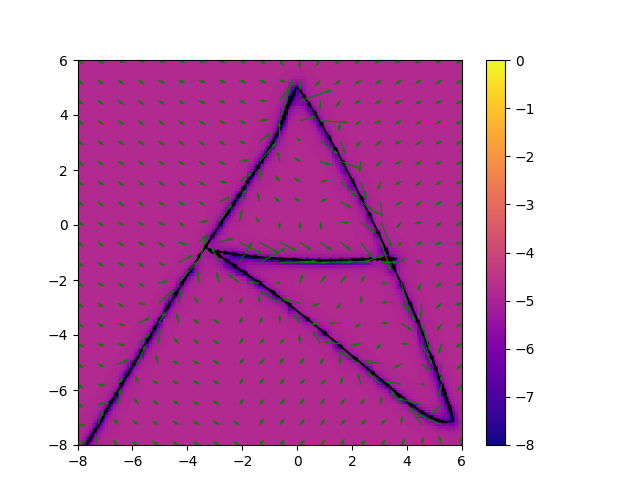}
	\caption{\label{fig:dyn_fixture:kmp_policy} 2D motion policy using a \gls{kmp}. Black arrows visualize velocities in the training data \cite{calinon2017learning}. The \gls{kmp} output is shown on the left (\Cref{sec:non_parametric_DS}), blue ellipsoids and red arrows depict centers, position covariances and velocities of the Gaussians in the reference \gls{gmm}.
The \textit{stabilizing policy} (\Cref{sec:dynamic_fixtures:base_policy}) is shown in the middle, final velocities resulting from the arbitration on the right.
The colormap of each plot depicts $\mathrm{log}(\mathrm{det}(\bm{\Sigma}))$.}
\end{figure}
We introduce automated \Acrlong{ds} based \Acrlongpl{vf} for \textbf{coarse guidance} modeling the relation between assistive velocities $\dot{\bm{x}}_\mathrm{DS}$ and robot state $\bm{x}_\mathrm{ee}$ probabilistically
\begin{equation}
	p(\dot{\bm{x}}_\mathrm{DS} | \bm{x}_\mathrm{ee}) = \mathcal{N}(\dot{\bm{x}}_\mathrm{DS} | \bm{\mu}_\mathrm{DS}, \bm{\Sigma}_\mathrm{DS}),\label{eq:dynsys_start}
\end{equation}
where $\bm{\mu}_\mathrm{DS}$ denotes the mean of the dynamical system evaluated at $\bm{x}_\mathrm{ee}$ and $\bm{\Sigma}_\mathrm{DS}$ the associated covariance.
In contrast to time-driven motions, these models encode state-dependent dynamic behaviors that support user progression along the task, adapting their guidance with varying granularity, such as modulating velocity magnitude based on the task phase or enabling periodic motions.
This distinguishes \gls{ds}-based fixtures from the trajectory-based fixtures (\Cref{sec:auto_position_based}), which can only encode a trajectory with start and end point.
By providing motion information for the whole robot workspace, they are especially suited to the approaching phase of a manipulation where the robot may start at arbitrary configurations.

We adopt an uncertainty-aware, non-parametric approach that is highly flexible to composition and modulation without requiring parameter re-computation.
Based on demonstration data, we use \glspl{kmp} (\Cref{sec:background:lfd}) to encode a specific task in a region of the workspace which results in $N_\mathrm{DS}$ concurrently active velocity fields on possibly different manifolds.
In the following, $p_n\left(\dot{\bm{x}}_\mathrm{DS}\right)$ denotes the $n$-th of $N_\mathrm{DS}-1$ probabilistic \gls{ds} models learned from demonstrations (see \Cref{sec:non_parametric_DS}), and $p_\mathrm{stab}(\dot{\bm{x}}_\mathrm{DS})$ a probabilistic policy (see \Cref{sec:dynamic_fixtures:base_policy}) with a constant, pre-defined uncertainty that drives the robot toward the closest known dynamic.
A proportional control scheme (\Cref{sec:dynamic_fixtures:vf}) enables the use of such learned \gls{ds} as a probabilistic \gls{vf}.
Through our arbitration scheme (\Cref{sec:problem:arbitration}), we leverage the epistemic uncertainty encoded by \glspl{kmp} to automatically prioritize models with low uncertainty.
When all models exhibit high uncertainty, the formulation defaults to the stabilizing policy $p_\mathrm{stab}(\dot{\bm{x}})$ or to another type of fixture, such as those introduced in \Cref{sec:auto_position_based,sec:vs_fixture}.

\subsection{Non-parametric Learning of Dynamical Systems}
\label{sec:non_parametric_DS}
We define the $n$-th \gls{ds} as a probabilistic mapping
\begin{equation}
	\label{eq:gp_approximate}
	p_n(\dot{\bm{x}}_\mathrm{DS} | \bm{x}_\mathrm{ee}) = \mathcal{N}(\dot{\bm{x}}_\mathrm{DS}|\bm{\mu}_{\mathrm{DS},n},\bm{\Sigma}_{\mathrm{DS},n})
\end{equation}
which can be learned from a dataset of demonstrations $\{\bm{x}_j, \dot{\bm{x}}_j\}_{j=1}^N$ where, e.g., a full pose $\bm{x} \in \mathbf{R}^3 \times \mathcal{S}^3$ ($\mathcal{M}_1$) or only the position part $\bm{x} \in \mathbf{R}^3$ with velocities in the tangent spaces $\dot{\bm{x}} \in \mathcal{T}_{\bm{x}} \mathcal{M}$ respectively $\dot{\bm{x}} \in \mathcal{T}_{\bm{x}} \mathbb{R}^3 = \mathbb{R}^3$ are used.
This dataset is subsampled with equal spacing of the input poses to achieve a trade-off between accuracy and computational cost.
The demonstrations are then used to learn a \gls{kmp} (\Cref{sec:background:lfd}).
For the use in a \gls{ds}, we customize the computation of the reference distribution to obtain covariances in pose space as detailed in \Cref{sec:appendix:covariance_pose_space}.

In our experiments, we use the on-manifold \gls{rbf} kernel with the distance function \eqref{eq:distance_function_onman_weighted}
\begin{equation}
	k(\bm{x}, \bm{x}_\mathrm{ee}) = \mathrm{exp}\left( -\frac{d_{\bm{I}}^{\mathcal{M}}(\bm{x}, \bm{x}_\mathrm{ee})}{2 l^2} \right),\label{eq:dynamic_fixtures:rbf}
\end{equation}
with length scale $l$.
Similarly to \glspl{gp}, we assume a zero-mean prior, so predictions from \eqref{eq:background:kmp_pred_1} yield $\dot{\bm{x}}_\mathrm{DS} = \bm{0}$ in regions far from the demonstrations. Moreover, \eqref{eq:background:kmp_pred_2} captures aleatoric uncertainty near the data and increasing epistemic uncertainty in regions with limited or no demonstrations.
The distance function \eqref{eq:dynamic_fixtures:rbf} naturally extends to the manifold $\mathcal{M}$ chosen for the task.
We treat the output as Euclidean, therefore, only the kernel has to be adapted to the manifold case assuming that the learned velocities are smooth and the length scale parameter of the kernel is small compared to the changes in the velocity.



\subsection{Probabilistic Stabilizing Policy}
\label{sec:dynamic_fixtures:base_policy}
Outside of the task space regions where demonstrations were provided, the robot actions computed from \eqref{eq:gp_approximate} are zero, leaving the robot stationary.
A probabilistic \textit{stabilizing policy}\footnote{This policy is not stabilizing in the sense of ensuring convergence but provides a \textit{soft} stability that brings the robot back to regions of learned \glspl{ds}.}
\begin{equation}
	p_\mathrm{stab}(\dot{\bm{x}}_\mathrm{DS} | \bm{x}_\mathrm{ee}) = \mathcal{N}(\dot{\bm{x}}_\mathrm{DS} | \bm{\mu}_\mathrm{stab}, \bm{\Sigma}_\mathrm{stab})
\end{equation}
then brings the robot back into the demonstrated areas.
To achieve this, we compute the distance to the reference poses of all $N_\mathrm{DS}$ velocity fields using \eqref{eq:distance_function_onman_weighted} to $d_j = d_{\bm{I}}^{\mathcal{M}_1}(\bm{x}_j, \bm{x}_\mathrm{ee})$ and obtain the closest known dynamic pose $\bm{x}_{j^*}$ with $j^* = \operatorname*{argmin}_j d_j$.
The normalized velocity towards this pose using the distance \eqref{eq:distance_function_onman_weighted} and default velocity $\dot{x}_\mathrm{stab}$ evaluates to
\begin{equation}
	\bm{\mu}_\mathrm{stab} = \dot{x}_\mathrm{stab} \frac{\mathrm{Log}_{\bm{x}_\mathrm{ee}}^{\mathcal{M}_1}(\bm{x}_{j^*})}{\sqrt{d_{\bm{I}}^{\mathcal{M}_1}(\bm{x}_\mathrm{ee}, \bm{x}_{j^*})}}
\end{equation}
which is furthermore equipped with a constant covariance $\bm{\Sigma}_\mathrm{stab} = \sigma_\mathrm{stab} \bm{I}_6$.
The covariance $\sigma_\mathrm{stab}$ of this policy is an important hyperparameter -- it has to be chosen such that $\bm{\Sigma}_\mathrm{stab}$ is larger than the covariances $\bm{\Sigma}_{\mathrm{DS},n}$ of the learned \glspl{ds} in the areas where demonstrations have been provided.
Thanks to \glspl{kmp} providing epistemic uncertainty, the covariances $\bm{\Sigma}_{\mathrm{DS},n}$ increase outside of the demonstrated areas, therefore activating the stabilizing policy which brings the robot back inside the demonstrated areas through the arbitration \eqref{eq:vf_covariance_arbitration}.

\Cref{fig:dyn_fixture:kmp_policy} shows our \gls{ds} learned on 2D data \cite{calinon2017learning}.
We use a \gls{rbf} kernel with $l=0.3$, the \gls{kmp} is initialized from a \gls{gmm} with $5$ Gaussians sampled at the input data points and \textit{stabilizing policy} with $\bm{\Sigma} = \mathrm{diag}(0.1, 0.1)$.
We show the advantages of this \gls{kmp}-based policy over \glspl{gp} in \Cref{sec:evaluation:dynamic_vf}.

\subsection[Control Law for Dynamical-Systems-based Virtual Fixtures]{Control Law for DS-based \Acrlongpl{vf}}
\label{sec:dynamic_fixtures:vf}
We use a proportional control law to compute wrenches
\begin{equation}
	\bm{w}_{\mathrm{VF},i} = \bm{D}_{\mathrm{VF},i} \left( \dot{\bm{x}}_{\mathrm{VF},i} - \dot{\bm{x}}_\mathrm{ee} \right) \label{eq:dynamic_fixtures:vel_ctrl}
\end{equation}
with end effector velocity $\dot{\bm{x}}_\mathrm{ee}$ and \gls{vf} velocity $\dot{\bm{x}}_{\mathrm{VF},i}$ ($\bm{\mu}_\mathrm{DS}$ from \eqref{eq:dynsys_start}) in $\mathcal{T}_{\bm{x}_\mathrm{ee}}\mathcal{M}$ and a constant damping $\bm{D}_{\mathrm{VF},i}$.
Note that \eqref{eq:dynamic_fixtures:vel_ctrl} can be derived from \eqref{eq:VF_wrench} by setting $\bm{K}_{\mathrm{VF},i} = \bm{0}$.

The fixture wrench $\bm{w}_{\mathrm{VF},i}$ is expressed in cotangent space $\mathcal{T}^*_{\bm{x}_\mathrm{ee}}\mathcal{M}$.
As detailed in \Cref{sec:problem:arbitration}, $\bm{w}_{\mathrm{VF},i}$ and the associated covariance $\bm{\Sigma}_{\mathrm{VF},i}$ have to be transformed to the (co)tangent space of $\mathbb{R}^3 \times \mathcal{S}^3$ which allows for a natural arbitration between different velocity fixtures and other types of \glspl{vf}.

\section{Position-based Trajectory Fixtures: Learning, Arbitration and Automation}
\label{sec:auto_position_based}
Position-based trajectory \Acrlongpl{vf} for \textbf{precise guidance} compute a probabilistic attractor from the robot state $\bm{x}_\mathrm{ee}$
\begin{equation}
	p(\bm{x}_\mathrm{PB} | \bm{x}_\mathrm{ee}) = \mathcal{N}(\bm{x}_\mathrm{PB} | \bm{\mu}_\mathrm{PB}, \bm{\Sigma}_\mathrm{PB})
\end{equation}
This formulation requires both a probabilistic trajectory and a distance function for mapping the end effector pose $\bm{x}_\mathrm{ee}$ to the closest pose on the trajectory.
To benefit from the different manifolds $\mathcal{M}$ introduced in \Cref{sec:background:on_manifold}, we show the extension of \cite{muehlbauer2024probabilistic} from the manifold $\mathbb{R}^3 \times \mathcal{S}^3$ to $\mathcal{M}$.
Furthermore, we introduce an automation of this fixture in \Cref{sec:auto_position_based:automation}.
The probabilistic wrench of each of $N_\mathrm{PB}$ concurrently active trajectory fixtures calculated through variable impedance control (\Cref{sec:problem:variable_impedance}) is again combined with all other fixtures through the arbitration (\Cref{sec:problem:arbitration}).

\subsection{Learning of Trajectory Fixtures on different Manifolds $\mathcal{M}$}
\label{sec:auto_position_based:learning}
We start from a dataset of trajectories $\{t_i, \bm{x}_i\}_{i=1}^N$ containing demonstrations with time $t \in \mathbb{R}^1$ and pose $\bm{x} \in \mathcal{M}$ for each of the $N_\mathrm{PB}$ fixtures.
\Gls{dtw} \cite{salvador2007toward} then allows us to align the individual demonstrations in $t \in [0, 1]$ to encode them in a \gls{gmm} \eqref{eq:background:probabilistic:joint_gmm_encoding} with $M$ components.
To benefit from the properties of different manifolds $\mathcal{M}$, we use the square root of the distance function defined in \eqref{eq:distance_function_onman_weighted} with $\bm{A} = \bm{I}$ as distance measure between the two poses $\bm{x}_j$ and $\bm{x}_k$ in the \gls{dtw} calculation.
This for example allows us to learn trajectories in cylindrical coordinates $\mathcal{M}_2$, see \Cref{sec:evaluation:auto_pb_robograv} for an evaluation.
The effect of this distance measure for Cartesian and cylindrical coordinate systems is shown in \Cref{fig:pb_fixture:distance_manifolds}.
The result of \gls{dtw} is comparable to the introduction of a phase variable \cite{ti2023geometric}.
Using \gls{gmr}, we condition the \gls{gmm} on time $t$ to obtain probabilistic poses $p(\bm{x} \vert t)$ \eqref{eq:background:probabilistic:gmm_multimodal}.

\subsection{On-Manifold Attractor Point Calculation}
\label{sec:auto_position_based:attractor_point}
For impedance control, a single attractor point as in \eqref{eq:background:probabilistic:gmm_unimodal} is required.
In the Euclidean case, a closed form solution exists for extracting this point on a time-based probabilistic trajectory \cite{raiola2017comanipulation}.
As on-manifold Gaussian operations require iterations with a variable number of steps \cite{zeestraten2017manifold}, they are not well suited for real-time control.
We therefore extract a trajectory from the conditional distribution $p(\bm{x} | t)$ by calculating $\{\bm{\mu}_n, \bm{\Sigma}_n\}_{n=1}^N$ for $N$ equally spaced samples of $t \in [0, 1]$.
This trajectory is then sent to the real-time controller for on-manifold interpolation of the attractor $\bm{x}_\mathrm{PB}$ as detailed in \Cref{sec:appendix:on_manifold_interpolation} and variable stiffness control (\Cref{sec:problem:variable_impedance}) to calculate $\bm{w}_{\mathrm{VF},i}$.

\subsection{Distance-based Covariance Adaptation}
\label{sec:auto_position_based:cov_adaptation}
The covariance $\bm{\Sigma}_\mathrm{PB}$ is used for variable impedance control (\Cref{sec:problem:variable_impedance}) and for arbitration with other fixtures (\ref{sec:problem:arbitration}).
By default, the covariance matrix only depends on the closest point of the trajectory, thus modeling aleatoric uncertainty.
This results in large wrenches $\bm{w}_{\mathrm{VF},i}$ when the end effector $\bm{x}_\mathrm{ee}$ is far away from the fixture even though the robot no longer follows it.
We therefore propose to adapt the original covariance output of the fixture by a distance-based scaling of the precision matrix $\bm{P}_\mathrm{PB} = \bm{\Sigma}_\mathrm{PB}^{-1}$ to $\hat{\bm{P}}_\mathrm{PB} = s \cdot \bm{P}_\mathrm{PB}$ using
\begin{align}
	s =
	\begin{cases}
		1, & d < d_\mathrm{min}\\
		1 - \frac{d_{\bm{P}_\mathrm{PB}}^{\mathcal{M}}(\bm{x}_\mathrm{PB}, \bm{x}_\mathrm{ee}) - d_\mathrm{min}}{d_\mathrm{max} - d_\mathrm{min}}, & d_\mathrm{min} \le d \le d_\mathrm{max}\\
		0, & d > d_\mathrm{max}
	\end{cases}
\end{align}
with the Mahalanobis distance $d^{\bm{P}_\mathrm{PB}}_{\mathcal{M}}(\bm{x}_\mathrm{PB}, \bm{x}_\mathrm{ee})$ between end effector $\bm{x}_\mathrm{ee}$ and attractor $\bm{x}_\mathrm{PB}$ weighted by the precision matrix $\bm{P}_\mathrm{PB}$.
The parameters $d_\mathrm{min}$ and $d_\mathrm{max}$ determine the distances at which the precision matrix scaling starts and ends, respectively.
Using the Mahalanobis distance takes the influence of specific \glspl{dof} into account, thus leaving the fixture active farther in directions with high uncertainty.
This is relevant to exploit the variable stiffness formulation (\Cref{sec:problem:variable_impedance}) where those directions get much lower stiffnesses, thus allowing the operator to move more freely.
The resulting behaviour is showcased in \Cref{sec:evaluation:all_fixtures_sara}.

\subsection{Automating the Fixture}
\label{sec:auto_position_based:automation}
As the position-based fixture is derived from a time-driven motion, we can compute a preferred direction along the trajectory.
Inspired by \cite{pezzementi2007dynamic}, we calculate the current direction
\begin{equation}
	\bm{\delta} = \bm{\mu}_{j+1,\mathrm{pos}} - \bm{\mu}_{j,\mathrm{pos}}
\end{equation}
where $\bm{\mu}_{j,pos}$ and $\bm{\mu}_{j+1,pos}$ are the $\mathbb{R}^3$ positions of the interpolation points of the fixture.
Subsequently, we normalize $\bm{\delta}$ to $\tilde{\bm{\delta}} = \frac{\bm{\delta}}{||\bm{\delta}||}$.
The same velocity controller as employed for \gls{ds} based fixtures (\Cref{sec:dynamic_fixtures:vf}) is used to calculate an automation wrench $\bm{w}_{\mathrm{VF,aut},i}$ from $\tilde{\bm{\delta}}$.
The resulting wrench is then added to the wrench of the impedance controller \eqref{eq:VF_wrench}
\begin{equation}
	\tilde{\bm{w}}_{\mathrm{VF},i} = \bm{w}_{\mathrm{VF},i} + \bm{w}_{\mathrm{VF,aut},i}.
\end{equation}

\section{Geometric Visual Servoing Fixture}
\label{sec:vs_fixture}
Probabilistic visual servoing \glspl{vf} introduced in \cite{muehlbauer2024probabilistic} for \textbf{very precise guidance} model the attractor based on visual input $\mathrm{I}$
\begin{equation}
	p(\bm{x}_\mathrm{VS} | \mathrm{I}) = \mathcal{N}\left(\bm{x}_\mathrm{VS} | \bm{\mu}_\mathrm{VS}, \bm{\Sigma}_\mathrm{VS}\right).
\end{equation}
We consider $N_\mathrm{VS}$ concurrently active visual servoing \glspl{vf}, each modeling $M$ individual fixtures, one for every probabilistic visual detection $p_m(\bm{x}_\mathrm{VS}|\bm{x}_\mathrm{ee})=\mathcal{N}\left(\bm{x}_\mathrm{VS}|\bm{\mu}_m,\bm{\Sigma}_m\right)$, depending on $\bm{x}_\mathrm{ee}$.
We assume that the uncertainty given by the visual detection incorporates uncertainties resulting from, e.g., ambiguous or occluded detections.
A \gls{moe} model \cite{jacobs1991adaptive, bishop2006pattern} allows us to model the fixture as a multi-modal distribution
\begin{equation}
	p(\bm{x}_{\mathrm{VS}}|\bm{x}_{\mathrm{ee}}) = \sum_{m=1}^M \hat{h}_m(\bm{x}_{\mathrm{ee}},\bm{\mu}_m) p_m(\bm{x}_{\mathrm{VS}}|\bm{x}_{\mathrm{ee}}).
	\label{eq:MoE}
\end{equation}
Variable impedance control (\Cref{sec:problem:variable_impedance}) is used to calculate a wrench for each of the $N_\mathrm{VS}$ fixtures which is then fused with all other fixtures through the arbitration (\Cref{sec:problem:arbitration}).

\begin{figure*}
	\centering
	\begin{subfigure}{0.32\textwidth}
		\centering
		\annotategraphicsmulti{
		    \includegraphics[width=\columnwidth,trim={200 250 200 176},clip]{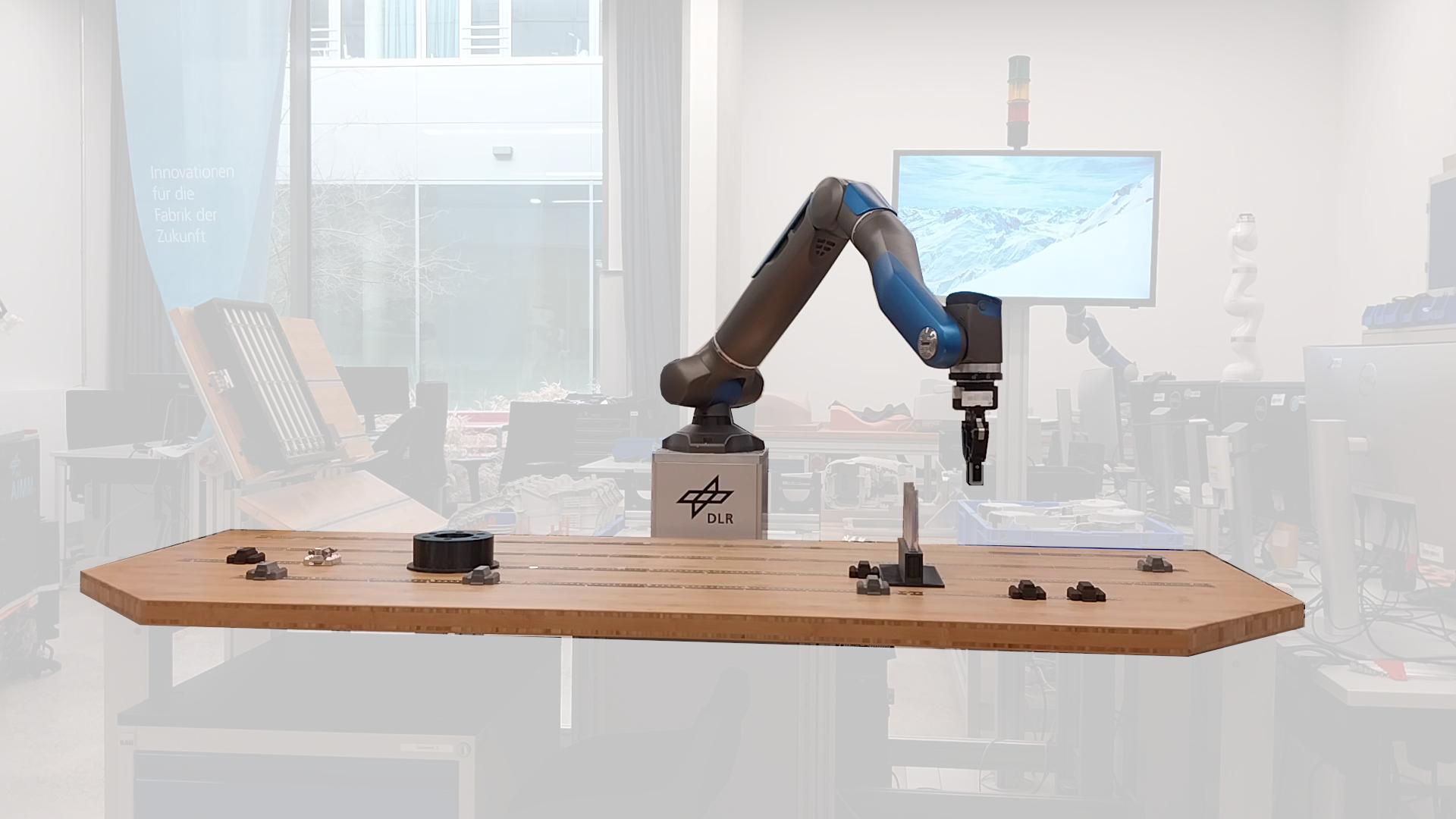}
		}{
	    }
    \end{subfigure}
	\begin{subfigure}{0.32\textwidth}
		\centering
		\annotategraphicsmulti{
		    \includegraphics[width=\columnwidth,trim={250 306 150 120},clip]{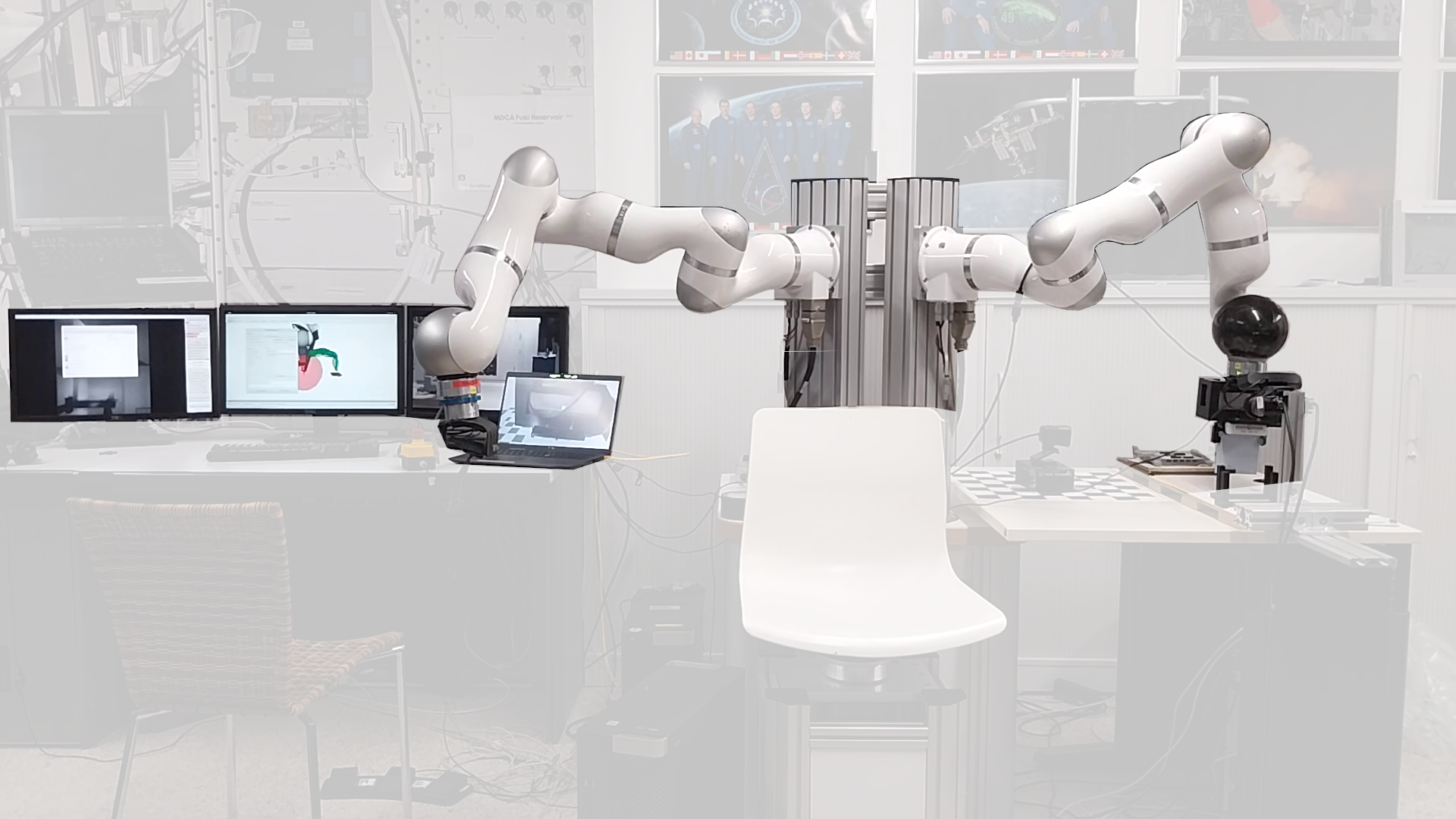}
		}{
	    }
    \end{subfigure}
	\begin{subfigure}{0.32\textwidth}
		\centering
		\annotategraphicsmulti{
		    \includegraphics[width=\columnwidth,trim={540 222 540 498},clip]{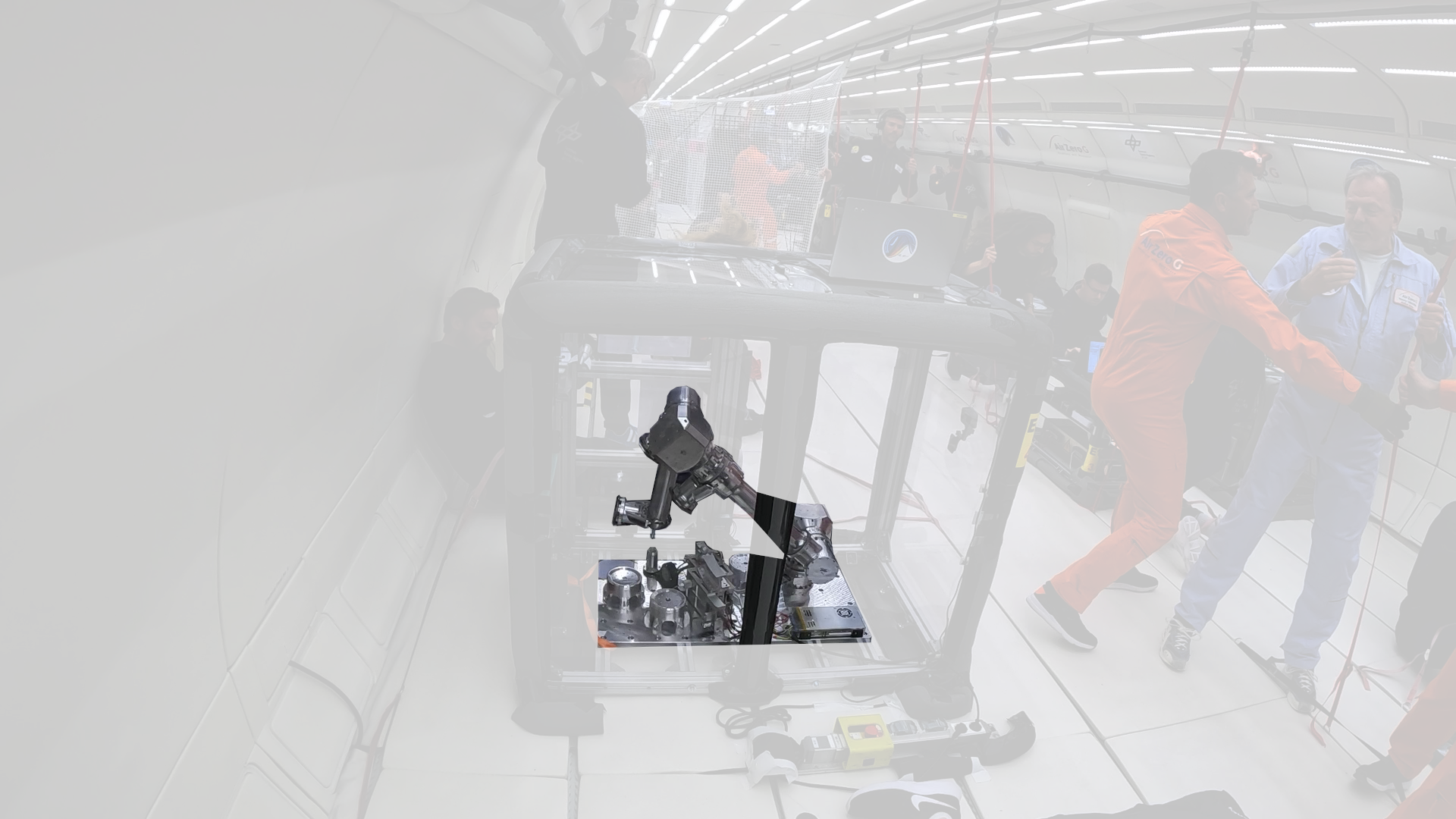}
		}{
	    }
    \end{subfigure}
	\caption{\label{fig:evaluation:robotic_systems} The \textit{DLR SARA robot} (\textbf{left}) is used in \textit{hand-guided} mode for evaluating individual fixtures in \Cref{sec:evaluation:coupled_stiffness,sec:evaluation:position_stiffness,sec:evaluation:dynamic_vf}, as well as their combinations in  \Cref{sec:evaluation:dynamic_vs_hug,sec:evaluation:all_fixtures_sara}.
	The \textit{HUG robot} with two torque-controlled 7-\gls{dof} manipulators (\textbf{center}) is used for evaluating the fixtures in a \textit{teleoperation} task in \Cref{sec:evaluation:dynamic_vs_hug}.
	The 4-\gls{dof} \textit{RoboGrav robot arm} (\textbf{right}) is used with \textit{fully automated} fixtures (\Cref{sec:evaluation:dynamic_vf,sec:evaluation:auto_pb_robograv,sec:evaluation:dynamic_pb_robograv}).
	}
\end{figure*}

\subsection{Geometry-aware Mixture of Experts Gating Functions}
In \cite{muehlbauer2022mixture}, the \textit{gating function} $h_m$ computes the influence of each expert based on the distance between expert and end effector $\mathcal{M}_1$.
Using the geometry-aware distance function \eqref{eq:distance_function_onman_weighted}, we can generalize to manifolds $\mathcal{M}_i$ defined in \Cref{sec:background:on_manifold}
\begin{align}
	h_m(\bm{x}_{\mathrm{ee}},\bm{\mu}_m) = \mathrm{exp}\left(\!-\frac{1}{2} d_{\bm{L}}^{\mathcal{M}}(\bm{x}_\mathrm{ee}, \bm{\mu}_m) \right)\!+\!\gamma
\label{eq:gating_function}
\end{align}
where $\gamma$ is a regularization factor and the hyperparameter $\bm{L} = \mathrm{diag}(l^2_0, l^2_1, l^2_2, l^2_{wx}, l^2_{wy}, l^2_{wz})^{-1}$ specifies the relevance of each direction.
For $\mathcal{M}_1$, $l_0$, $l_1$ and $l_2$ correspond to $x$, $y$ and $z$.
For $\mathcal{M}_2$, $l_0$ and $l_1$ scale angular \glspl{dof} and radius $r$.
In case of $\mathcal{M}_3$, $l_0$ and $l_1$ weigh the two angular \glspl{dof} and $l_2$ radius $r$.
When far from all detections, $\gamma$ assigns equal weights to each expert, reflecting the overall uncertainty of all detections.

Similarly to \cite{muehlbauer2024probabilistic}, we compute a unimodal distribution of \eqref{eq:MoE} via moment matching resulting in mean $\bm{\mu}_\mathrm{VS}$ and covariance $\bm{\Sigma}_\mathrm{VS}$, used as attractor in \eqref{eq:VF_wrench} with variable stiffness (\Cref{sec:problem:variable_impedance}).
The fixture wrench is again combined with all other fixture wrenches through the arbitration (\Cref{sec:problem:arbitration}).

\subsection{Geometric Expert Customization}
The expert customization of \cite{muehlbauer2024probabilistic} needs flexibilisation and formalisation.

\subsubsection{Zero force along insertion axis}
We assume that the insertion that should be controlled by the operator is to be performed along one \gls{dof} of the chosen manifold.
Therefore, the stiffness in the corresponding row and column of $\bm{K}$ is set to zero to not generate any forces along this axis and allow the operator full control over the insertion.

\subsubsection{Deadzones}
In the vicinity of a connector, the operator should receive strong guidance.
This can be achieved by modifying $\mathrm{Log}_{\bm{x}_{ee}}^{\mathcal{M}}\!\!\left( \bm{\mu}_m \right)$, setting its entries to zero for distances smaller than a predefined radius and scaling it for larger distances.
We use a length vector $\bm{l}_\mathrm{dead}$ to deform the difference vector as well as the scalar deadzone value $r_\mathrm{dead}$ to calculate the modified logarithm $\mathrm{Log}_{\bm{x}_{ee}}^{'\mathcal{M}}\!\!\left( \bm{\mu}_m \right)$ (\Cref{sec:appendix:modified_log}).
The modified value is then used in the gating function \eqref{eq:gating_function}.

\subsubsection{Expert Initialization}
As in \cite{muehlbauer2024probabilistic}, we initialize the \gls{moe} with an additional expert at the end effector pose with high covariance.
This additional expert ensures that the fixture does not generate forces outside its valid region.
It is parameterized with length scale $\bm{l}_\mathrm{dead,add}$ and dead zone $r_\mathrm{dead,add}$, calculating the modified difference $\mathrm{Log}_{\bm{x}_{ee}}^{'\mathcal{M}}\!\!\left( \bm{\mu}_m \right)$ (\Cref{sec:appendix:modified_log}).
Its influence factor $h_{M+1}$ is then calculated as in \cite{muehlbauer2024probabilistic}
\begin{multline}
	h_{M+1}(\bm{x}_\mathrm{ee}, \bm{x}_\mathrm{targ}) =
	1 - \mathrm{exp} \left( - \frac{1}{2} d_{\bm{L}_\mathrm{add}}^{'\mathcal{M}}\left(\bm{x}_\mathrm{ee}, \bm{\mu}_m \right) \right),
\end{multline}
where $\bm{L}_\mathrm{add} = \mathrm{diag}(\bm{l}_\mathrm{add})$ is the length vector for the additional expert, $\bm{x}_\mathrm{targ}$ the expected mean of the experts and the modified $\mathrm{Log}_{\bm{x}_{ee}}^{'\mathcal{M}}\!\!\left( \bm{\mu}_m \right)$ is used in $d_{\bm{L}_\mathrm{add}}^{'\mathcal{M}}\left(\bm{x}_\mathrm{ee}, \bm{\mu}_m \right)$.


\section{Evaluation}
\label{sec:evaluation}
Our framework is implemented on three robotic systems (\Cref{fig:evaluation:robotic_systems}) in different \textit{automation levels} and \textit{interaction modes} for evaluating task-specific metrics (\Cref{tab:experiment_classifications}).
The \glspl{vf} are implemented on standard computers using Simulink and C++ code running in hard real time at the robot's control rate of up to \SI{8}{\kilo\hertz}.
As first system (left image), we use the DLR SARA robot in \textit{hand-guided mode}, i.e., an operator directly interacts with the arm, representing human-robot collaboration in a factory context.
On this robot, we evaluate components of our framework and use cases in all automation levels.
To emulate a space application, we use the HUG robot with two torque-controlled 7-\gls{dof} manipulators (middle image) where one robot arm is used as haptic input device and the other as remote robot to test the fixtures in a \textit{teleoperation} setup in \textit{semi-automated} mode.
Finally, we use the 4-\gls{dof} space-ready RoboGrav setup (right image) \textit{without} interaction in \textit{fully automated} mode.
Implementation on a space-grade computer showcases applicability of our method on space hardware.
This setup is designed to be tested in outer space; experiments in this paper were conducted on the 42nd DLR parabolic flight experiment in June 2024.

We start with an evaluation of our novel variable stiffness formulation (\Cref{sec:problem:variable_impedance}).
We demonstrate coupled translational and rotational \glspl{dof} in \Cref{sec:evaluation:coupled_stiffness} and low stiffness along \glspl{dof} of $\mathcal{M}_3$ in \Cref{sec:evaluation:position_stiffness}.
Next, in \Cref{sec:evaluation:dynamic_vf}, we compare our novel \gls{ds}-based \gls{vf} with a \gls{gp} baseline, \textit{quantifying} how much repeated executions differ in their trajectories.
Finally, we \textit{quantify} the repeatability of the automated position-based fixture (\Cref{sec:auto_position_based}) over multiple fully autonomous executions in \Cref{sec:evaluation:auto_pb_robograv}.

Using the arbitration (\Cref{sec:problem:arbitration}), we combine multiple fixtures.
We start with \gls{ds}- and position-based \glspl{vf} in \textit{fully automated} execution (\Cref{sec:evaluation:dynamic_pb_robograv}).
Next, we combine \gls{ds}-based and visual servoing \glspl{vf} to automate where possible with human interaction where required (\Cref{sec:evaluation:dynamic_vs_hug}).
Finally, we combine all fixture types (\Cref{sec:evaluation:all_fixtures_sara}).

To obtain user feedback for our approach\footnote{Ethical approval obtained from the ``Geschäftsstelle Forschungsethik'' of the German Aerospace Center (DLR) under number 15/24.}, we have asked six expert users ($2$ female, $4$ male) aged 27-42 (\SI{35.2 \pm 5.6}{\year}) with experience in shared control methods and haptic interfaces who were not involved in this work to perform experiments in \Cref{sec:evaluation:coupled_stiffness,sec:evaluation:dynamic_vs_hug,sec:evaluation:all_fixtures_sara}.%
\footnote{The experiment with arbitration in \Cref{sec:evaluation:all_fixtures_sara} has been conducted for a first version of the manuscript, while the experiments in \Cref{sec:evaluation:coupled_stiffness,sec:evaluation:dynamic_vs_hug} and the remaining case without arbitration for \Cref{sec:evaluation:all_fixtures_sara} (in this order) were conducted for a second version with more thorough comparisons. Through extended familiarization with the system, this protocol leads to a bias favoring the baseline of no arbitration in \ref{sec:evaluation:all_fixtures_sara}; the other experiments are counterbalanced. The duration of experiments was \SI{20}{\minute} per user in the first version of the manuscript, followed by an extension of \SI{40}{\minute} for more thorough comparisons in the second version. Fatigue was neither observed nor reported by the users. Due to the small sample size, the statistical analysis conducted has limited power. All comparisons use Wilcoxon signed-rank tests on per-subject means ($N=6$).\label{footnote:long_experiments}}
We measure quantitative metrics in the controller and collect task performance metrics using both NASA TLX~\cite{hart1988development} (range: $1-20$, lower = better) and SUS~\cite{brooke1996sus} (range: $0-100$, higher = better) questionnaires as well as a rating of the forces experienced by the user from $0$ (too low) to $10$ (too high).
We complement the evaluation by a \textit{semi-structured interview} about the expected guidance, its implementation with our method and about transitions between different task phases.

\begin{table}
\scriptsize
\caption{\label{tab:experiment_classifications}\centering \scshape Automation, interaction and evaluation metrics.}
\centering
\begin{tabular}{@{}c|ccc@{}}
Section & Automation & Interaction & Evaluation Metrics\\
\hline
\ref{sec:evaluation:coupled_stiffness} & manual & hand-guided & time, precision, usability\\
\ref{sec:evaluation:position_stiffness} & manual & hand-guided & qualitative\\
\ref{sec:evaluation:dynamic_vf} & full & without & repeatability\\
\ref{sec:evaluation:auto_pb_robograv} & full & without & tracking error\\
\ref{sec:evaluation:dynamic_pb_robograv} & full & without & qualitative\\
\ref{sec:evaluation:dynamic_vs_hug} & semi & teleop \& hand-g. & time, forces, usability\\
\ref{sec:evaluation:all_fixtures_sara} & semi & hand-guided & time, forces, success, usability
\end{tabular}
\end{table}

\subsection{Visual Servoing Fixture with Coupled Variable Stiffness}
\label{sec:evaluation:coupled_stiffness}
\begin{figure}
	\centering
	\includegraphics[width=0.8\columnwidth,trim={1230 550 1030 200},clip]{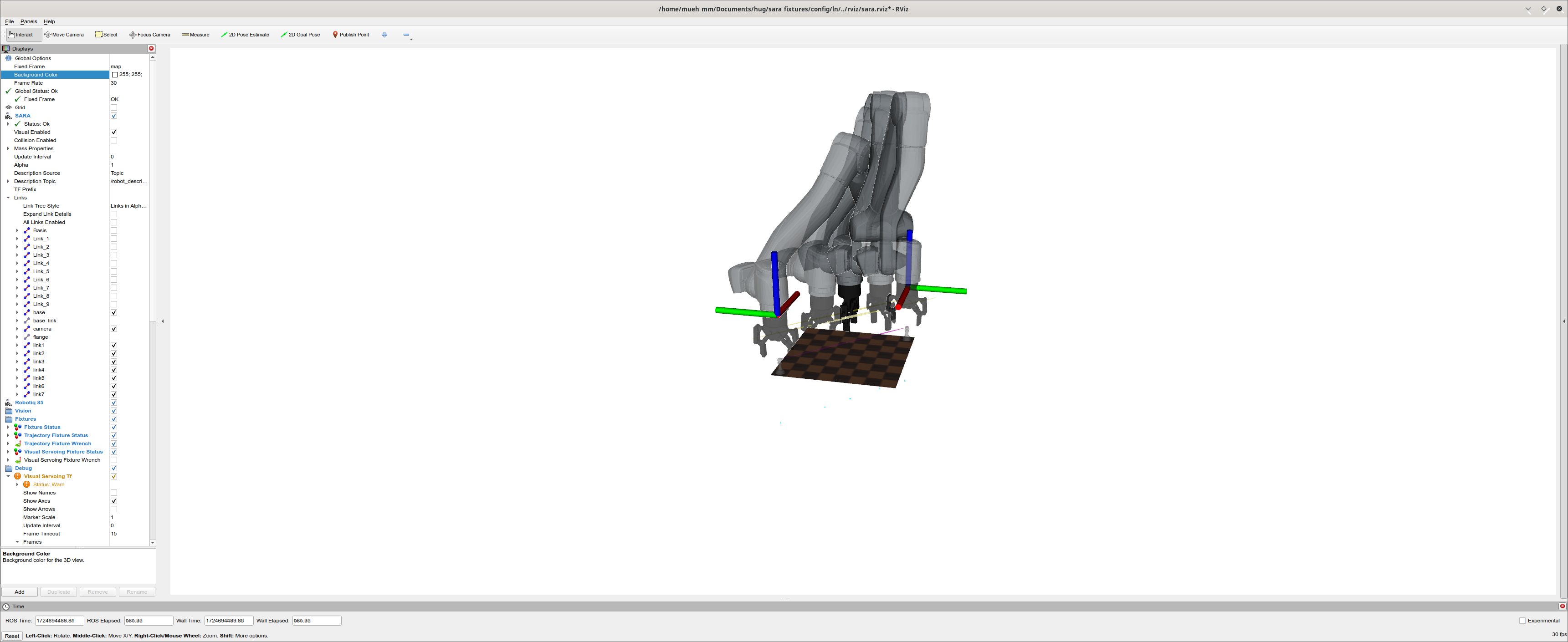}
	\caption{\label{fig:evaluation:variable_stiffness} Visual servoing fixture (\Cref{sec:vs_fixture}) on $\mathcal{M}_1$ with two targets with an orientation difference of \SI{180}{\degree} around the $z$ axis and position differences both along the $x$ and $y$ axes. This leads to a covariance matrix with couplings both inside the positional as well as between positional and rotational \glspl{dof}, therefore necessitating a fully populated stiffness matrix.}
\end{figure}
\Cref{fig:evaluation:variable_stiffness} shows the use case of robotic chess playing with the visual servoing fixture ($N_\mathrm{VS}=1$, $M=2$) on a chess field with size of $\SI{40}{\centi\metre} \times \SI{40}{\centi\metre}$.
Two chess figure detections with covariance $\bm{\Sigma} = 5 \times 10^{-6} \cdot \bm{I}_6$ and a rotation difference of \SI{180}{\degree} are simulated on fields a1 and h8.
With parameters $l_x = l_y = l_z = 0.06$, $l_{wx} = l_{wy} = l_{wz} = 0.2$, $\gamma = 1\times 10^{-20}$, $\lambda_\mathrm{rot}^- = \lambda_\mathrm{trans}^- = 1000$, $\lambda_\mathrm{rot}^+ = \lambda_\mathrm{trans}^+ = 2500$, $k_\mathrm{trans,nom} = 1000$ and $k_\mathrm{rot,nom} = 40$, we obtain the stiffness matrix
\begin{equation}\label{eq:coupled_stiffness_k}
	\bm{K} =
{\scriptsize
\begingroup
\setlength\arraycolsep{2pt}
		\begin{bmatrix}
			1000.0 & * & * & * & * & -111.4\\
			* & 1000.0 & * & * & * & 111.5\\
			* & * & 1000.0 & * & * & -0.8\\
			* & * & * & 40.0 & * & *\\
			* & * & * & * & 40.0 & *\\
			-111.4 & 111.5 & -0.8 & * & * & 24.8
		\end{bmatrix},
\endgroup
}
\end{equation}
with entries $|*| < 10^{-3}$, guiding along the geodesic between both detections, coupling translations with an orientation change (also see the supplementary video; further discussion in \Cref{sec:discussion:stiffness}).
Unlike in \cite{muehlbauer2024probabilistic}, the operator is not always attracted to a detection as the gating function \eqref{eq:gating_function} assigns equal weights when the end effector is located in between.

In the evaluation with expert users, we asked them to move a chess piece from field h8 and place it on a1 with either the attractor averaging between both detections with variable stiffness enabled or only the two attractors (passivated using \cite{muehlbauer2026stabilizing}) as baseline.
After a familiarization phase with variable stiffness, we conducted both cases with one repetition each, counterbalancing the order of the two conditions.
\begin{table}
\scriptsize
\centering
\caption{\label{tab:evaluation:variable_stiffness}\centering \scshape Variable stiffness: quant. evaluation (M $\pm$ SD).}
\begin{tabular}{@{}c|ccc@{}}
Metric & Var. Stiffness & No Stiffness & Wilcoxon signed rank\\
\hline
\textit{Time} & \textbf{\SI[detect-all=true]{7.66 \pm 1.20}{\second}} & \SI{11.47 \pm 4.04}{\second} & $W=0$, $\mathbf{p < 0.05}$\\
\textit{Rot. Error} & \textbf{\SI[detect-all=true]{1.43 \pm 1.77}{\degree}} & \SI{15.79 \pm 18.25}{\degree} & $W=0$, $\mathbf{p < 0.05}$\\
\textit{NASA TLX} & $\mathbf{3.1 \pm 0.6}$ & $5.0 \pm 2.1$ & $W=0$, $\mathbf{p < 0.05}$\\
\textit{SUS} & $\mathbf{94.2 \pm 2.9}$ & $79.6 \pm 17.4$ & $W=0$, $\mathbf{p < 0.05}$
\end{tabular}
\end{table}
\Cref{tab:evaluation:variable_stiffness} summarizes the quantitative evaluation of this experiment.
The rotational error is measured in a vicinity of \SI{\pm 1}{\centi\metre} of the target poses.
We have obtained significant differences for all metrics.

\subsection{Position-based Fixture with Variable Stiffness}
\label{sec:evaluation:position_stiffness}
\begin{figure}
	\centering
	\begin{subfigure}{0.4\columnwidth}
		\includegraphics[width=\columnwidth,trim={1200 250 1200 150},clip]{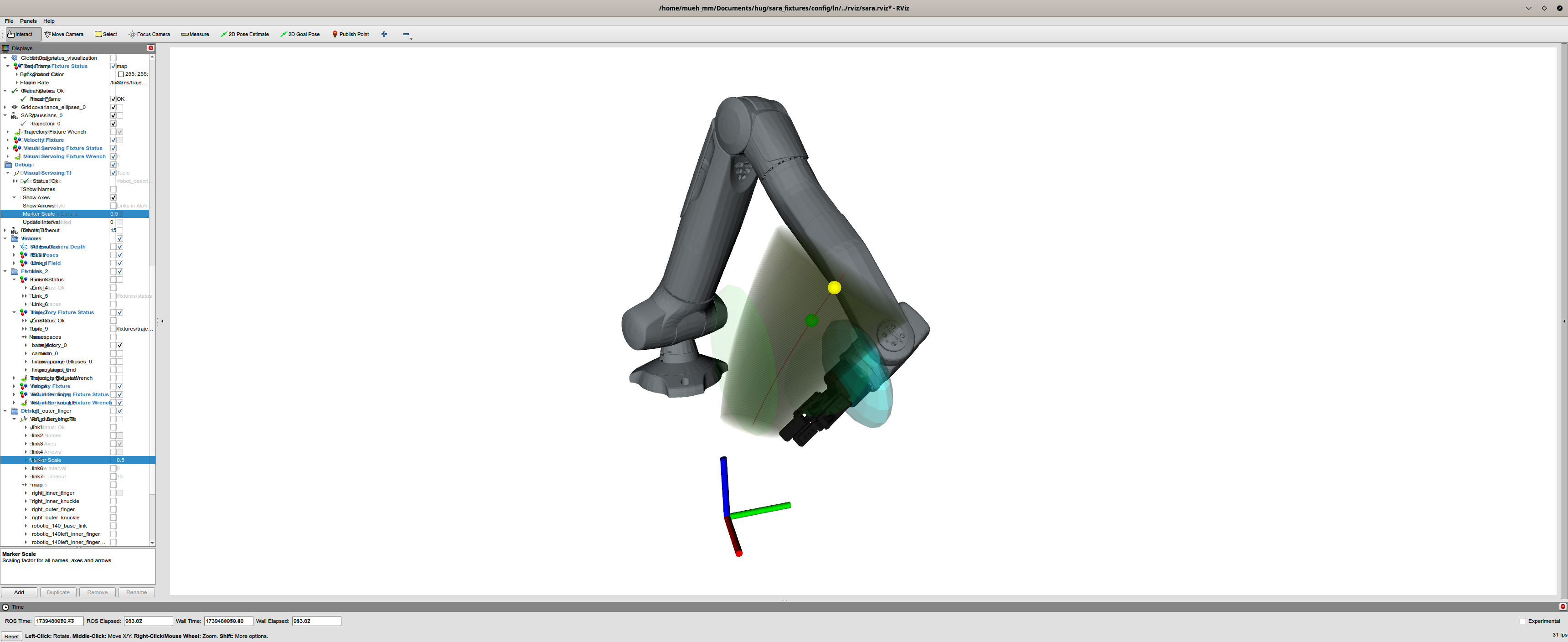}
	\end{subfigure}
	\hfill
	\begin{subfigure}{0.4\columnwidth}
		\includegraphics[width=\columnwidth,trim={100 20 70 40},clip]{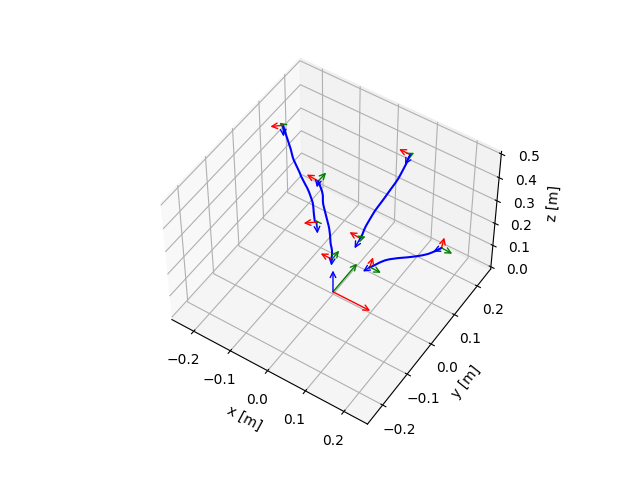}
%
%
	\end{subfigure}
	\caption{\label{fig:evaluation:position_stiffness} Position-based fixture (\Cref{sec:auto_position_based}) on the spherical manifold $\mathcal{M}_3$ ($\mathcal{S}^2 \times \mathbb{R}^1 \times \mathcal{S}^3$) with variable stiffness learned from four demonstrations (right side). The trajectory with red mean and covariance visualized as yellow tube consists of $M=2$ Gaussians plotted as green ellipsoids.
	The green dot on the red mean trajectory depicts the attractor point computed using the Mahalanobis distance ($\bm{A} = \bm{\Sigma}^{-1}$ in \eqref{eq:distance_function_onman_weighted}) while the yellow dot would be computed using a non-weighted distance metric ($\bm{A} = \bm{I}$ in \eqref{eq:distance_function_onman_weighted}).}
\end{figure}

\Cref{fig:evaluation:position_stiffness} shows the position-based trajectory fixture ($N_\mathrm{PB} = 1$, $M=2$) used for a pointing task on the spherical manifold $\mathcal{M}_3$ learned from four demonstrations of a movement towards the coordinate origin.
The resulting fixture exhibits a large covariance around the two rotational \glspl{dof} of $\mathcal{S}^2$ (visualized by the turquoise Gaussian at the end effector), resulting in a precision matrix with small entries $P_{1,1}$, $P_{2,2}$ and $P_{3,3}$.
With $\lambda_\mathrm{rot}^- = 0.5$, $\lambda_\mathrm{rot}^+ = 1.5$, $\lambda_\mathrm{trans}^- = 400$, $\lambda_\mathrm{trans}^+ = 500$, $k_\mathrm{trans,nom} = 500$ and $k_\mathrm{rot,nom} = 40$, we obtain a corresponding stiffness
\begin{gather}
\underbrace{\scriptsize
\begingroup
\setlength\arraycolsep{2pt}
		\begin{bmatrix}
			\num{344} & \num{-23} & \num{-109} & \num{-110} & \num{280} & \num{-166}\\
			\num{-23} & \num{46} & \num{58} & \num{133} & \num{48} & \num{31}\\
			\num{-109} & \num{58} & \num{723} & \num{-74} & \num{-22} & \num{78}\\
			\num{-110} & \num{133} & \num{-74} & \num{794} & \num{58} & \num{112}\\
			\num{280} & \num{48} & \num{-22} & \num{58} & \num{533} & \num{-81}\\
			\num{-166} & \num{31} & \num{78} & \num{112} & \num{-81} & \num{94}
		\end{bmatrix}
\endgroup
}_{\bm{P}}
\quad
\underbrace{\scriptsize
\begingroup
\setlength\arraycolsep{2pt}
		\begin{bmatrix}
			\num{5} & \num{-2} & \num{-20} & * & \num{2} & \num{-3}\\
			\num{-2} & * & \num{7} & * & * & \num{1}\\
			\num{-20} & \num{7} & \num{74} & \num{-3} & \num{-9} & \num{12}\\
			* & * & \num{-3} & \num{40} & * & *\\
			\num{2} & * & \num{-9} & * & \num{41} & \num{-1}\\
			\num{-3} & \num{1} & \num{12} & * & \num{-1} & \num{42}
		\end{bmatrix}
\endgroup
}_{\bm{K}}
\nonumber
\end{gather}
with entries $|*| < 1$.
Compared to the nominal stiffness matrix $\bm{K}_\mathrm{nom} = \mathrm{diag}(80, 80, 500, 40, 40, 40)$ at $r = \SI{16}{\centi\metre}$, these values allow a human operator to freely move around the object centered at the coordinate origin while always pointing at it which is, e.g., valuable for inspection tasks.

Crucial for this fixture is also the correct attractor selection through
\eqref{eq:pb_fixture:l0ee}, \eqref{eq:pb_fixture:mahalanobis_projection}, ensuring that the attractor stays the same when moving along a zero-force direction extracted by the variable impedance control (\Cref{sec:problem:variable_impedance}).
This is achieved by computing both attractor as well as stiffness from the precision $\bm{P}_\mathrm{VF} = \bm{\Sigma}_\mathrm{VF}^{-1}$, ensuring that length scales in the distance calculation correspond to stiffness scales (\Cref{fig:evaluation:position_stiffness}).

\subsection{Probabilistic Dynamical System Virtual Fixtures}
\label{sec:evaluation:dynamic_vf}
\begin{figure}
	\centering
	\annotategraphicsmulti{
	    \includegraphics[width=0.9\columnwidth,trim={850 150 800 150},clip]{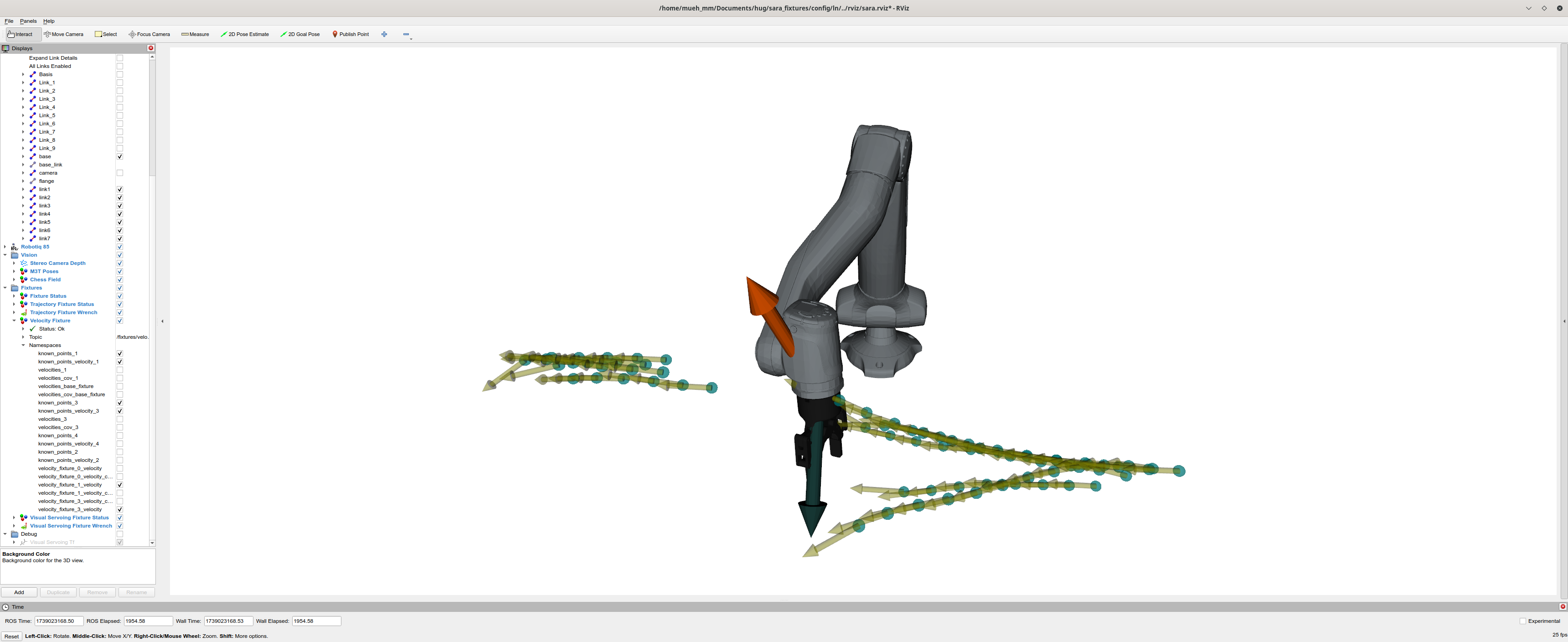}
	}{
        \node at (0.75,0.3) {1};
        \node at (0.25,0.3) {2};
    }
	\caption{\label{fig:evaluation:velocity_gp} \gls{ds} based \gls{vf} using a \gls{gp} model. Yellow arrows starting at turquoise dots visualize known velocities. The green arrow corresponds to the output of the velocity policy \protect\lref{1} on the right side while the red arrow depicts the velocity output of the velocity policy \protect\lref{2}. Both counteract each other due to erroneous velocity measurements at the borders of the dataset, leading to a stuck evolution of the system unable to transition from \protect\lref{1} to \protect\lref{2}.}
\end{figure}

\begin{figure}
	\centering
	\annotategraphicsmulti{
	    \includegraphics[width=0.9\columnwidth,trim={850 150 800 150},clip]{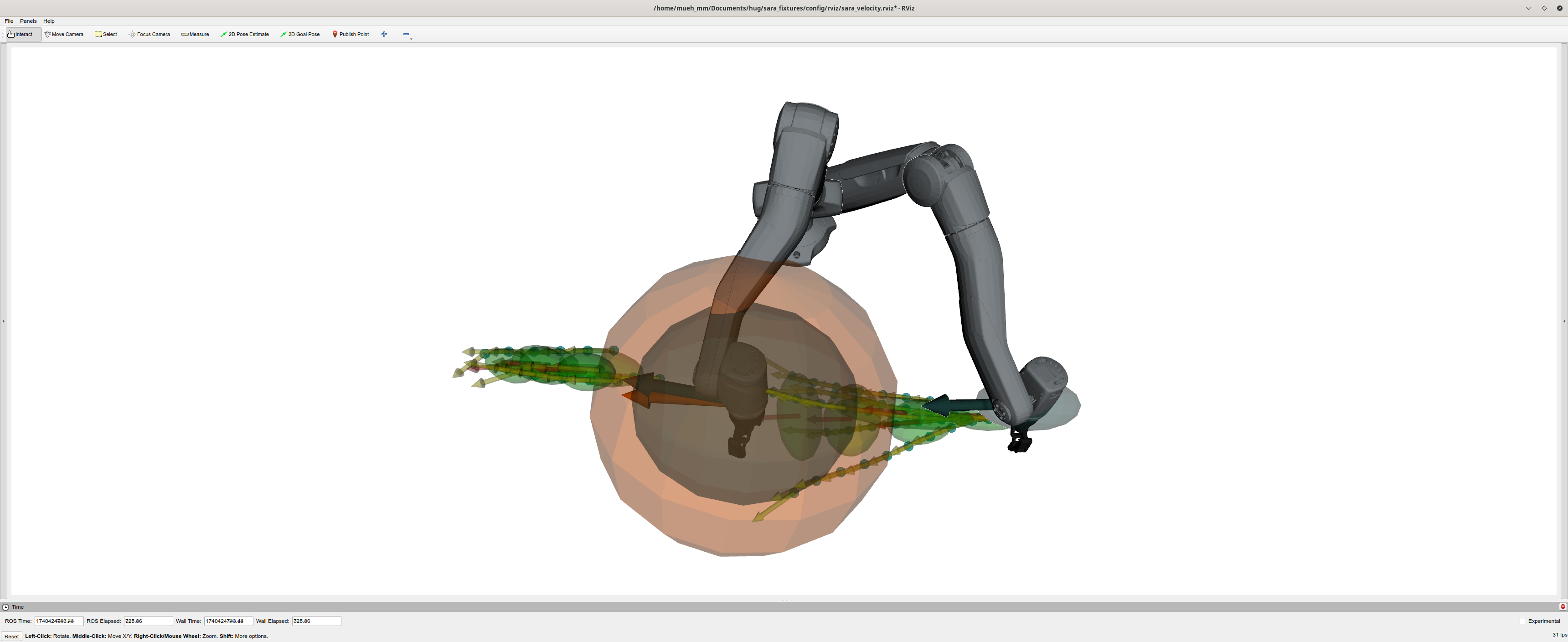}
	}{
        \node at (0.72,0.65) {s};
        \node at (0.47,0.7) {t};
        \node at (0.65,0.2) {1};
        \node at (0.2,0.3) {2};
    }
	\caption{\label{fig:evaluation:velocity_kmp} \gls{ds}-based \gls{vf} using a \gls{kmp} model for the robot in start~\protect\lref{s} and transition~\protect\lref{t} configurations.
	Yellow arrows starting at turquoise dots visualize known velocities.
	Green Gaussians depict the positional uncertainty of the underlying \gls{gmm} with mean velocity as red arrows.
	The dark green Gaussian with green velocity arrow at the end effector represents policy~\protect\lref{1}.
	The orange-red Gaussian with red velocity arrow visible in~\protect\lref{t} configuration corresponds to policy~\protect\lref{2}.
	In~\protect\lref{t}, a smooth transition between policies is happening.
	In~\protect\lref{s} configuration, no output for policy~\protect\lref{1} is visible due to its high uncertainty.}
\end{figure}

We evaluate the novel \gls{ds} based \gls{vf} (\Cref{sec:dynamic_fixtures}) on two \gls{ds} policies consisting of transport motions from right to left in the robot's workspace (\Cref{fig:evaluation:velocity_gp,fig:evaluation:velocity_kmp}).
The right motion consists of five demonstrations diverging towards the middle of the workspace while the left motion consists of four demonstrations which are very close together.
For evaluation purposes we compare the performance of \gls{gp} and \gls{kmp} representations on $\mathcal{M}_1$.
To limit the amount of data used in the models, the recorded trajectories are subsampled at a distance of \SI{5}{\centi\metre} before calculating the velocities based on time differences.

Through empirical trials we found that with human interaction, policies using a full pose $\bm{x} \in \mathcal{M}$ as input and outputting a velocity $\dot{\bm{x}} \in \mathbb{R}^6$ lead to non-smooth behavior due to the curse of dimensionality.
When a human perturbs the robot's orientation, the epistemic uncertainty of the learned policy increases, activating the \textit{stabilizing policy}.
This policy only acts towards the closest pose $\bm{x}$ without forward motion component $\dot{\bm{x}}$, therefore halting the evolution of the \gls{ds}.
Splitting the \gls{ds} in two components, one with position input $\bm{x}_\mathrm{pos}\in\mathbb{R}^3$ and output velocity $\dot{\bm{x}}_\mathrm{pos}\in\mathbb{R}^3$ and the other with the full pose as input $\bm{x}\in\mathcal{M}$ and rotational velocity $\bm{\omega} \in \mathcal{T}_{\bm{x}_\mathrm{rot}}^{\mathcal{S}^3}$ as output, this problem can be mitigated.
The full set of velocity policies for this experiment therefore consists of $N_\mathrm{DS} = 5$ concurrently active policies: one \textit{stabilizing policy} and two policies each for the left and the right side of the motion.
All policies are fused through the arbitration (\Cref{sec:problem:arbitration}).

\begin{figure}
	\centering
	\annotategraphicsmulti{
	    \includegraphics[width=\columnwidth,trim={0 0 0 0},clip]{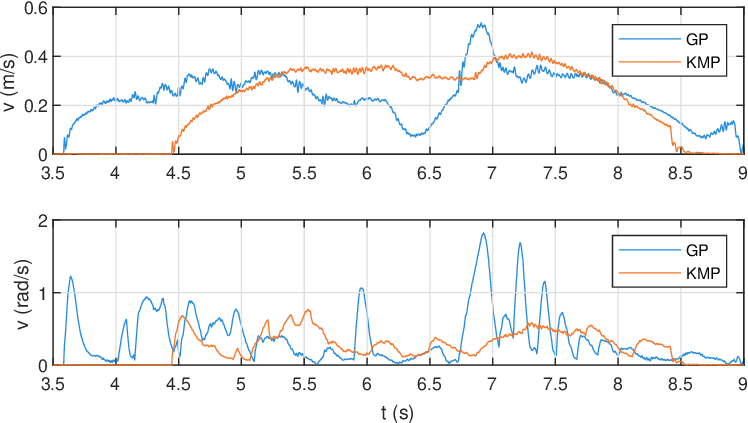}
	}{
    }
	\caption{\label{fig:evaluation:velocity_profile_joined} Translational (upper plot) and rotational (lower plot) velocities observed during a human interaction with \gls{gp}- and \gls{kmp}-based velocity fixtures.
	With the \gls{kmp}, both rotational velocities as well as the transition from policy~\protect\lref{1} to~\protect\lref{2} at $t = \SI{6.5}{\second}$ are much smoother compared to the \gls{gp}.}
\end{figure}

In \Cref{fig:evaluation:velocity_gp}, policies encoded using a \gls{gp} are visualized.
For the positional \gls{gp} we use a \gls{rbf} kernel \eqref{eq:dynamic_fixtures:rbf} with $l = 0.1$ and for the rotational \gls{gp} $l = 0.03$.
We set both process variances to $\lambda = 0.01$.
As shown in the supplementary video, the evolution of the \gls{ds} closely follows the demonstration trajectories but fails to capture their variance, both for left and right policies.
Furthermore, the velocities generated by the \gls{gp}, as can be seen in \Cref{fig:evaluation:velocity_profile_joined}, exhibit abrupt changes, suggesting limited smoothness in the generated trajectories.
A higher value of $\lambda$ could smoothen the prediction of the process, which would however have a global effect and not be restricted to the demonstrated high-variance zone.
The robot also fails to transition between policies, halting at the final pose of policy~\protect\lref{1} and requiring operator input to proceed.

\begin{figure}
	\centering
	\annotategraphicsmulti{
	    \includegraphics[width=\columnwidth,trim={0 0 0 0},clip]{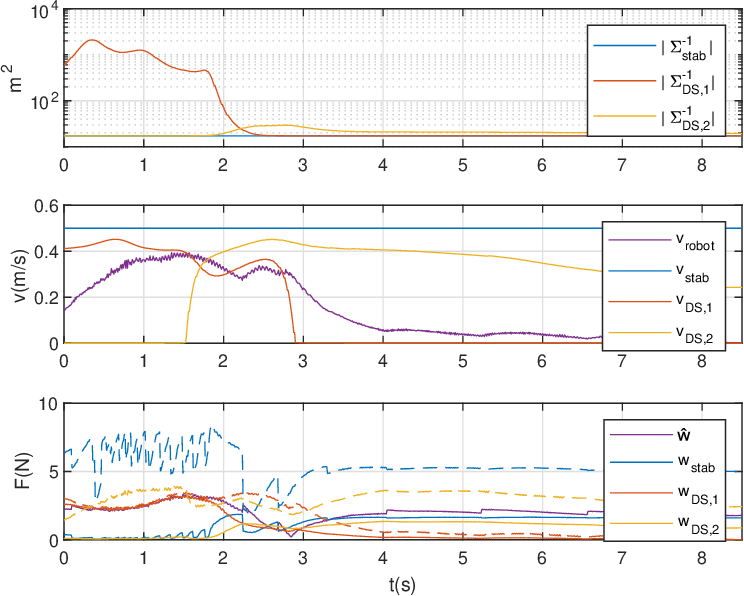}
	}{
    }
	\caption{\label{fig:evaluation:ds_pose_diffs} Norm of precisions (\textbf{top}), target velocities (\textbf{middle}), raw (dashed, $\bm{w}_{\mathrm{VF},i}$) and arbitrated (solid lines, $\hat{\bm{\Sigma}}_{\mathrm{VF}} \bm{\Sigma}^{-1}_{\mathrm{VF},i} \bm{w}_{\mathrm{VF},i}$ resp. $\hat{\bm{w}}$, see \eqref{eq:vf_covariance_arbitration}) fixture forces (\textbf{bottom}) of the \textit{stabilizing policy} and learned \gls{kmp}-\gls{ds}.}
\end{figure}

\Cref{fig:evaluation:velocity_kmp} shows the same data encoded in a \gls{ds} using a \gls{kmp} based on a \gls{gmm} with $M=5$ Gaussians with hyperparameters $\lambda = 0.05$, $\lambda_c = 10$, $\alpha = 0.1$ and $l = 0.1$ in the \gls{rbf} kernel for the positional respectively $l = 0.03$ for the rotational \gls{kmp}.
The \gls{kmp} encodes a small covariance at~\protect\lref{s} and a bigger covariance at~\protect\lref{t} for the right motion policy.
This also enables a system evolution from right to left policy where the \gls{gp}-based policy failed.
Unlike the previous model, the learned covariance permits smoother motion in the high-variance region of the right policy, allowing the robot to deviate from the demonstrations when appropriate, as also seen in the supplementary video.
Furthermore, a much smoother velocity profile is obtained (\Cref{fig:evaluation:velocity_profile_joined}).
The arbitration (\Cref{sec:problem:arbitration}) plays a key role in coordinating the \textit{stabilizing policy} and the learned policies which is shown for the position components in \Cref{fig:evaluation:ds_pose_diffs}.
First, \gls{ds}~\protect\lref{1} dominates the execution, then followed by a mix of \gls{ds}~\protect\lref{2} and the \textit{stabilizing policy} as can be seen from arbitrated forces which sum to $\hat{\bm{w}}$ \eqref{eq:vf_covariance_arbitration}.
The motion converges towards the end of policy~\protect\lref{2}, when learned and \textit{stabilizing policy} as well as robot friction cancel each other.
Over $5$ \gls{kmp} policy executions, a standard deviation of $(0.006, 0.012, 0.006)~\mathrm{m}$ and $(0.012, 0.005, 0.012)~\mathrm{rad}$ between end effector poses as well as $(0.007, 0.011, 0.008)~\mathrm{\frac{m}{s}}$ and $(0.026, 0.019, 0.053)~\mathrm{\frac{rad}{s}}$ for velocities at the same elapsed time from start of execution results.
Note that an error between desired and measured velocity as visible in \Cref{fig:evaluation:ds_pose_diffs} is required for the proportional controller \eqref{eq:dynamic_fixtures:vel_ctrl} to generate a wrench.

{}

\begin{figure}
	\centering
	\annotategraphicsmulti{
	    \includegraphics[width=0.8\columnwidth,trim={800 110 600 190},clip]{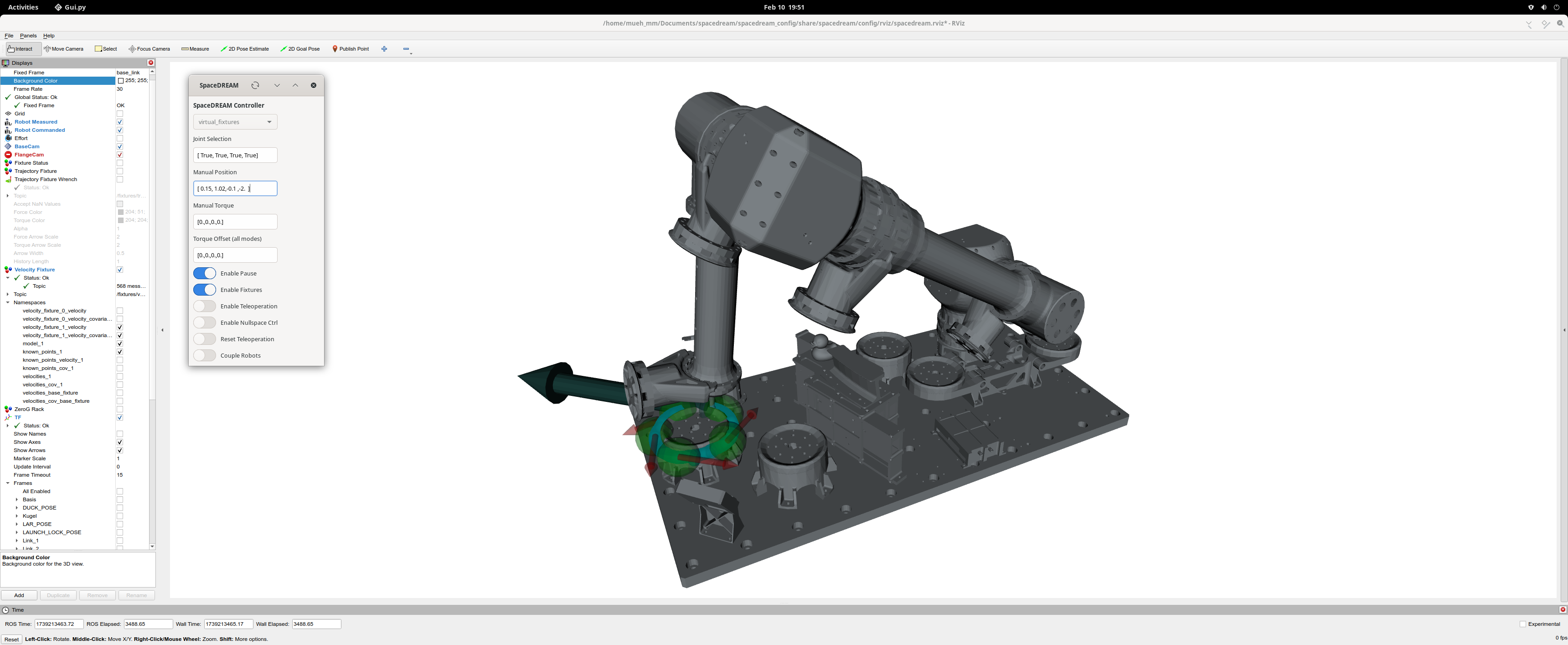}
	}{
    }
	\caption{\label{fig:eval:dynamic_robograv}Dynamical system \gls{vf} on the RoboGrav space robot. A repeating motion in $\mathcal{S}^1 \times \mathbb{R}^2$ around the launch lock is overlaid with a local policy exerting a force perpendicular to the launch lock. The $5$ green ellipsoids visualize the position-based covariances of the underlying \gls{gmm} model with the red arrows depicting its velocities. The green arrow in the back visualizes the current output of the velocity fixture along the launch lock.}
\end{figure}

\begin{figure}
	\centering
	\annotategraphicsmulti{
	    \includegraphics[width=0.94\columnwidth,trim={0 0 0 0},clip]{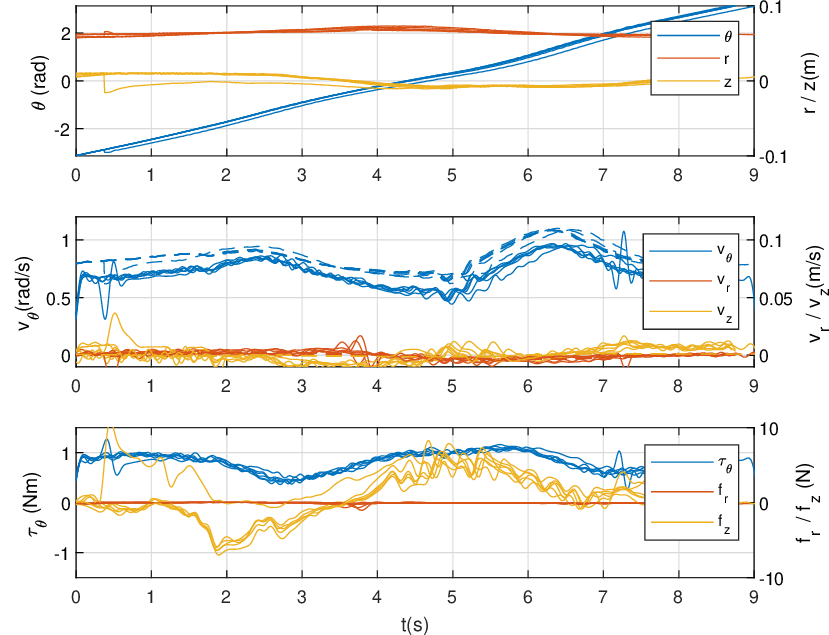}
	}{
    }
	\caption{\label{fig:evaluation:ds_flight} Robot poses on $\mathcal{M}_2$ (\textbf{top}), desired (dashed) and measured velocities (\textbf{middle}) and computed forces (\textbf{bottom}) for $6$ repetitions of the circular motion shown in \Cref{fig:eval:dynamic_robograv} during the \SI{0}{\gravity} phase of the parabolic flight.
	$\theta$, $r$ and $z$ are obtained by computing $\mathrm{Log}_{\bm{e}}^{\mathcal{M}_2} ( \cdot )$ with $\bm{e} = \left( 0, 1, 0, 0, 0, 0, 0, 1 \right)^\top \in \mathcal{M}_2$.
	Wrenches and velocities are plotted in the (co)tangent space of $\bm{x}_\mathrm{ee}$.
	One trajectory deviates from the rest with a disturbance in $z$ caused by starting the robot already before the \SI{0}{\gravity} phase, but recovers over time as the \textit{stabilizing policy} outputs a desired velocity and thus force in $+z$ direction.}
\end{figure}

\begin{figure}
	\centering
	\begin{subfigure}{0.64\columnwidth}
		\annotategraphicsmulti{
	    		\includegraphics[width=\columnwidth,trim={500 200 700 450},clip]{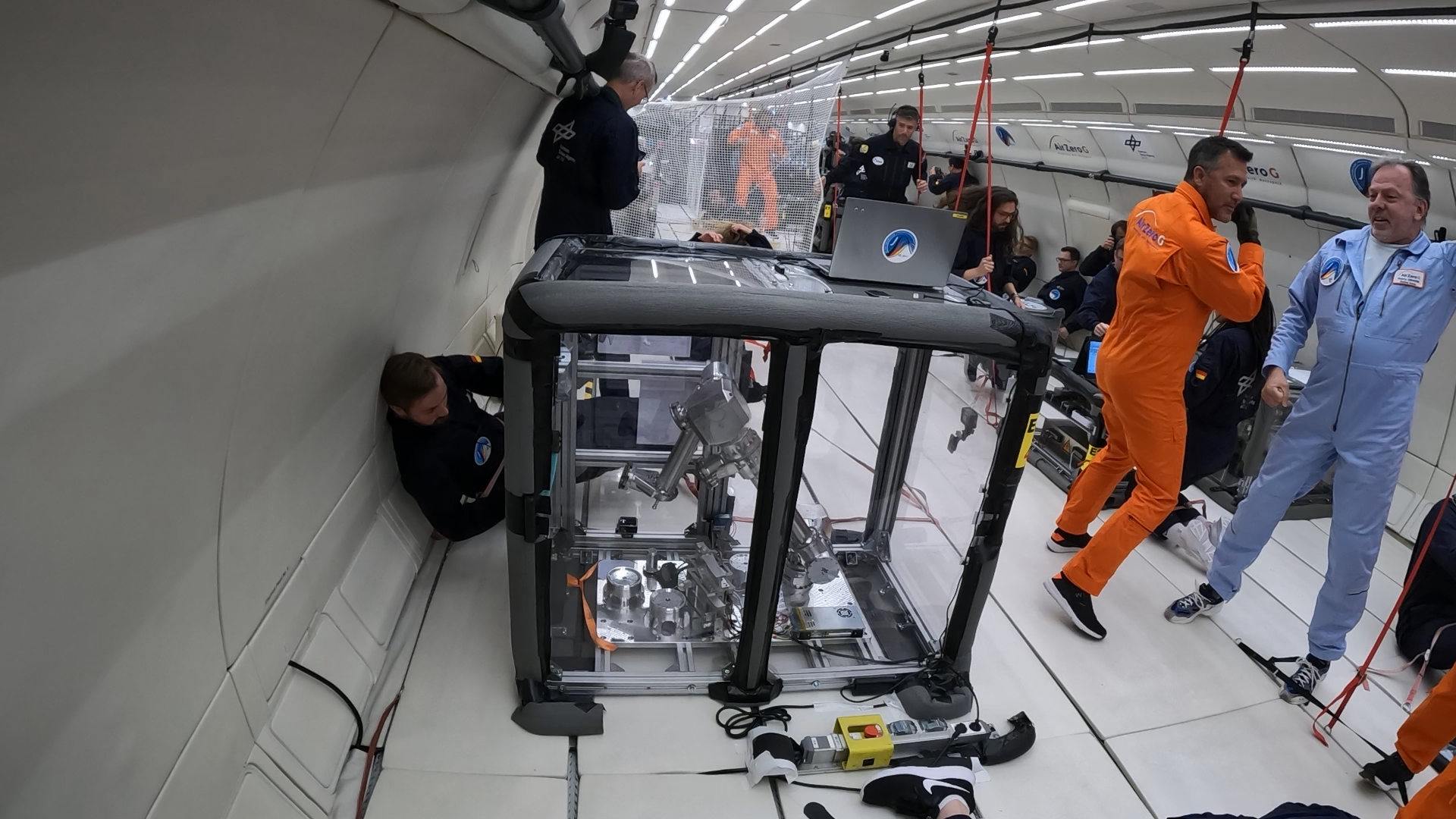}
		}{
        		\node at (0.4,0.17) {LL};
        		\node at (0.5,0.25) {S};
        		\node at (0.6,0.35) {LAR};
    		}
    	\end{subfigure}
	\begin{subfigure}{0.34\columnwidth}
		\annotategraphicsmulti{
	    		\includegraphics[width=\columnwidth,trim={1300 210 1300 150},clip]{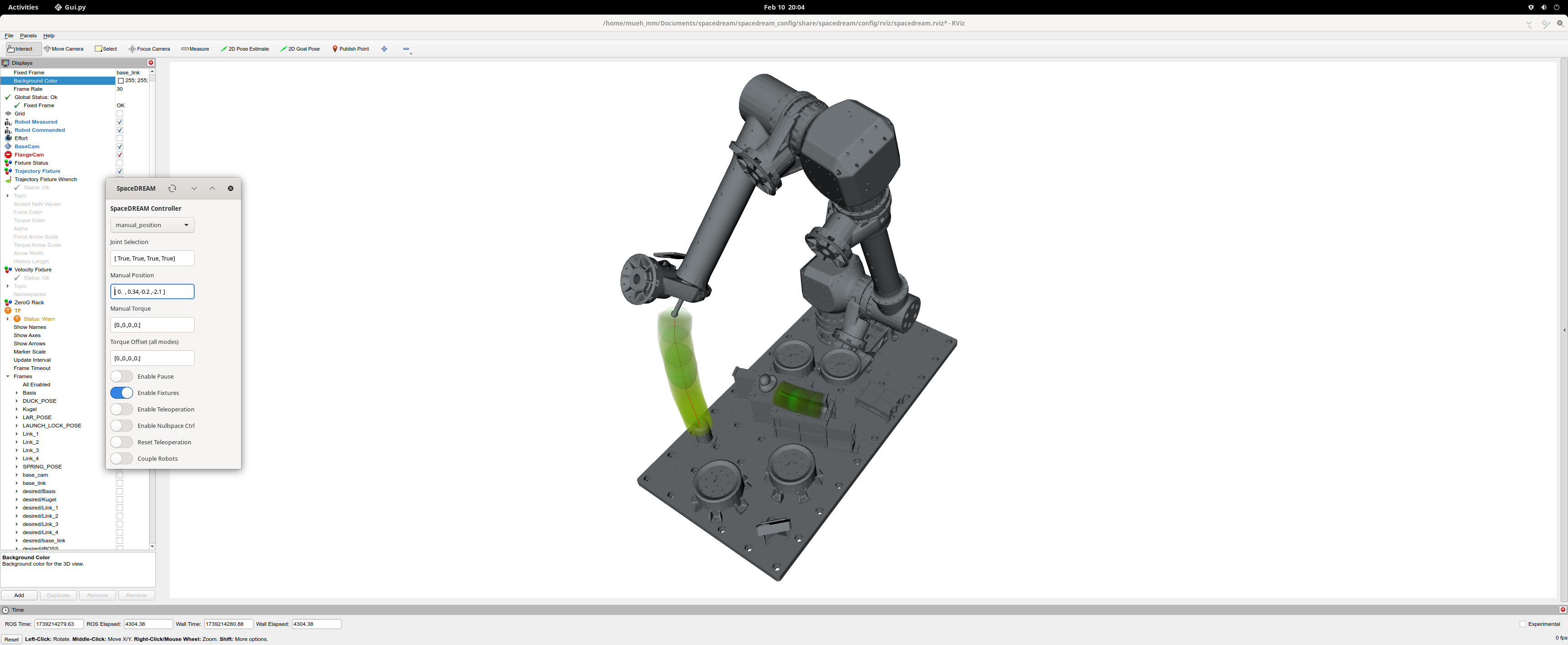}
		}{
        		\node at (0.35,0.5) {PB1};
        		\node at (0.5,0.3) {PB2};
        		\node at (0.34,0.3) {S};
        		\node at (0.6,0.44) {LAR};
    		}
	\end{subfigure}
	\caption{Left image: The RoboGrav robot approaching the spring~\protect\lref{S} using an automated position-based \gls{vf} during the \SI{0}{\gravity} phase of a parabolic flight. Also shown are the launch adapter ring segment~\protect\lref{LAR} and the launch lock~\protect\lref{LL}.
	Right image: \protect\lref{PB1} denotes a position-based fixture for approaching and pressing the spring assembly~\protect\lref{S} while \protect\lref{PB2} shows the second position-based fixture for the robot moving along the launch adapter ring segment~\protect\lref{LAR}.}
	\label{fig:eval:robograv_spring_approach}
\end{figure}
\begin{figure}
	\centering
	\annotategraphicsmulti{
	    \includegraphics[width=0.9\columnwidth,trim={0 0 0 0},clip]{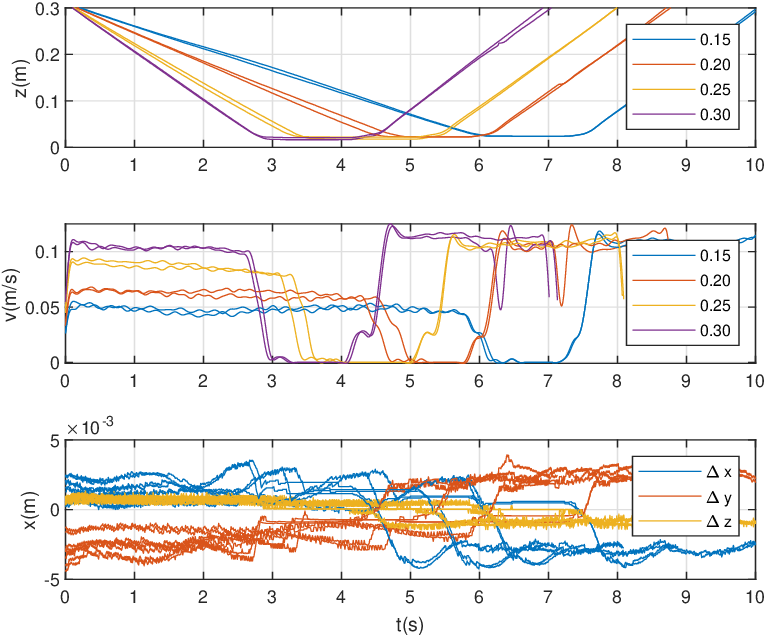}
	}{
    }
	\caption{\label{fig:evaluation:robograv_spring_flight} Repeated execution of the spring pressing task (\Cref{fig:eval:robograv_spring_approach}) with different velocities. Differences between the setpoint velocities of \SI{0.15}{\metre\per\second}, \SI{0.2}{\metre\per\second}, \SI{0.25}{\metre\per\second} and \SI{0.3}{\metre\per\second} for pressing - for retracting, we always use \SI{-0.3}{\metre\per\second} - and the velocities visible in the middle plot are caused by friction in the joints that has to be overcome by the velocity controller with a gain of \SI{150}{\newton\second\per\metre}.}
\end{figure}
\begin{figure*}
\centering
\begin{subfigure}{0.24\textwidth}
  \centering
  \annotategraphicsmulti{
  \includegraphics[width=\columnwidth,trim={0 0 250 0},clip,page=1]{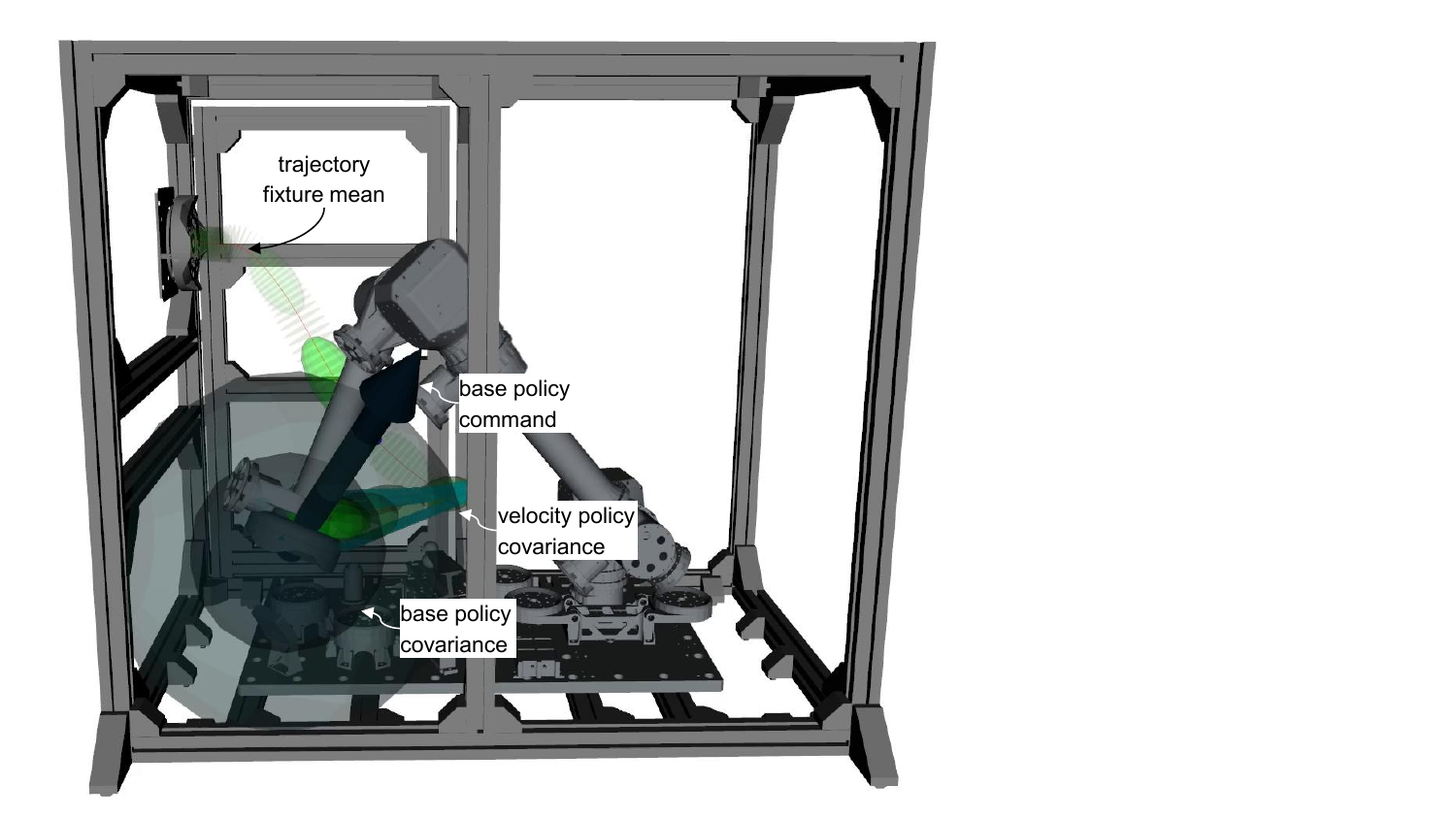}
  }{
        \node at (0.52,0.41) {DS};
        \node at (0.28,0.6) {PB};
        \node at (0.13,0.8) {DOCK};
  }
  \subcaption{\label{fig_moe_atpoint} \textit{Stabilizing policy} active.}
\end{subfigure}
\hfill
\begin{subfigure}{0.24\textwidth}
  \centering
  \annotategraphicsmulti{
  \includegraphics[width=\columnwidth,trim={0 0 250 0},clip,page=2]{bilder.pdf}
  }{
        \node at (0.52,0.41) {DS};
        \node at (0.28,0.6) {PB};
        \node at (0.13,0.8) {DOCK};
  }
  \subcaption{\label{fig_moe_between} Approaching the trajectory.}
\end{subfigure}
\hfill
\begin{subfigure}{0.24\textwidth}
  \centering
  \annotategraphicsmulti{
  \includegraphics[width=\columnwidth,trim={0 0 250 0},clip,page=3]{bilder.pdf}
  }{
        \node at (0.52,0.41) {DS};
        \node at (0.28,0.6) {PB};
        \node at (0.13,0.8) {DOCK};
  }
  \subcaption{\label{fig_moe_far} The trajectory takes over.}
\end{subfigure}
\hfill
\begin{subfigure}{0.24\textwidth}
  \centering
  \annotategraphicsmulti{
  \includegraphics[width=\columnwidth,trim={0 0 250 0},clip,page=4]{bilder.pdf}
  }{
        \node at (0.52,0.41) {DS};
        \node at (0.28,0.6) {PB};
        \node at (0.13,0.8) {DOCK};
  }
  \subcaption{\label{fig_moe_right} Successful docking.}
\end{subfigure}
\caption{\gls{ds}-based \gls{vf} \protect\lref{DS} combined with a position-based fixture \protect\lref{PB} on the RoboGrav space robot. The \gls{ds}-based \gls{vf} guides the robot towards the start of the automated position-based fixture performing a docking of the interface mounted at the end effector to the interface mounted on the rack \protect\lref{DOCK}.}
\label{fig:eval:dynamic_pb_robograv}
\end{figure*}

We encode repetitive motions in the same policy representation (\Cref{fig:eval:dynamic_robograv}).
This fixture ($N_\mathrm{DS}=2$) is learned on $\mathcal{M}_2$ from one demonstration of circling the launch adapter ring of the RoboGrav space robot $4$ times.
It is encoded in a \gls{kmp} based on a \gls{gmm} with $M=5$ Gaussians with hyperparameters $\lambda = 0.1$, $\lambda_c = 10$, $\alpha = 0.1$, $h = 1$ and $l = 0.03$.
We use the arbitration to overlay a wrench pointing along $-r$ with
\begin{gather}
	\label{eq:robograv_local_policy}
	\bm{w} = (0, -\SI{10}{\newton}, 0)^\top,
	\quad
	\bm{\Sigma} = \mathrm{diag}(0, \SI{2e-4}{\metre\squared}, 0).
\end{gather}
We evaluate the motion both on ground as well as under \SI{0}{\gravity} conditions, controlling only the position of the robot.
\Cref{fig:evaluation:ds_flight} shows trajectories, computed forces, desired and measured velocities of $6$ repetitions of the policy on the parabolic flight; the same plot on ground looks very similar and is therefore omitted.
We obtain a standard deviation of $(\SI{0.069}{\radian}, \SI{0.001}{\metre}, \SI{0.003}{\metre})$ for positions as well as of $(\SI{0.493}{\radian\per\second}, \SI{0.013}{\metre\per\second}, \SI{0.018}{\metre\per\second})$ for velocities during flight as well as of $(\SI{0.055}{\radian}, \SI{0.001}{\metre}, \SI{0.001}{\metre})$ and $(\SI{0.207}{\radian\per\second}, \SI{0.007}{\metre\per\second}, \SI{0.007}{\metre\per\second})$ on ground.
Higher deviations during flight are caused by stopping the robot in between the \SI{0}{\gravity} phases.
Despite only aiming for \textbf{coarse guidance}, small standard deviations in both experiments demonstrate a remarkably high repeatability of the \gls{ds} based fixture.
Note that this metric does not measure correctness of guidance, as a systematic bias could also result in repeatable behavior.

\subsection{Automated Position-based Fixtures}
\label{sec:evaluation:auto_pb_robograv}
For higher precision motions, where following a defined trajectory is required, an automated position-based trajectory fixture (\Cref{sec:auto_position_based}) is well suited.
We first demonstrate this fixture on the fully automated task of pressing a spring on the RoboGrav robot, again only controlling the position, with the pin end effector of the robot as visualized in \Cref{fig:eval:robograv_spring_approach}.
The fixture is learned from two kinesthetic demonstrations of pressing the spring using $M=5$ Gaussians.
Through \gls{gmr}, we retrieve a reference trajectory denoted as \lref{PB1}.
For the evaluation, we first move the robot to the start position of the fixture.
During the \SI{0}{\gravity} phase of the parabolic flight, we press the spring four times with different velocities as shown in the supplementary video and \Cref{fig:evaluation:robograv_spring_flight}.
During two repetitions of this procedure, standard deviations of the end effector pose from the attractor point on the fixture of $(0.002, 0.002, 0.001)~\mathrm{m}$ both during flight and on ground result at a constant stiffness of \SI{3000}{\newton\per\metre}.
Those deviations lie in the expected range for the robot's capabilities and model tuning.
Similar results on ground as well as under \SI{0}{\gravity} conditions can be explained by slight errors in the gravity compensation model for the former as well as disturbances during flight for the latter.

Concurrently ($N_\mathrm{PB} = 2$), a second position-based fixture \protect\lref{PB2} learned from $4$ demonstrations of a circular motion along the \protect\lref{LAR} on the $\mathcal{S}^1 \times \mathbb{R}^2$ manifold using $M=5$ Gaussians is present in the robot workspace.
However, thanks to the distance-based covariance adaptation (\Cref{sec:auto_position_based:cov_adaptation}) only the fixture close to the end effector creates forces.
For activating \protect\lref{PB2}, the end effector is placed close to the \lref{LAR}.
As in the previous section, we overlay a probabilistic force of \SI{-10}{\newton} along the radius of the cylindrical manifold, leading to the robot pressing against the \lref{LAR}.
The motion is executed both during the \SI{0}{\gravity} phase of the flight and on ground.
With a standard deviation of the position error of $(\SI{0.001}{\radian}, \SI{0.002}{\metre}, \SI{0.001}{\metre})$ both during flight and on ground, we achieve a more \textbf{precise guidance} than with \gls{ds}-based fixtures.

\subsection{Combining Dynamical System and Position-based Fixtures}
\label{sec:evaluation:dynamic_pb_robograv}
\begin{figure}
	\centering
	\annotategraphicsmulti{
	    \includegraphics[width=\columnwidth,trim={0 0 0 10},clip]{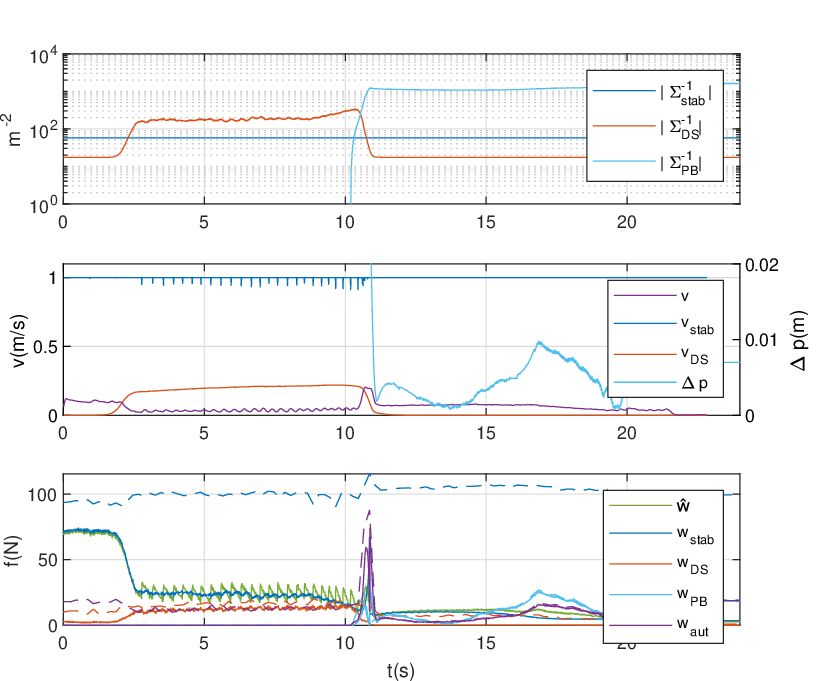}
	}{
    }
	\caption{\label{fig:evaluation:docking_parabolic} Norm of precisions (\textbf{top}), target velocities and position offset (\textbf{middle}), raw (dashed, $\bm{w}_{\mathrm{VF},i}$) and arbitrated (solid lines, $\hat{\bm{\Sigma}}_{\mathrm{VF}} \bm{\Sigma}^{-1}_{\mathrm{VF},i} \bm{w}_{\mathrm{VF},i}$ resp. $\hat{\bm{w}}$, see \eqref{eq:vf_covariance_arbitration}) fixture forces (\textbf{bottom}) of the \textit{stabilizing policy}, \gls{ds}-based and trajectory ($\bm{w}_\mathrm{PB}$, $\bm{w}_\mathrm{aut}$ from \ref{sec:auto_position_based:automation}) fixtures used for docking (\Cref{fig:eval:dynamic_pb_robograv}).}
\end{figure}
With the uncertainty-aware probabilistic fusion of fixtures through \eqref{eq:vf_covariance_arbitration} being at the core of our method, we now combine \glspl{vf} based on dynamical systems and position-based fixtures on the RoboGrav robot.
\Cref{fig:eval:dynamic_pb_robograv} shows the scenario of docking the iBOSS ``iSSI'' interface mounted to the end effector of the space robot with its counterpart mounted to the rack, which we solve by combining a dynamical system \gls{vf} \lref{DS} ($N_\mathrm{DS} = 2$) with a position-based trajectory fixture \lref{PB} ($N_\mathrm{PB} = 1$), both on the $\mathbb{R}^3$ manifold.
The \gls{ds} is encoded in a \gls{kmp} based on a \gls{gmm} with $M=5$ Gaussians with hyperparameters $\lambda = 0.1$, $\lambda_c = 10$, $\alpha = 0.1$, $h = 1$ and $l = 0.03$.
The trajectory fixture is encoded in a \gls{gmm} with $M=5$ Gaussians.
\Cref{fig:eval:dynamic_pb_robograv} shows that both fixtures have similar covariance values thanks to the formulation chosen in \Cref{sec:dynamic_fixtures}, allowing for an arbitration of their wrenches even though they are computed from different quantities.
We thus combine a more flexible velocity fixture with the precision of a position-based trajectory fixture that can model the exact approach required for a successful mating of the interface.
\Cref{fig:evaluation:docking_parabolic} visualizes fixture target poses and velocities, resulting forces as well as their arbitrated forces summing to $\hat{\bm{w}}$ \eqref{eq:vf_covariance_arbitration}.
The plot shows that first, the \textit{stabilizing policy} moves the end effector towards the learned \gls{ds} which is active until the position-based fixture takes over.
Overall, the system behaves as expected even though the \textit{stabilizing policy} is equipped with both non-optimal covariance $\sigma_\mathrm{stab} = \SI{0.04}{\metre\squared}$ and default velocity $\dot{x}_\mathrm{stab} = \SI{1}{\metre\per\second}$, causing it to also generate forces when only other fixtures should be active.

\subsection{Combining Dynamical System and Visual Servoing Fixtures}
\label{sec:evaluation:dynamic_vs_hug}
As explored in previous works \cite{muehlbauer2022multiphase,muehlbauer2024probabilistic,muehlbauer2024ai}, visual input is often required to successfully accomplish the precision requirements of a task.
We therefore analyze the task of CubeSat subsystem assembly on the HUG robot with the aim of automating it to the extent possible.
To this end, \gls{ds}-based \glspl{vf} allow for a flexible but relatively \textbf{coarse} automation.
As such, they are well suited to be paired with the very precise visual servoing fixtures under human control.

We thus train two velocity fixtures (one for position and one for rotation guidance) resulting in $N_\mathrm{DS} = 3$ on the data of \cite{muehlbauer2022multiphase}, modeling it using a \gls{kmp} with $\lambda = 0.1$, $\lambda_c = 10$, $\alpha = 0.1$ and $l = 0.03$ based on a \gls{gmm} model with each $M=5$ Gaussians.
The velocity fixtures bring the subsystem close to the assembly position where the visual servoing fixture ($N_\mathrm{VS} = 1$) taken from \cite{muehlbauer2024probabilistic} using in-hand camera measurements ensures precise alignment.
\Cref{fig:eval:hug_fixtures} shows the full set of fixtures.

At the end effector pose visualized in \Cref{fig:eval:hug_fixtures}, the covariance of the visual servoing fixture is, for the first time in the experiment, smaller than the covariance of the \gls{ds}-based fixture.
Thus, the velocity output of the latter (yellow arrow) is not guiding the robot anymore.
Instead, only the visual servoing fixture is active, reproducing the assistive behaviour introduced in \cite{muehlbauer2024probabilistic}, but combining it with the increased flexibility of the novel \gls{ds}-based fixture.
The force introduced by a \textbf{coarse} fixture with high covariance was found to be very small therein, so also in our setup, precise telemanipulation using the \textbf{very precise} visual servoing fixture is not impaired.
Notably, this enables the robot to start anywhere in its workspace, eliminating the need for initial alignment.
\Cref{sec:appendix:teleop} describes the teleoperation coupling used in the experiment.

\begin{figure}
	\centering
	\annotategraphicsmulti{
	    \includegraphics[width=0.78\columnwidth,trim={110 0 70 5},clip,page=5]{bilder.pdf}
	}{
    }
	\caption{Semi-automated CubeSat subsystem assembly \cite{muehlbauer2022multiphase,muehlbauer2024probabilistic,muehlbauer2024ai}: A \gls{ds}-based fixture takes the operator in the vicinity of the insertion pose where the manual visual servoing fixture takes over. This enables a collaborative automation of the first phase with \textbf{coarse} precision while the \textbf{very precise} visual servoing fixture with human corrections enables a successful connector insertion.}
	\label{fig:eval:hug_fixtures}
\end{figure}
\begin{figure}
	\centering
	\annotategraphicsmulti{
	    \includegraphics[width=\columnwidth,trim={0 115 0 15},clip,page=6]{bilder.pdf}
	}{
        \node at (0.68,0.22) {H1};
        \node at (0.31,0.24) {H2};
        \node at (0.45,0.7) {DS};
    }
	\caption{Combination of \gls{ds} and visual servoing fixtures for transporting test tubes from a linear holder \protect\lref{H1} on the right to a circular holder \protect\lref{H2} on the left. The operator is supported by a visual servoing fixture at each holder as well as a dynamical system \protect\lref{DS} aiding the transportation between both.}
	\label{fig:eval:test_tubes}
\end{figure}

In another experiment, we move test tubes from a linear holder \lref{H1} to a cylindrical holder \lref{H2}, showing the extension of the visual servoing fixture to the cylindrical $\mathcal{M}_2$, visualized in \Cref{fig:eval:test_tubes} with $N_\mathrm{VS} = 2$.
We simulate visual measurements for all tube holders with $\bm{\Sigma} = 2.25 \times 10^{-6} \cdot \bm{I}_6$ (\lref{H1}) and $\bm{\Sigma} = 2.5 \times 10^{-7} \cdot \bm{I}_6$ (\lref{H2}).
For the visual servoing fixture in $\mathcal{M}_1$, we use the length scale $l = (0.006, 0.006, 0.2, 0, 0, 0)$ and a deadzone of \SI{5}{\milli\metre} in the $xy$-plane.
For the visual servoing fixture in $\mathcal{M}_2$, we set the length scale $l = (0.1, 0.05, 0.2, 0, 0, 0)$ and use a deadzone of \SI{0.2}{\radian} along the angular \gls{dof}.
For the \gls{ds} ($N_\mathrm{DS} = 3$), we again use a \gls{kmp} based on a \gls{gmm} with $M=5$ Gaussians with hyperparameters $\lambda = 0.05$, $\lambda_c = 10$, $\alpha = 0.1$, $h = 1$ and $l = 0.1$ for positions respectively $0.03$ for orientations.
This setup allows an operator to easily choose a test tube to pick up through guidance from a visual servoing fixture, transport it to the other holder while the velocity fixture supports with a velocity field and by keeping the tube upright where the other visual servoing fixture eases the placement.

\begin{table}
\scriptsize
\centering
\caption{\label{tab:evaluation:tubes}\centering \scshape Tube experiment: quantitative evaluation (M $\pm$ SD).}
\begin{tabular}{@{}c|ccc@{}}
Metric & With Arbitration & Without Arbitration & Wilcoxon signed rank\\
\hline
\textit{Time} & \textbf{\SI[detect-all=true]{23.98 \pm 4.15}{\second}} & \SI{28.10 \pm 11.22}{\second} & $W=1$, $\mathbf{p < 0.1}$\\
\textit{Measured Forces} & \textbf{\SI[detect-all=true]{15.28 \pm 1.90}{\newton}} & \SI{19.70 \pm 3.06}{\newton} & $W=0$, $\mathbf{p < 0.05}$\\
\textit{Perceived Forces} & $\mathbf{5.8 \pm 1.5}$ & $6.6 \pm 2.0$ & $W=6$, n.s.\\
\textit{NASA TLX} & $\mathbf{3.86 \pm 1.3}$ & $6.8 \pm 1.6$ & $W=0$, $\mathbf{p < 0.05}$\\
\textit{SUS} & $\mathbf{85.0 \pm 14.7}$ & $66.3 \pm 14.0$ & $W=1$, $\mathbf{p < 0.1}$
\end{tabular}
\end{table}

In the supplementary video, we show this experiment both with arbitration (\Cref{sec:problem:arbitration}) enabled and without arbitration, adding all fixture wrenches with equal weight.
It can be observed that motions are smoother with the arbitration enabled; furthermore, interaction with the cylindrical visual servoing \gls{vf} is more intuitive.
We observe a success rate of \SI{100}{\percent} in both cases and summarize quantitative findings in \Cref{tab:evaluation:tubes}.
We find significant differences for measured forces and the NASA TLX score and a non-significant trend for completion time and the SUS metric.
A possible explanation for both trends is that the arbitration mainly influences the short transition period between fixtures, thus only causing a small effect.
The force ripples in the \gls{ds}-based fixture and high forces for the circular visual servoing \gls{vf} used for placing the test tube, caused by the missing arbitration, also do not lead to significant differences in the perceived forces.
Both are, however, criticized in the answers to the semi-structured questions.
Combined with the significant results, this underlines the necessity and effectiveness of our proposed arbitration scheme.

\subsection{Semi-Automated Combination of All Fixtures}
\label{sec:evaluation:all_fixtures_sara}
Finally, we combine position-based, velocity and visual servoing fixtures ($N_\mathrm{PB} = 1$, $N_\mathrm{DS} = 3$, $N_\mathrm{VS} = 1$; \Cref{fig:evaluation:all_fixtures_sara}).
The task is to move a bottle \lref{B} in the crate \lref{C} with a position-based trajectory fixture for picking the bottle, a velocity fixture for transporting it to the crate and a visual servoing fixture enabling the selection from multiple placement locations.

From $3$ demonstrations of picking up the bottle with approaches from different angles we learn a trajectory fixture supported by a \gls{gmm} with $M=2$ Gaussians on the cylindrical manifold $\mathcal{M}_2$ centered inside the bottle.
The precision and stiffness matrices at the robot configuration of \Cref{fig:evaluation:all_fixtures_sara}, using $\lambda_\mathrm{rot}^- = 100$, $\lambda_\mathrm{rot}^+ = 500$, $\lambda_\mathrm{trans}^- = 100$, $\lambda_\mathrm{trans}^+ = 500$, $k_\mathrm{trans,nom} = 300$ and $k_\mathrm{rot,nom} = 100$, evaluate to
\begin{equation}
\underbrace{\scriptsize
\begingroup
\setlength\arraycolsep{2pt}
		\begin{bmatrix}
			\num{9} & \num{10} & \num{7} & \num{-12} & \num{13} & \num{32}\\
			\num{10} & \num{760} & \num{-17} & \num{80} & \num{18} & \num{49}\\
			\num{7} & \num{-17} & \num{990} & \num{-8} & \num{28} & *\\
			\num{-12} & \num{80} & \num{-8} & \num{720} & \num{96} & \num{-2}\\
			\num{13} & \num{18} & \num{28} & \num{96} & \num{340} & \num{37}\\
			\num{32} & \num{49} & * & \num{-2} & \num{37} & \num{830}
		\end{bmatrix}.
\endgroup
}_{\bm{P}}
\quad
\underbrace{\scriptsize
\begingroup
\setlength\arraycolsep{2pt}
		\begin{bmatrix}
			* & \num{4.1} & \num{2.3} & * & * & *\\
			\num{4.1} & \num{300} & * & \num{31} & \num{7.2} & \num{19}\\
			\num{2.3} & * & \num{300} & \num{-1.9} & \num{8.7} & *\\
			* & \num{31} & \num{-1.9} & \num{99} & \num{14} & \num{2.6}\\
			* & \num{7.2} & \num{8.7} & \num{14} & \num{52} & \num{-1.7}\\
			* & \num{19} & * & \num{2.6} & \num{-1.7} & \num{100}
		\end{bmatrix}
\endgroup
}_{\bm{K}}\nonumber
\end{equation}
with $|*| < 1$.
We thus obtain a very low stiffness for the angular \gls{dof}, allowing the bottle to be picked up from multiple angles (\Cref{fig:evaluation:all_fixtures_sara}).
Combined with the covariance-aware attractor calculation \eqref{eq:pb_fixture:l0ee} and \eqref{eq:pb_fixture:mahalanobis_projection}, the robot can be moved in the plane while always pointing the gripper towards the bottle.

Using a covariance adaptation using the Mahalanobis distance (\Cref{sec:auto_position_based:cov_adaptation}) with a maximum distance value of $5$ for this unitless distance, the fixture remains active for larger displacements around the rotational \gls{dof} (up to \SI{41}{\degree}) than for other \glspl{dof} (up to \SI{7.1}{\centi\metre}) for the given precision.
Once the bottle has been picked up, the operator can easily escape this fixture by achieving a displacement larger than this threshold.

\begin{figure}
	\begin{subfigure}{0.49\columnwidth}
        	\annotategraphicsmulti{
	        \includegraphics[width=\columnwidth,trim={130 50 210 50},clip,page=7]{bilder.pdf}
        	}{
            \node at (0.41,0.35) {B};
            \node at (0.35,0.5) {C};
	    }
    \end{subfigure}
    \begin{subfigure}{0.49\columnwidth}
	    \annotategraphicsmulti{
            \includegraphics[width=\columnwidth,trim={230 50 110 50},clip,page=8]{bilder.pdf}
	    }{
            \node at (0.68,0.33) {B};
            \node at (0.28,0.54) {C};
        }
    \end{subfigure}
	\caption{\label{fig:evaluation:all_fixtures_sara} Picking up a bottle~\protect\lref{B} for moving it into a crate~\protect\lref{C} using position-based trajectory, velocity and visual servoing fixtures. The rotationally symmetric bottle can be picked up from multiple angles thanks to a probabilistic trajectory fixture on $\mathcal{M}_2$  with variable stiffness (left). After~\protect\lref{B} has been picked up, the velocity fixture takes over, fusing a \textit{stabilizing} with the learned policy (right). This fixture aids the operator to move to~\protect\lref{C} where a probabilistic visual servoing \gls{vf} enables selection of placement position.}
\end{figure}

After leaving the position-based \gls{vf}, the \gls{ds}-based \gls{vf} takes over.
Learned from $5$ demonstrations, it is encoded in a \gls{kmp} based on a \gls{gmm} with $M=5$ Gaussians with hyperparameters $\lambda = 0.1$, $\lambda_c = 10$, $\alpha = 0.1$, $h = 1$ and $l=0.1$ for the position velocity and $l = 0.03$ for the orientation velocity field.
This \gls{ds} moves the operator towards the visual servoing \gls{vf} while keeping the bottle upright.

\begin{figure}
	\centering
	\annotategraphicsmulti{
	    \includegraphics[width=\columnwidth,trim={0 0 0 0},clip]{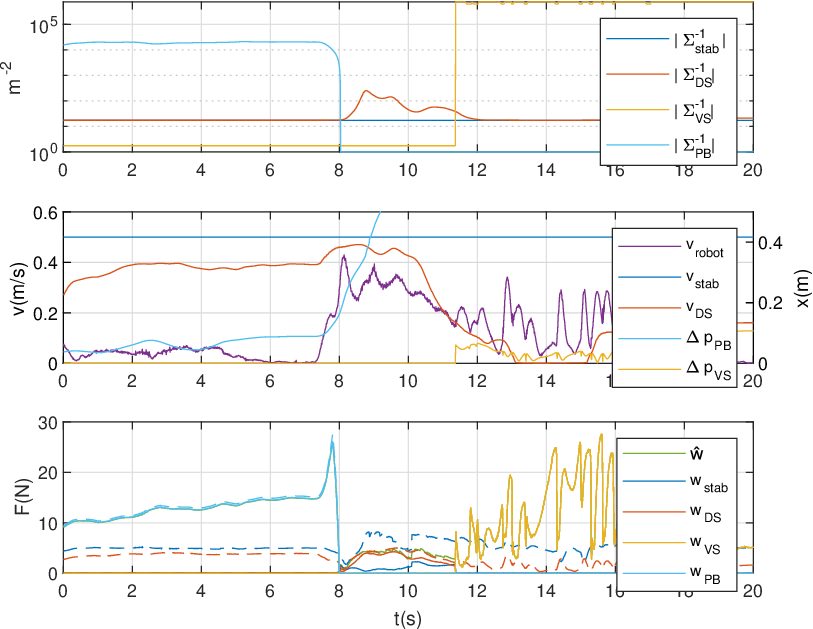}
	}{
    }
	\caption{\label{fig:evaluation:all_fixtures_sara_plot}  Norm of precisions (\textbf{top}), target velocities and position offset (\textbf{middle}), raw (dashed, $\bm{w}_{\mathrm{VF},i}$) and arbitrated (solid lines, $\hat{\bm{\Sigma}}_{\mathrm{VF}} \bm{\Sigma}^{-1}_{\mathrm{VF},i} \bm{w}_{\mathrm{VF},i}$ resp. $\hat{\bm{w}}$, see \eqref{eq:vf_covariance_arbitration}) fixture forces (\textbf{bottom}) of the \textit{stabilizing policy}, \gls{ds}-based, trajectory and visual servoing fixtures used for manipulating the bottle.}
\end{figure}

We simulate visual measurements for $M = 20$ crate positions with $\bm{\Sigma} = 2.25 \times 10^{-1} \cdot \bm{I}_6$, using a deadzone of \SI{5}{\milli\metre} in the $xy$ plane and $l = (0.006, 0.006, 0.2, 0, 0, 0)$.

\Cref{fig:evaluation:all_fixtures_sara_plot} visualizes covariances, target velocities, position offset as well as raw and arbitrated forces summing to $\hat{\bm{w}}$ \eqref{eq:vf_covariance_arbitration} of the exemplary execution from the supplementary video.

\begin{table}
\scriptsize
\centering
\caption{\label{tab:evaluation:cup}\centering \scshape Bottle experiment: quantitative evaluation (M $\pm$ SD).}
\begin{tabular}{@{}c|ccc@{}}
Metric & With Arbitration & Without Arbitration & Wilcoxon signed rank\\
\hline
\textit{Time} & \textbf{\SI[detect-all=true]{10.76 \pm 2.92}{\second}} & \SI{12.16 \pm 4.41}{\second} & $W=6$, n.s.\\
\textit{Measured Forces} & \textbf{\SI[detect-all=true]{6.18 \pm 1.19}{\newton}} & \SI{10.81 \pm 2.32}{\newton} & $W=0$, $\mathbf{p < 0.05}$\\
\textit{Perceived Forces} & $6.7 \pm 1.1$ & $\mathbf{6.0 \pm 0.8}$  & $W=5.5$, n.s.\\
\textit{NASA TLX} & $\mathbf{4.7 \pm 1.4}$ & $\mathbf{4.7 \pm 1.5}$ & $W=9.5$, n.s.\\
\textit{SUS} & $\mathbf{89.0 \pm 6.6}$ & $87.9 \pm 10.0$ & $W=7$, n.s.
\end{tabular}
\end{table}

We ask the expert users to perform the task of picking the bottle and placing it in a predefined position in the crate using our fixtures.
For placing, users had to drop the bottle from a distance of \SI{9}{\centi\metre} above the holder inside the crate; to succeed, a maximum offset of \SI{3}{\milli\metre} to each side must be achieved, otherwise, the bottle topples instead of landing in its target position.
After familiarization with the system, users performed the task first with arbitration and then with an equal-weight addition of fixtures four times each, with either no obstacle, an obstacle close to the picking position, an obstacle on the way to the crate, or both obstacles.
We systematically vary the trial order in a Latin square design.
Over all $24$ trials in each method, users succeeded in placing the bottle with a success rate of \SI{87.5}{\percent} with and \SI{70.8}{\percent} without arbitration.

We summarize quantitative findings and a pairwise statistical evaluation in \Cref{tab:evaluation:cup}.
Note that this evaluation only contains $23$ of the $24$ conducted trials due to a failed recording.
Statistically significant differences are observed for the measured forces.
We attribute non-significant results for other metrics to the fact that in this setting the fixtures themselves are already very helpful.
This is, e.g., also reflected in very similar values of the NASA TLX and SUS scores.
In the semi-structured interview, users reported a close-to-optimal guidance - except for force ripples in the case of no arbitration - with no notable transitions, but highlighted that the forces when switching between crate positions were too high.
This is also reflected in a higher than the average physical demand TLX score ($7.2 \pm 3.9$ with and $6.0 \pm 2.8$ without arbitration).
This shortcoming is easily corrected by lowering the stiffness of the visual servoing \gls{vf}, as confirmed via an additional trial after the experiment.

\section{Discussion}
\label{sec:discussion}
\subsection[Dynamical Systems based Virtual Fixtures]{Dynamical-Systems-based \Acrlongpl{vf}}
\label{sec:discussion:dynamical_fixtures}
\glspl{vf} based on \glspl{ds} as evaluated also in combination with other fixtures in \Cref{sec:evaluation:dynamic_vf,sec:evaluation:dynamic_pb_robograv,sec:evaluation:dynamic_vs_hug,sec:evaluation:all_fixtures_sara} allow for a collaborative automation.
By default, the task is being performed autonomously while human interaction is always possible as can be seen in the supplementary video.

A challenging problem in \gls{ds}-based \glspl{vf} is the need to find the right hyperparameters both for the learned as well as for the \textit{stabilizing policy}.
When using a \gls{gp} for modeling the \gls{ds}, it is especially important to ensure that its covariance output fits the other fixtures present, i.e., it should not output an overly low covariance when another, better suited fixture can take over.
On the other hand, its uncertainty should also not be too high as it would not take any effect otherwise.
This modeling is greatly simplified through the use of \gls{kmp}-based policies, where the underlying \gls{gmm} already models the covariance appropriately.
This leaves the covariance tuning to the constant covariance $\bm{\Sigma}_\mathrm{stab}$ of the \textit{stabilizing policy}.
Future work should consider approaches to automatically select viable hyperparameters in the fixture's kernel as well as for this constant covariance value from demonstrations.

A limitation of the current formulation is that only a global, but no per-\gls{dof} epistemic uncertainty estimate of the velocity policy is available through the kernels \eqref{eq:dynamic_fixtures:rbf}.
This was mitigated by using different policies for positional and orientational control; however, as with the coupled stiffness (\Cref{sec:problem:variable_impedance}), a policy unifying all \glspl{dof} would be desirable.
Such a policy could then, through the arbitration \eqref{eq:vf_covariance_arbitration}, progress some \glspl{dof} through the \gls{ds} evolution while the remaining \glspl{dof} would be brought back to the demonstration data by the \textit{stabilizing policy}.
Such behavior is especially relevant in \glspl{vf}, as perturbations induced by the operator are a desired property of the overall system.

\subsection{Probabilistic Policy Arbitration via Products of Experts}
\label{sec:discussion:arbitration}
Previous works have already shown that a probabilistic arbitration scheme allows for an optimal combination of different \glspl{vf} \cite{muehlbauer2024probabilistic}.
Throughout the evaluation, we have seen how this arbitration naturally extends to other types of \glspl{vf} and manifolds and is essential for combining learned and \textit{stabilizing} policies.
One is thus free to choose the best \gls{vf} representation for a specific task phase without manually designing transitions, highlighting the advantages of our fully probabilistic \gls{vf} formulation.
One key distinction to previous works is that our formulation supports different manifolds $\mathcal{M}$ by transforming the wrench to a common representation in the cotangent space of $\mathbb{R}^3 \times \mathcal{S}^3$.
Performing arbitration in wrench space, different types of \glspl{vf} are naturally supported.
This makes it possible to easily fuse fixtures calculating an attractor point, a target velocity or directly a wrench in a unified formulation which in turn allows us to model guiding behavior for each portion of a task using the best available fixture representation.

\subsection{Variable Stiffness}
\label{sec:discussion:stiffness}
Previous works \cite{muehlbauer2024probabilistic,abifarraj2017learning} either considered diagonal or block-diagonal stiffness matrices.
While such formulations are well suited when high covariance only appears within positions or orientations, block-diagonal stiffness matrices fail when a lower stiffness is required along a coupled \gls{dof} as in \Cref{sec:evaluation:coupled_stiffness}.
To the best of our knowledge, our approach is the first to enable such coupled variable stiffness.
In this experiment, our approach achieves to model a stiffness that makes the robot's end effector follow the geodesic between two detections.
Our approach furthermore provides reasonable stiffness values for the precision matrices observed in all experiments.
This underlines that the proposed approach is suitable for the generation of stiffness matrices from arbitrary precision matrices.

\subsection{Comparison with Potential Fields}
\label{sec:discussion:potential_fields}
Both the position-based trajectory (\Cref{sec:auto_position_based}) and the geometric visual servoing fixture (\Cref{sec:vs_fixture}) can be seen as potential fields vanishing at the attractor pose or trajectory, shaped by the stiffness matrix derived in \Cref{sec:problem:variable_impedance}.
Our work is comparable to artificial potential fields such as those learned by \cite{khansari2015learning} and used as repulsive \cite{kastritsi2022passive} as well as attractive \cite{papageorgiou2020passive} constraints.
Note, however, that those works do not consider the arbitration of different constraints as done by our approach (\Cref{sec:problem:arbitration}).
Furthermore, potential fields cannot model recurring motions as our \gls{ds}-based fixtures can (\Cref{sec:dynamic_fixtures}; evaluation: \Cref{sec:evaluation:dynamic_vf}  (\Cref{fig:eval:dynamic_robograv,fig:evaluation:ds_flight})).

\section{Conclusion}
In this work we introduce a unified, probabilistic \Acrlong{vf} framework providing different types of assistance to operators -- particularly \textbf{coarse}, \textbf{precise}, and \textbf{very precise} guidance -- where each type of fixture can either be manually defined or learned from human demonstrations.
To address a gap in the literature -- namely, the limited attention given to learned virtual fixtures that actively support task progression -- we propose a novel, uncertainty-aware \Acrlong{ds}-based \Acrlong{vf} formulation enabling flexible task automation while keeping the operator in control.
We further introduce geometry-awareness in shared control through object-specific coordinate systems, including Cartesian, cylindrical and spherical frames.
Combined with a novel variable impedance formulation -- which robustly captures correlations between \glspl{dof} -- our framework brings together the different fixture types using a \textit{product of experts} approach, enabling a principled and uncertainty-aware fusion of assistance commands.

We have validated that the approach can be readily applied across diverse use cases, thanks to its ease of programming and flexibility with respect to fixture representations, input modalities, and uncertainty sources.
The evaluation with $6$ expert users suggests high usability of our method, paired with low workload for human operators.
While we demonstrated its use in factory automation and space scenarios, and in an initial experiment with expert users, we believe that the approach is also well-suited for medical and personal assistance robotics, which we plan to investigate in future work.
In this regard, a full user study should be conducted to precisely validate the effectiveness of our approach.
We furthermore envision developing methods to interactively modify position-based fixtures, allowing an operator to modify them adaptively based on novel task needs \cite{quere2024probabilistic}.
To ease finding working hyperparameters for the geometric visual servoing fixture, methods for automatic tuning should be investigated.
Finally, to enable the use of the proposed framework in high-latency applications such as on-orbit servicing, we plan to integrate it with controllers that can provide stable force feedback despite such delays.


{}


\bibliographystyle{./style/IEEEtran}
\bibliography{literatur}

\begin{appendix}

\section{Supplementary Derivations}
\subsection{Key Notations and Hyperparameter Tuning}
\label{sec:appendix:key_notations}
\Cref{tab:notation} summarizes the key notations of our framework.
Parameters introduced in this work can be tuned as follows:
The choice of stiffness eigenvalues $\lambda^-$ / $\lambda^+$ depends on the desired guidance behavior -- we suggest observing the decomposition in \Cref{sec:problem:variable_impedance} for a task and set $\lambda^+$ to the minimum value for which full stiffness is desired and $\lambda^-$ to the maximum value for which the operator should be able to move freely.
The stabilizing policy velocity $\dot{x}_\mathrm{stab}$ can be set to the mean velocity of the learned \glspl{ds} while $\sigma_\mathrm{stab}$ can be set to a value smaller than the \gls{kmp} uncertainty prediction when far from known points, e.g., $0.9 \cdot \alpha$.
$d_\mathrm{min}$ and $d_\mathrm{max}$ can be set to support a small transition region, e.g., to $d_\mathrm{min} = 1$ and $d_\mathrm{max} = 1.2$ to deactivate a fixture once the operator moves outside the $1 \sigma$ region of a fixture.
For the geometric visual servoing fixture, we set $\lambda = 1 \times 10^{-20}$ in our experiments and the deadzone parameters to the desired values. The length scales $\bm{L}$ are then modified until the desired modulation is achieved, followed by the parameters for the additional expert to control activation.
A more principled approach could rely on likelihood optimization from demonstrations \cite{kuehnoel2024visual}.
\begin{table}\caption{\centering \scshape Key notations used in our framework.}
	\centering 
	\begin{tabular}{p{0.25\columnwidth} l p{0.58\columnwidth}}
		\toprule
		\rowcolor{Light0}
		$N  \in \mathbb{N}$ & $\triangleq$ & Number of data points per demonstration\\
		\rowcolor{Light0}
		$N_\mathrm{DS}$, $N_\mathrm{PB}$, $N_\mathrm{VS}$ & $\triangleq$ & Number of fixtures of each type\\
		\rowcolor{Light0}
		$M$ & $\triangleq$ & Number of Gaussian components in a fixture\\

		\rowcolor{Light0}
		$\bm{x}_{\mathrm{VF},i}$ & $\triangleq$ & Attractor of the $i$-th fixture \\
		\rowcolor{Light0}
		$\bm{w}_{\mathrm{VF},i}$ & $\triangleq$ & Wrench of the $i$-th fixture \\
		\rowcolor{Light0}
		$\bm{\Sigma}_{\mathrm{VF},i} = \bm{P}_{\mathrm{VF},i}^{-1}$ & $\triangleq$ & Covariance of the $i$-th fixture \\

		\rowcolor{Light1}
		$\bm{K}_{\mathrm{VF},i}$ & $\triangleq$ & Stiffness matrix of the $i$-th fixture \\
		\rowcolor{Light1}
		$\bm{D}_{\mathrm{VF},i}$ & $\triangleq$ & Damping matrix of the $i$-th fixture \\
		\rowcolor{Light1}
		$\bm{k}_\mathrm{nom}$ & $\triangleq$ & Diagonal nominal stiffness values \\
		\rowcolor{Light1}
		$\lambda_\mathrm{trans}^-$, $\lambda_\mathrm{trans}^+$ & $\triangleq$ & ``High'' and ``low'' trans. stiffness eigenvalues\\
		\rowcolor{Light1}
		$\lambda_\mathrm{rot}^-$, $\lambda_\mathrm{rot}^+$ & $\triangleq$ & ``High'' and ``low'' rot. stiffness eigenvalues\\

		\rowcolor{Light2}
		$\lambda$, $\lambda_c$, $\alpha$ & $\triangleq$ & \gls{kmp} regularization and scaling factors \\
		\rowcolor{Light2}
		$l$ & $\triangleq$ & Length scale for the \Acrshort{rbf} kernel\\
		\rowcolor{Light2}
		$\dot{x}_\mathrm{stab}$, $\sigma_\mathrm{stab}$ & $\triangleq$ & \textit{Stabilizing policy} velocity and covariance\\
		
		\rowcolor{Light3}
		$d_\mathrm{min}$, $d_\mathrm{max}$ & $\triangleq$ & Start / end distance of covariance adaptation\\

		\rowcolor{Light4}
		$\bm{L}$ & $\triangleq$ & Length scale of the visual servoing fixture\\
		\rowcolor{Light4}
		$\gamma$ & $\triangleq$ & Regularization factor\\
		\rowcolor{Light4}
		$\bm{L}_\mathrm{dead}$, $r_\mathrm{dead}$ & $\triangleq$ & Length scale and radius of the deadzone\\
		\rowcolor{Light4}
		$\bm{L}_\mathrm{add}$, $\bm{L}_\mathrm{dead,add}$, $r_\mathrm{dead,add}$ & $\triangleq$ & Length scale and deadzone parameters of the additional expert\\
		\bottomrule
	\end{tabular}
	\label{tab:notation}
\end{table}

\subsection{Variable Impedance Control on $\mathcal{M}_2$ and $\mathcal{M}_3$}
\label{sec:appendix:var_stiffness_rot}
The variable impedance formulation of \Cref{sec:problem:variable_impedance} can also be extended to $\mathcal{M}_2$ and $\mathcal{M}_3$ introduced in \Cref{sec:background:on_manifold}.
For this, we scale the nominal translational stiffness $k_{\mathrm{nom},j}$ by the radius $r$ for the angular \glspl{dof} ($\mathcal{M}_2$: $j = 1$, $\mathcal{M}_3$: $j = 1, 2$):
\begin{equation}
	k_{\mathrm{nom},j}^* = r \cdot k_{\mathrm{nom},j}.
\end{equation}
This ensures a stiffness comparable to the Euclidean case for those \glspl{dof}, avoiding too high stiffness values for $r \ll 1$.
As the calculation of the stiffness values is performed in a coordinate system rotated by $\bm{R}_{\mathrm{diag},i}^\top$, the maximum translational stiffness values for each \gls{dof} in those coordinates has to be limited further s.t. when rotating the stiffness matrix back, the nominal stiffness value is not exceeded.
This can be achieved by limiting the maximum value of the stiffness $\bm{k}_\mathrm{nom,trans}$ denoting the translational ($j=1,2,3$) \glspl{dof} of $\bm{k}_\mathrm{nom}$ in rotated \glspl{dof} through
\begin{align}
	\bm{k}_\mathrm{rot} &= \mathrm{diag}(\bm{R}_{\mathrm{diag},i}^T \mathrm{diag}(\bm{k}_\mathrm{nom,trans}) \bm{R}_{\mathrm{diag},i})\label{eq:stiffness:rotforward}\\
	\bm{k}_\mathrm{rotback} &= \mathrm{diag}(\bm{R}_{\mathrm{diag},i} \mathrm{diag}(\bm{k}_\mathrm{rot}) \bm{R}_{\mathrm{diag},i}^\top)\label{eq:stiffness:rotbackward}\\
	\beta &= \mathrm{max}(\bm{k}_\mathrm{rotback} \oslash \bm{k}_\mathrm{nom,trans})\label{eq:stiffness:factor}\\
	\bm{k}_\mathrm{nom,trans}' &= \beta\bm{k}_\mathrm{rot}
\end{align}
where $\mathrm{diag}$ transforms a vector to a diagonal matrix respectively extracts the diagonal of a matrix and $\oslash$ is the element wise vector division.
In \eqref{eq:stiffness:rotforward}, we take the diagonal elements of $\bm{k}_\mathrm{nom,trans}$ rotated into coordinates of $\bm{P}_{\mathrm{VF},i}'$.
As the maximum values of the translational stiffnesses correspond to $\mathrm{diag}(\bm{k}_\mathrm{rot})$, we can rotate that matrix back and check for potentially increased stiffness values in \eqref{eq:stiffness:factor}.
The maximum value of this increase is then used to scale the translational stiffnesses.

\subsection{Optimal Damping using Double Diagonalization}
\label{sec:appendix:damping}
We design the damping matrix $\bm{D}_i$ using double diagonalization as presented in \cite{albuschaeffer2003cartesianimpedance} with $\zeta = 0.7$, in particular
\begin{equation}
	\bm{D}_i = 2 \zeta \bm{Q} \bm{K}_{i,0}^\frac{1}{2} \bm{Q}^\top
\end{equation}
with mass matrix $\bm{M}_{i,\mathcal{M}} = \bm{Q} \bm{Q}^\top$ and stiffness matrix $\bm{K}_i^* = \bm{Q} \bm{K}_{i,0} \bm{Q}^\top$ where $\bm{K}_{i,0}$ is a diagonal matrix.
Similar to the wrench transformation in the previous section, a Cartesian mass matrix $\bm{M}_\mathrm{cart}$ has to be transformed to the manifold using
\begin{equation}
	\bm{M}_{i,\mathcal{M}} = \left( \bm{J}_{i,\mathcal{M}} \bm{M}_\mathrm{cart}^{-1} \bm{J}_{i,\mathcal{M}}^\top \right)^{-1}.
\end{equation}
To also dampen zero-stiffness \glspl{dof}, we calculate $\bm{K}_i^*$  with scaling factors $s_j$ used in \eqref{eq:variable_stiffness:torque_realization} and \eqref{eq:variable_stiffness:wrench_realization} lower-bounded to a small $\epsilon > 0$, thus ensuring smoothly decaying robot motions.

\subsection{Teleoperation System}
\label{sec:appendix:teleop}
To supply the operator with force feedback, we use the bilateral teleoperation system of \cite{muehlbauer2022multiphase,muehlbauer2024probabilistic}.
We assume that the haptic input device can also be controlled using Cartesian wrenches; in case of a torque-controlled robot, \eqref{eq:robot_torque_control} can be used to compute joint torques.
Using a simple position-computed force architecture that does not require a force-torque sensor at the end effector, the Cartesian wrenches of remote robot $\bm{w}_{\mathrm{ee, rem}}$ and input device $\bm{w}_{\mathrm{ee, inp}}$ are
\begin{align}
	\bm{w}_{\mathrm{ee, rem}} &= \chi \left(\bm{K} \mathrm{Log}_{\bm{x}_{\mathrm{ee}}}^{\mathcal{M}}\!\!\left(\bm{x}_{\mathrm{inp}}\right) + \bm{D} \frac{\mathrm{d}}{\mathrm{d}t}\mathrm{Log}_{\bm{x}_{\mathrm{ee}}}^{\mathcal{M}}\!\!\left(\bm{x}_{\mathrm{inp}}\right)\right) + \bm{w}_{\mathrm{VF}} \nonumber\\
	\bm{w}_{\mathrm{ee, inp}} &= - \chi \bm{Ad}_{\mathrm{ir}} \bm{w}_{\mathrm{ee, rem}}.\label{eq:wrench_remote}
\end{align}
The user receives force feedback from both the environment as well as the \glspl{vf} through the coupling introduced by $\chi$, the adjoint $\bm{Ad}_{\mathrm{ir}}$ transforms wrenches between coordinate systems of the remote robot and input device.
Applying the \gls{vf} wrench to the remote robot, we can support a user while avoiding potential inaccuracies resulting from a teleoperated coupling.

\subsection{KMP with Covariance in Pose Space for DS}
\label{sec:appendix:covariance_pose_space}
First, we approximate the joint distribution between $\bm{x}$ and $\dot{\bm{x}}$ in a \gls{gmm} with $M$ components and use \gls{gmr} to compute the probabilistic reference velocities $\mathcal{N}(\dot{\bm{x}}_\mathrm{DS} | \bm{\mu}_\mathrm{GMR}, \bm{\Sigma}_\mathrm{GMR})$.
Through the averaging of the \gls{gmm}, this results in a smooth velocity field.
As the wrench calculated from this \gls{ds} is later fused with wrenches from other \glspl{vf}, we only use the velocity $\bm{\mu}_\mathrm{GMR}$ resulting from the mixture regression.
The covariance output $\bm{\Sigma}_\mathrm{GMR}$ of the regression is discarded, as it would correspond to a covariance in velocity space which is not well suited for the fusion with fixtures with covariance in pose space.
To obtain a better-suited covariance in pose space, we decompose data points $\bm{\xi}_j$ and the $M$ Gaussians of the \gls{gmm}
\begin{gather}
	\bm{\xi}_j = \begin{bmatrix}
		\bm{x}_j\\
		\dot{\bm{x}}_j
	\end{bmatrix}, \quad
	\bm{\mu}_m = \begin{bmatrix}
		\bm{\mu}_m^{I}\\
		\bm{\mu}_m^{O}
	\end{bmatrix}, \quad
	\bm{\Sigma}_m = \begin{bmatrix}
		\bm{\Sigma}_m^{I} & \bm{\Sigma}_m^{IO}\\
		\bm{\Sigma}_m^{OI} & \bm{\Sigma}_m^{O}
	\end{bmatrix}
\end{gather}
which in turn allows us to write a marginalized \gls{gmm} using the weight factors $\pi_m$ from the joint encoding \eqref{eq:background:probabilistic:joint_gmm_encoding}
\begin{equation}
	p_n(\bm{x}) = \sum_{m=1}^M \pi_m \mathcal{N}(\bm{x} | \bm{\mu}_m^{I}, \bm{\Sigma}_m^{I})\label{eq:ds:input_gmm}
\end{equation}
and to compute the likelihood of each Gaussian generating $\bm{x}$
\begin{equation}
	e_m = \frac{\pi_m \mathcal{N}(\bm{x} | \bm{\mu}_m^{I}, \bm{\Sigma}_m^{I})}{\sum_{m=1}^M \pi_m \mathcal{N}(\bm{x} | \bm{\mu}_m^{I}, \bm{\Sigma}_m^{I})}.
\end{equation}
Using those weighting factors, we compute $\hat{\bm{\Sigma}}$ through a unimodal approximation \eqref{eq:background:probabilistic:gmm_unimodal} of the \gls{gmm} \eqref{eq:ds:input_gmm}.
This results in a probabilistic velocity field $\mathcal{N}(\bm{\mu}_\mathrm{GMR}, \hat{\bm{\Sigma}})$.
We sample this velocity field at a subsampled set of $N_\mathrm{ref}$ poses from the demonstrated trajectories to obtain a probabilistic reference $\{\bm{\mu}_{\mathrm{GMR}, n}, \hat{\bm{\Sigma}}_n \}_{n=1}^{N_\mathrm{ref}}$ which is then encoded into the \gls{kmp}.
Finally, \gls{kmp} mean and covariance predictions \eqref{eq:background:kmp_pred_1}, \eqref{eq:background:kmp_pred_2} provide $\bm{\mu}_\mathrm{DS}$, $\bm{\Sigma}_\mathrm{DS}$ in \eqref{eq:gp_approximate} at each end effector pose $\bm{x}_\mathrm{ee}$.

\subsection{On-Manifold Covariance-aware Interpolation}
\label{sec:appendix:on_manifold_interpolation}
\begin{figure}
\centering
\begin{DIFnomarkup}
\begin{tikzpicture}[scale=0.8]
\begin{axis}[
  axis lines=middle,
  axis line style={-Stealth,very thick},
  xmin=0,xmax=4.8,ymin=0,ymax=4.8,
  xtick distance=1,
  ytick distance=1,
  xlabel=$x$,
  ylabel=$y$,
  xticklabels={,,},
  yticklabels={,,},
  grid=major,
  grid style={thin,densely dotted,black!20},
  axis equal image]
		\draw[-,red] (0,0.2) -- (5,0.2);
		\filldraw[red] (0,0.2) circle[radius=2pt];
		\filldraw[red] (1,0.2) circle[radius=2pt] node[anchor=west,rotate=90]{$\bm{\mu}_{j,\mathrm{cart}}$};
		\filldraw[red] (2,0.2) circle[radius=2pt] node[anchor=north west,rotate=90]{$\bm{\mu}_{j+1,\mathrm{cart}}$};
		\filldraw[red] (3,0.2) circle[radius=2pt] node[anchor=west,rotate=90]{$\bm{\mu}_{j,\mathrm{cyl}}$};
		\filldraw[red] (4,0.2) circle[radius=2pt] node[anchor=west,rotate=90]{$\bm{\mu}_{j+1,\mathrm{cyl}}$};

		\draw[style={thin,dashed,black!50}] circle[radius=4];
		\draw[style={thin,dashed,black!50}] (1.65,0) -- (1.65,4.5);

		\filldraw[black] (1.65,3.65) circle[radius=2pt] node[anchor=south west]{$\bm{x}_{\mathrm{ee}}$};
		\filldraw[black] (1.65,0.2) circle[radius=2pt] node[anchor=south]{$\bm{x}_{\mathrm{cart}}$};
		\filldraw[black] (3.9,0.2) circle[radius=2pt] node[anchor=south east]{$\bm{x}_{\mathrm{cyl}}$};


\end{axis}
\end{tikzpicture}
\end{DIFnomarkup}
\caption{\label{fig:pb_fixture:distance_manifolds}2D projection of $\bm{x}_{\mathrm{ee}}$ and the closest poses $\bm{\mu}_j$ and $\bm{\mu}_{j+1}$ on a trajectory (red points) in $\mathcal{M}_1$ ($\bm{x}_{\mathrm{cart}}$) as well as $\mathcal{M}_23$ ($\bm{x}_{\mathrm{cyl}}$). The $x$ axis through $\bm{x}_{\mathrm{ee}}$ as well as the circle with the same radius on $\mathcal{S}^1 \times \mathbb{R}^2 \times \mathcal{S}^3$ is plotted in dashed gray, note that according to the manifold metric the closest pose is much further to the right than for Cartesian coordinates.}
\end{figure}

In the real-time controller, the two closest poses $\bm{\mu}_j$ and $\bm{\mu}_{j+1}$ of this trajectory according to the on-manifold Mahalanobis distance $d_{\bm{\Sigma_j^{-1}}}^{\mathcal{M}}(\bm{\mu}_n, \bm{x}_\mathrm{ee})$ \eqref{eq:distance_function_onman_weighted}
are extracted.
To guarantee constant runtime, we first extract the $10$ closest poses in a background thread while the actual real time computations only use those $10$ closest poses.
On the Intel Core i5-14600 CPU with the controller running at \SI{8}{\kilo\hertz} on a single core, this allows us to calculate one trajectory fixture at \SI{40}{\percent} and two fixtures at \SI{50}{\percent} core usage of the full controller.
For simplicity, we use only the position part of the pose in the distance function, where we set the last three entries of $\bm{\mu}_j$ and $\bm{x}_\mathrm{ee}$, which correspond to the orientation, to the identity.
We found only using the position to be sufficient in manipulation scenarios.
Adding the orientation to the distance function could help in scenarios where parts of the trajectory pass very close to each other, at the cost of having to tune a scaling factor between position and orientation.
We then perform linear interpolation between both extracted closest poses along the covariance-deformed geodesic on the manifold (\Cref{fig:pb_fixture:distance_manifolds}), using
\begin{gather}
	\Delta\bm{\mu}_{j,j+1} = \mathrm{Log}_{\bm{\mu}_j}^{\mathcal{M}} \bm{\mu}_{j+1}, \quad
	\Delta\bm{\mu}_{j,\mathrm{ee}}  = \mathrm{Log}_{\bm{\mu}_j}^{\mathcal{M}} \bm{x}_{\mathrm{ee}},\label{eq:pb_fixture:l0ee}\\
	\nu = \frac{\Delta\bm{\mu}_{j,\mathrm{ee}}^\top \bm{\Sigma}_j^{-1} \Delta\bm{\mu}_{j,j+1}}{\Delta\bm{\mu}_{j,j+1}^\top \bm{\Sigma}_j^{-1} \Delta\bm{\mu}_{j,j+1}}, \quad
	\bm{x}_{\mathrm{PB}} = \mathrm{Exp}_{\bm{\mu}_j}^{\mathcal{M}} \left(\nu \cdot \Delta\bm{\mu}_{j,j+1}\right)\raisetag{1em}\label{eq:pb_fixture:mahalanobis_projection}
\end{gather}
where $\Delta\bm{\mu}_{j,j+1}$ is the vector between $\bm{\mu}_j$ and $\bm{\mu}_{j+1}$ and $\Delta\bm{\mu}_{j,\mathrm{ee}}$ between $\bm{\mu}_j$ and $\bm{x}_{\mathrm{ee}}$ in tangent space and $\bm{\Sigma}_j^{-1}$ the precision matrix corresponding to $\bm{\mu}_j$.
The interpolation factor $0 \le \nu \le 1$ represents the closeness of $\bm{x}_{\mathrm{ee}}$ to $\bm{\mu}_j$ and $\bm{\mu}_{j+1}$ and $\bm{x}_{\mathrm{PB}}$ the final \gls{vf} pose which is used in \eqref{eq:VF_wrench} to calculate the wrench $\bm{w}_{\mathrm{VF},i}$ associated to the fixture.
Its computation through \eqref{eq:pb_fixture:mahalanobis_projection} takes the Mahalanobis distance between the end effector and the two closest means into account.
As we assume the covariance matrix to only vary slowly between points we set $\bm{\Sigma}_\mathrm{PB} = \bm{\Sigma}_j$ which is associated with $\bm{\mu}_j$; however, interpolation could be performed similarly on the manifold of symmetric positive definite matrices~\cite{calinon2020gaussians}.

The manifold-aware attractor calculation through \eqref{eq:pb_fixture:l0ee} and \eqref{eq:pb_fixture:mahalanobis_projection} ensures that only forces orthogonal to the trajectory are being applied by the fixture and thus the user is in full control of motion along the trajectory.

\subsection{Modified Logarithm Map with Deadzone}
\label{sec:appendix:modified_log}
We calculate $\mathrm{Log}_{\bm{x}_{ee}}^{'\mathcal{M}}\!\!\left( \bm{\mu}_m \right)$ from $\mathrm{Log}_{\bm{x}_{ee}}^{\mathcal{M}}\!\!\left( \bm{\mu}_m \right)$ using the length scale vector $\bm{l}_\mathrm{dead}$ and the crop length $r_\mathrm{crop}$ as follows:
\begin{align}
	\bm{L}_\mathrm{dead} &= \mathrm{diag}(\bm{l}_\mathrm{dead}^{-2})\label{eq:moe:deadzone_first}\\
	r &= \sqrt{d_{\bm{L}_\mathrm{dead}}^{\mathcal{M}}\left(\bm{x}_\mathrm{ee}, \bm{\mu}_m \right)}\label{eq:moe:deadzone_distance}\\
	\bm{d} &= \frac{\mathrm{Log}_{\bm{x}_{ee}}^{\mathcal{M}}\!\!\left( \bm{\mu}_m \right)}{r}\\
	r_\mathrm{crop} &= \mathrm{min}(r, r_\mathrm{dead})\\
	d'_j &= \begin{cases} d_j, & l_{\mathrm{dead},j} > 0\\ 0, & l_{\mathrm{dead},j} = 0 \end{cases}\label{eq:moe:deadzone_some_zero}\\
	\mathrm{Log}_{\bm{x}_{ee}}^{'\mathcal{M}}\!\!\left( \bm{\mu}_m \right)&= \mathrm{Log}_{\bm{x}_{ee}}^{\mathcal{M}}\!\!\left( \bm{\mu}_m \right) - r_\mathrm{crop} \cdot \bm{d}'\label{eq:moe:deadzone_last}
\end{align}
\eqref{eq:moe:deadzone_distance} uses the on-manifold distance \eqref{eq:distance_function_onman_weighted}, \eqref{eq:moe:deadzone_some_zero} ensures that only directions with length vector $>0$ get modified.

\subsection{Coordinate System Conversions}
\label{sec:appendix:coordsys}
\subsubsection{Conversion between $\mathcal{S}^1 \times \mathbb{R}^2 \times \mathcal{S}^3$ and $\mathbb{R}^3 \times \mathcal{S}^3$}
Position quantities can be converted from $\mathcal{M}_1$ to $\mathcal{M}_2$ using
\begin{gather}
	\theta_c = x / r, \quad
	\theta_s = y / r, \quad
	r = \sqrt{x^2 + y^2}, \quad
	z = z\nonumber
\end{gather}
where $\theta_c$ and $\theta_s$ denote the angle $\theta$ around the z axis as a complex number.
For small $r$, the angle $\theta$ is not well-defined.
The quaternion has to be rotated by $-\theta$, specifically
\begin{equation}
	\bm{q}^{\mathcal{M}_2} = \mathrm{Exp}_{\bm{I}}^{\mathcal{S}^3} \begin{bmatrix} 0, 0, -\theta \end{bmatrix}^\top \cdot \bm{q}^{\mathcal{M}_1}.\nonumber
\end{equation}

The inverse conversion can be performed as follows
\begin{gather}
	x = r \cdot \theta_c, \quad
	y = r \cdot \theta_s, \quad
	z = z, \quad
	\bm{q}^{\mathcal{M}_1} = \mathrm{Exp}_{\bm{I}}^{\mathcal{S}^3} \begin{bmatrix} 0\\ 0\\ \theta \end{bmatrix} \cdot \bm{q}^{\mathcal{M}_2}\nonumber
\end{gather}
with $\theta = \mathrm{arctan2}(\theta_s, \theta_c)$ where $\mathrm{arctan2}$ is the modified arc tangent mapping to the full circle $[-\pi, \pi]$.

\subsubsection{Conversion between $\mathcal{S}^2 \times \mathbb{R}^1 \times \mathcal{S}^3$ and $\mathbb{R}^3 \times \mathcal{S}^3$}
Position quantities can be converted from $\mathcal{M}_1$ to $\mathcal{M}_3$ using
\begin{gather}
	s_0 = x / r, \quad
	s_1 = y / r, \quad
	s_2 = z / r, \quad
	r = \sqrt{x^2 + y^2 + z^2}\nonumber
\end{gather}
For very small $r$, values $s_i$ are not well-defined.
The quaternion expressing the orientation has to be rotated by the inverse of the rotation $\bm{q}_\mathrm{align}$ aligning $[0,0,1]$ with $[s_0, s_1, s_2]$:
\begin{equation}
	\bm{q}^{\mathcal{M}_3} = \bm{q}_\mathrm{align}^{-1} \cdot \bm{q}^{\mathcal{M}_1}.\nonumber
\end{equation}

The inverse conversion can be performed as follows
\begin{gather}
	x = r \cdot s_0, \quad
	y = r \cdot s_1, \quad
	z = r \cdot s_2, \quad
	\bm{q}^{\mathcal{M}_1} = \bm{q}_\mathrm{align} \cdot \bm{q}^{\mathcal{M}_3}.\nonumber
\end{gather}

\subsection{Manifold Jacobians}
\label{sec:appendix:jacobians}
The manifold Jacobian $\bm{J}_{\mathcal{M}} = \frac{\partial \bm{x}_{\mathrm{ee},i,\mathcal{M}}}{\partial \bm{x}_{\mathrm{ee},\mathbb{R}^3 \times \mathcal{S}^3}}$  relating $\dot{\bm{x}}^\mathcal{M} = \bm{J}_{\mathcal{M}} \dot{\bm{x}}^{\mathbb{R}^3 \times \mathcal{S}^3}$ and $\bm{w}^{\mathbb{R}^3 \times \mathcal{S}^3} = \bm{J}_{\mathcal{M}}^\top \bm{w}^\mathcal{M}$ is derived as in \cite{dyck2022impedance}
\begin{equation}
	\bm{J}_\mathcal{M} =
	\begin{bmatrix}
		\bm{J}_{\bm{px}} & \bm{0}\\
		\bm{J}_{\bm{\omega x}} & \bm{J}_{\bm{\omega \omega}}
	\end{bmatrix}
\end{equation}
where $\bm{J}_{\bm{px}}$ denotes position quantities and $\bm{J}_{\bm{\omega x}}$ the coupling between position and orientation.
Both are given for $\mathcal{M}_2$ and $\mathcal{M}_3$ in the following.
We furthermore have $\bm{J}_{\bm{\omega \omega}} = \bm{I}$ for $\mathcal{M}$ as the axes of the orientation always coincide.

\subsubsection{$\mathcal{M}_2$ Jacobian}
Cartesian $x$ and $y$ coordinates have an influence on the angular \gls{dof} of the manifold and on $r$.
As the $y$ axis of the unit orientation points in direction of increasing $r$, Cartesian $x$ / $y$ and rotational velocities are coupled
\begin{gather}
	\underbrace{\begin{bmatrix}
		\frac{-y}{x^2 + y^2} & \frac{x}{x^2 + y^2} & 0\\
		\frac{x}{\sqrt{x^2 + y^2}} & \frac{y}{\sqrt{x^2 + y^2}} & 0\\
		0 & 0 & 1
	\end{bmatrix}}_{\bm{J}_{\bm{px}}}
	\quad
	\underbrace{\bm{R}^\top
	\begin{bmatrix}
		0 & 0 & 0\\
		0 & 0 & 0\\
		\frac{y}{x^2 + y^2} & \frac{-x}{x^2 + y^2} & 0
	\end{bmatrix}}_{\bm{J}_{\bm{\omega x}}}\nonumber
\end{gather}
where $\bm{R}^\top$ denotes the orientation in manifold coordinates.

\subsubsection{$\mathcal{M}_3$ Jacobian}
$x$, $y$, $z$ contribute to angular and $r$ \glspl{dof}
\begin{gather}
	\underbrace{\begin{bmatrix}
		\frac{1}{r} & 0 & 0\\
		0 & \frac{1}{r} & 0\\
		0 & 0 & 1
	\end{bmatrix}
	\cdot \bm{R}_\mathrm{align}^\top}_{\bm{J}_{\bm{px}}}
	\quad
	\underbrace{\bm{R}^\top \cdot
	\begin{bmatrix}
		0 & \frac{1}{r} & 0\\
		-\frac{1}{r} & 0 & 0\\
		0 & 0 & 0
	\end{bmatrix}
	\cdot \bm{R}_\mathrm{align}^\top}_{\bm{J}_{\bm{\omega x}}}\nonumber
\end{gather}
where $\bm{R}_\mathrm{align}$ is the rotation matrix equivalent to $\bm{q}_\mathrm{align}$ and $\bm{R}^\top$ denotes the orientation in manifold coordinates.
\end{appendix}

\clearpage
\raggedbottom
\begin{IEEEbiography}[{\biophoto{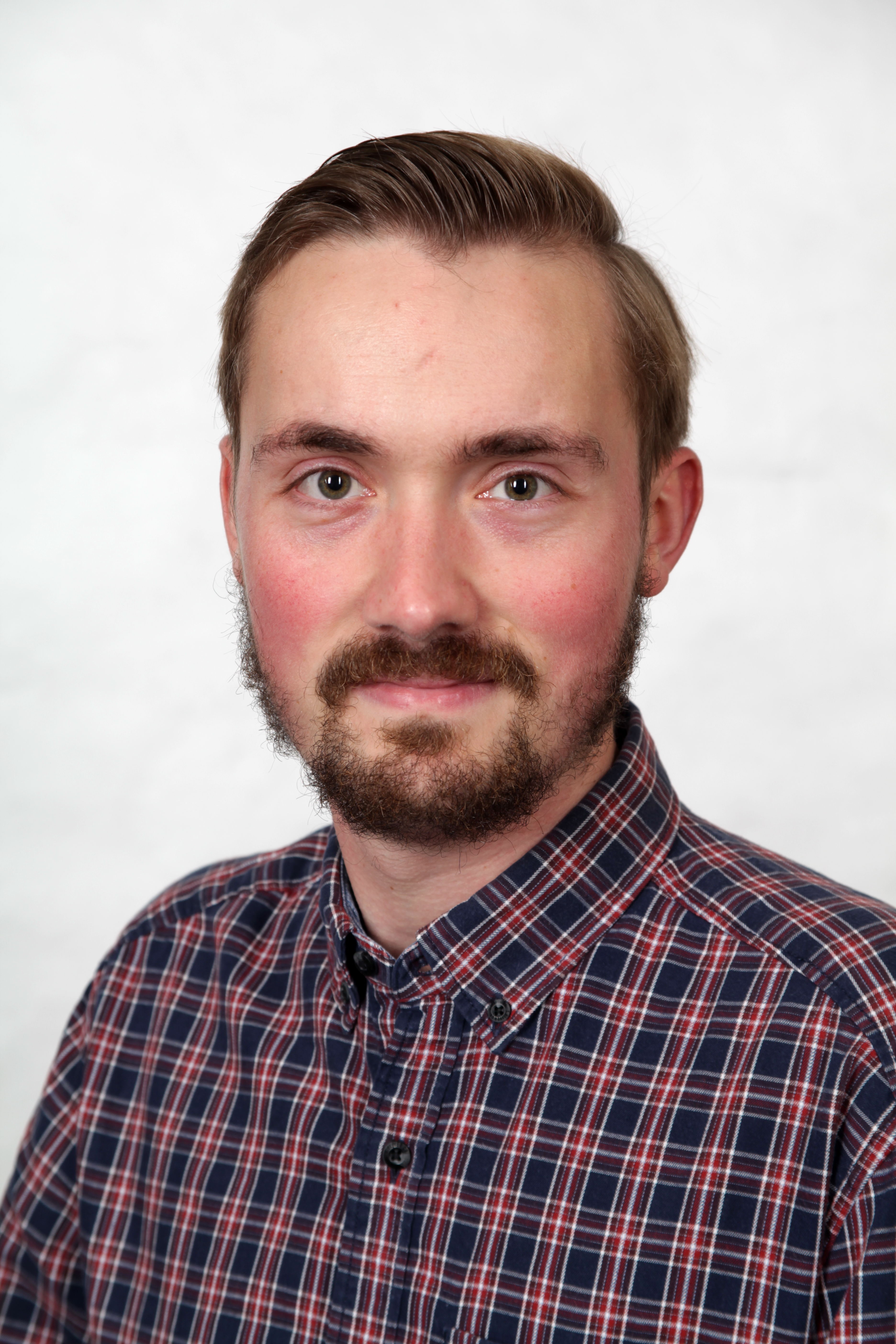}}]{Maximilian Mühlbauer}
received his B.Sc. and M.Sc. degrees in Mechanical Engineering respectively Robotics from the Technical University of Munich (TUM).
He is a research scientist in the department of Cognitive Robotics at the Institute of Robotics and Mechatronics at the German Aerospace Center (DLR) and a Ph.D. student at the Technical University of Munich.
His research focuses on probabilistic machine learning and control methods for human-robot collaboration with applications in space, medical and industrial robotics.
\end{IEEEbiography}

\begin{IEEEbiography}[{\biophoto{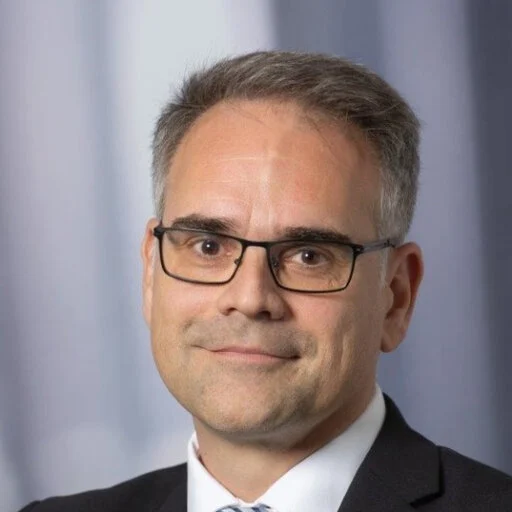}}]{Bernhard Weber}
received his Dipl.-Psych. and PhD degree at the University of Würzburg, Germany, in 2004 and 2008, respectively.
From 2008–2010, he was with the German Aerospace Center (DLR), Institute of Flight Guidance, Brunswick, Germany and since 2010 at the DLR Institute of Robotics and Mechatronics, Wessling, Germany, as a human factors expert.
His main research interests are human factors in telerobotic systems, evaluation of haptic interaction technology, and sensorimotor performance under conditions of microgravity.
\end{IEEEbiography}

\begin{IEEEbiography}[{\biophoto{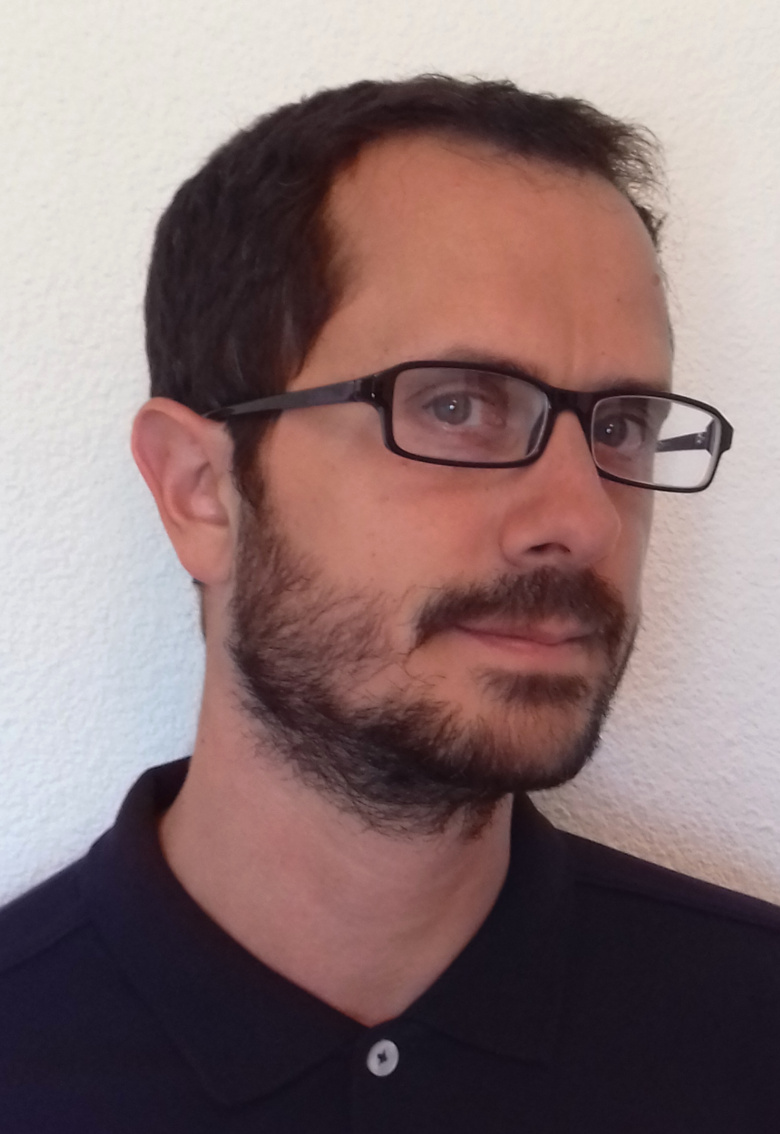}}]{Sylvain Calinon}
received the Ph.D. degree in robotics from the École Polytechnique Fedérale de Lausanne (EPFL), Lausanne, Switzerland, in 2007.
He is currently a Senior Research Scientist with the Idiap Research Institute, Martigny, Switzerland, and a Lecturer with the EPFL.
From 2009 to 2014, he was a Team Leader with the Italian Institute of Technology, Genoa, Italy.
From 2007 to 2009, he was a Postdoc with EPFL.
His research interests include human–robot collaboration, robot learning, robot manipulation skills, geometry and model-based optimization.
\end{IEEEbiography}

\begin{IEEEbiography}[{\biophoto{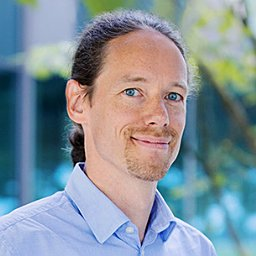}}]{Freek Stulp}
received the doctorate degree in computer science from the Technical University of Munich, Munich, Germany, in 2007.
He is currently the Head of the Department of Cognitive Robotics, Institute of Robotics and Mechatronics, German Aerospace Center.
Previously, he was an Assistant Professor with the École Nationale Supérieure de Techniques Avancées.
\end{IEEEbiography}

\vfill{}
\newpage

\begin{IEEEbiography}[{\biophoto{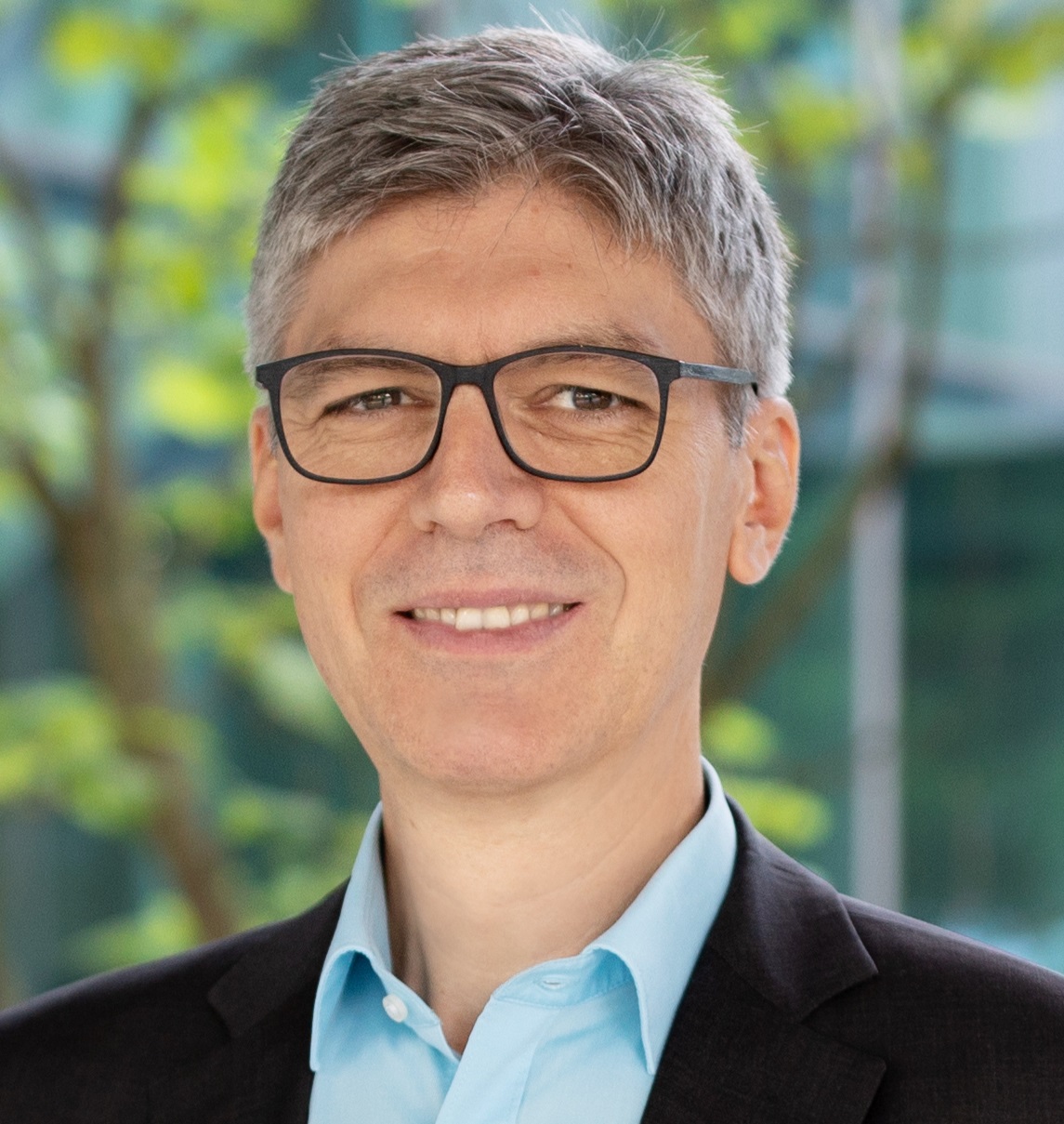}}]{Alin Albu-Schäffer}
received his M.S. in electrical engineering from the Technical University of Timisoara, Romania in 1993 and his Ph.D. in automatic control from the Technical University of Munich in 2002.
Since 2012 he is the head of the Institute of Robotics and Mechatronics at the German Aerospace Center (DLR).
Moreover, he is a professor at the Technical University of Munich, holding the Chair for "Sensor Based Robotic Systems and Intelligent Assistance Systems" at the School of Computation, Information and Technology.
His personal research interests include robot design, modeling and control, nonlinear control, flexible joint and variable compliance robots, impedance and force control, physical human-robot interaction, bio-inspired robot design and control.
He received several awards, including the IEEE King-Sun Fu Best Paper Award of the Transactions on Robotics in 2012 and 2014; several ICRA and IROS Best Paper Awards as well as the DLR Science Award.
He was strongly involved in the development of the DLR light-weight robot and its commercialization through technology transfer to KUKA.
He is the coordinator of euROBIN, the European network of excellence on intelligent robotics, IEEE Fellow and RAS-member.
\end{IEEEbiography}

\begin{IEEEbiography}[{\biophoto{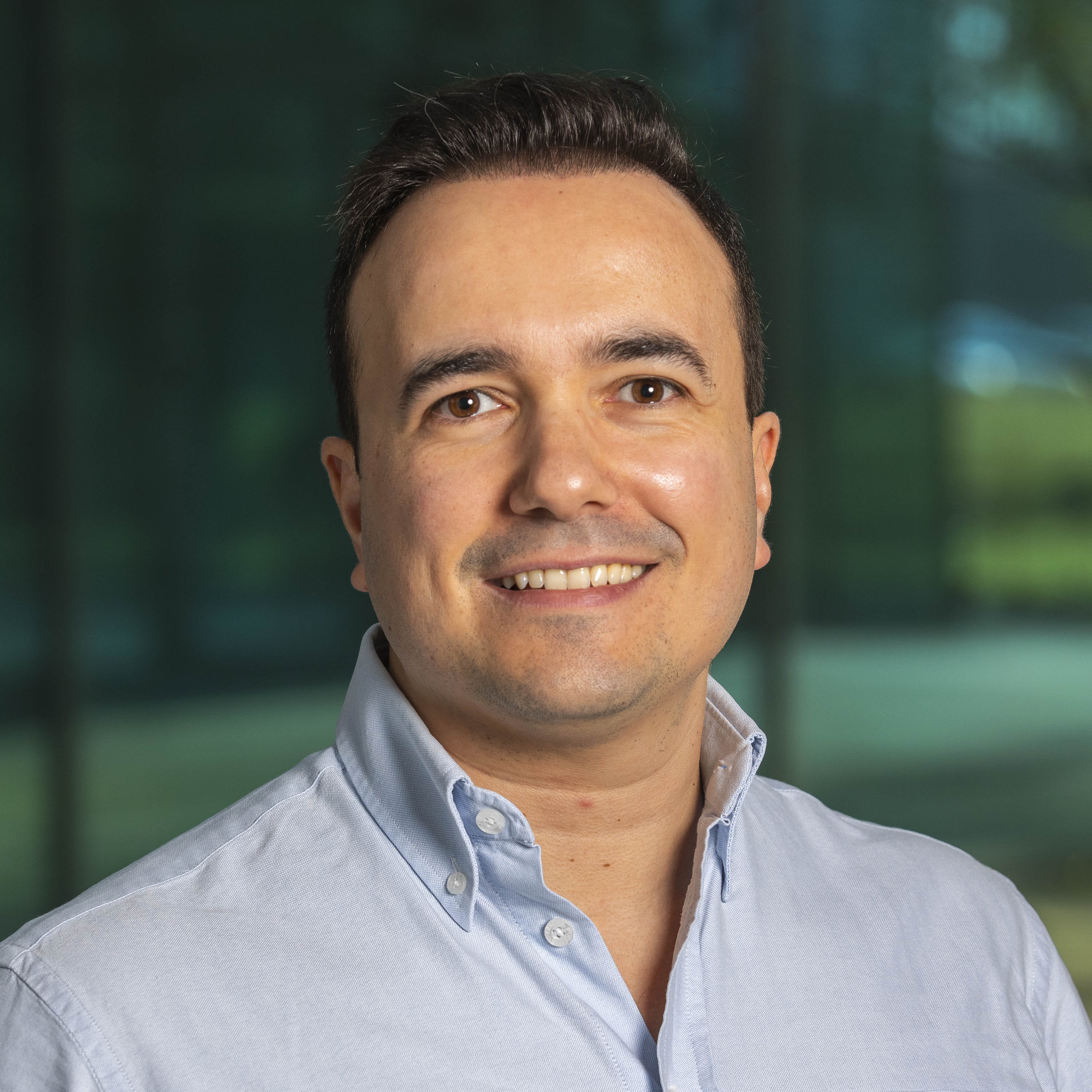}}]{João Silvério}
is a senior researcher and group leader at the German Aerospace Center (DLR) in Munich, where he leads the Interactive Skill Learning group since 2022.
He received a PhD in Robotics from the Italian Institute of Technology (IIT), Genoa, in 2017.
From 2017 to 2019, he was a postdoctoral researcher at IIT, and from 2019 to 2022, he held a postdoctoral position at the Idiap Research Institute in Martigny, Switzerland.
His research focuses on machine learning for robotics, with particular interest in learning for control, imitation learning, and reinforcement learning.
\end{IEEEbiography}
\vfill{}
\end{document}